\documentclass[]{fairmeta}
\usepackage{amsmath,amsfonts,bm}

\def\eqref#1{equation~\ref{#1}}
\def\1{\bm{1}}

\def\ve{{\bm{e}}}

\def\vh{{\bm{h}}}

\def\mW{{\bm{W}}}

\DeclareMathAlphabet{\mathsfit}{\encodingdefault}{\sfdefault}{m}{sl}
\SetMathAlphabet{\mathsfit}{bold}{\encodingdefault}{\sfdefault}{bx}{n}

\usepackage{amssymb}
\usepackage{graphicx}
\usepackage{pgfplots}

\usepackage{wrapfig}
\usepackage{caption}
\usepackage{algorithm}
\usepackage{algorithmic}
\usepackage{dsfont}

\usepackage{siunitx}
\PassOptionsToPackage{table}{xcolor}    % added to load xcolor properly
\usepackage[table]{xcolor}
\usepackage{colortbl}                   % added to load xcolor properly
\usepackage{multirow}
\usepackage{adjustbox}
\usepackage{enumitem}

\newif\ifshowcomments
\showcommentsfalse     % uncomment to hide feedback in PDF

\usepackage{array}

\title{LPDS: Evaluating LLM Robustness Through Logic-Preserving Difficulty Scaling}

\author[1,2,3,*]{Philipp Mondorf}
\author[4, *, \dagger]{Samuel J. Bell}
\author[1,\dagger]{Jesse Dodge}
\author[1,\dagger]{Dieuwke Hupkes}

\affiliation[1]{FAIR at Meta}
\affiliation[2]{LMU Munich}
\affiliation[3]{Munich Center for Machine Learning}
\affiliation[4]{Slingshot AI}

\contribution[*]{Work done at Meta}
\contribution[\dagger]{Joint last author, in random order}

\abstract{
As large language models (LLMs) are increasingly deployed to perform tasks with minimal human oversight, it is crucial that these models operate robustly. In particular, a model that can solve a given problem should not fail simply because certain entities\textemdash{}such as names, numbers, or other contextual details\textemdash{}have changed while the underlying problem logic remains the same. Prior work suggests that current LLMs still struggle with this form of robustness: they often succeed on some variations of a problem but fail on others. However, existing evaluations often lack a systematic way to identify which logic-preserving variations are most likely to induce failure. Instead, they typically test a random subset of allowable variations, which can overstate robustness. To address this gap, we introduce \emph{logic-preserving difficulty scaling} (LPDS), a framework that (i) quantifies the difficulty of a problem variation and (ii) systematically searches the space of allowable variations to find those that maximize difficulty and expose failures. We show that as difficulty increases, performance declines and errors in the models' reasoning chains become more pronounced. We further demonstrate that LPDS efficiently finds difficult problem variations for a model, resulting in performance drops up to $5\times$ larger compared to random sampling. Finally, we show that fine-tuning on more difficult variations leads to more consistent robustness gains than training on easier ones.
}

\date{\today}
\correspondence{\email{p.mondorf@lmu.de}}

\begin{document}

\maketitle

\begin{figure*}[h!]
  \centering
  \begin{minipage}[c]{1.0\textwidth}
    \centering
    \raisebox{0.55\height}{\rotatebox{90}{\textbf{\scriptsize{Accuracy}}}}%
    \hspace{0.3em}%
    \includegraphics[width=0.85\textwidth]{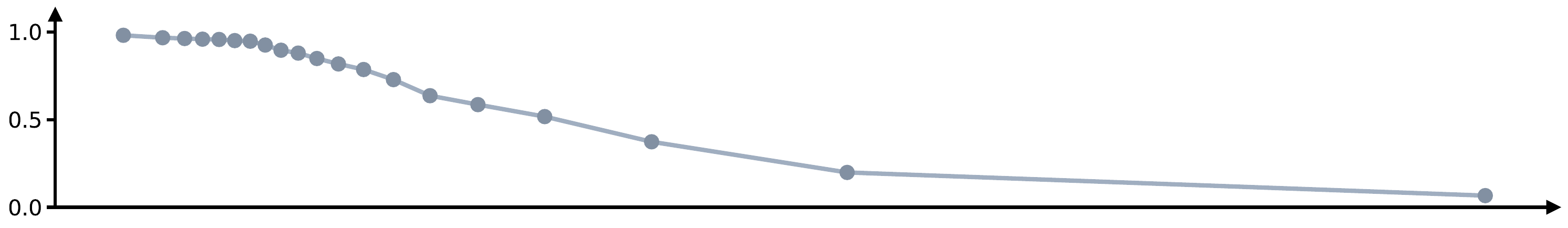}
    \hspace{0.3em}%
    \raisebox{0.4\height}{\textbf{\scriptsize{Difficulty $\left(\text{MD}_{\mathcal{H}}\right)$}}}%
  \end{minipage}
  
  \vspace{0.375em}

  % Second row
  \begin{minipage}[t!]{0.325\textwidth}
    \centering
    \input{tikz/intro_examples/low_md_example}
  \end{minipage}
  \hspace{0.00\textwidth}
  \begin{minipage}[t!]{0.325\textwidth}
    \centering
    \input{tikz/intro_examples/medium_md_example}
  \end{minipage}
  \hspace{0.00\textwidth}
  \begin{minipage}[t!]{0.325\textwidth}
    \centering
    \input{tikz/intro_examples/high_md_example}
  \end{minipage}

  \caption{\textbf{Conceptual overview.} As the estimated difficulty $\mathrm{MD}_{\mathcal{H}}$ of problem variations increases, accuracy declines (top) and model responses deviate more strongly from the correct solution trace (bottom).}
  \label{fig:intro_figure}
\end{figure*}

\section{Introduction}
\label{section:intro}
Large language models have demonstrated notable performance across various tasks and domains~\citep{10.1145/3641289, 10.1145/3747588, kamalloo-etal-2023-evaluating}. As a result, they are increasingly deployed as autonomous agents to carry out tasks with minimal human oversight~\citep{Xi2025, liu2025advances}. However, despite progress on specific benchmarks~\citep{ni2025survey, rein2024gpqa, NEURIPS2024_ad236edc}, recent studies suggest that LLMs still struggle to solve problems \emph{robustly}~\citep{10.5555/3692070.3692325, mondorf2024beyond, weber-etal-2023-mind}. For instance, models are often able to solve certain formulations of a problem while failing on others, even though the underlying problem logic remains the same~\citep{srivastava2024functional, 10.1162/tacl_a_00681}. This unreliability poses significant risks, as models may fail in ways that are difficult to anticipate, potentially leading to serious consequences in high-stakes scenarios~\citep{zhang2024agent}. To ensure the safe and reliable deployment of LLMs, it is therefore crucial to develop systematic ways to evaluate their robustness.

One approach to evaluating the robustness of LLMs is through \emph{symbolic templates}~\citep{mirzadeh2025gsmsymbolic, srivastava2024functional}. These templates represent a problem symbolically by specifying variables, constraints, and the corresponding solution procedure, while treating numbers, names, and other contextual details as parameters (see Figure~\ref{fig:symb_template_examples}). Symbolic templates can generate a diverse set of problem variations that differ in wording and values, while preserving the problem's underlying logic. Evaluating models on such a set of variations can reveal performance inflation due to data contamination and reliance on superficial patterns~\citep{mirzadeh2025gsmsymbolic}. However, because the space of possible problem variations defined by a symbolic template is often too large to evaluate exhaustively, existing studies typically assess models on a randomly sampled subset of instances~\citep{xie2025finchain, gull2025engtrace}. This can overestimate robustness, as challenging variations that are more likely to induce failure may be overlooked, leading to unwarranted confidence in the model.

To address this limitation, we introduce \emph{logic-preserving difficulty scaling} (LPDS), a systematic method for finding problem variations defined by a symbolic template that cause models to fail. First, we identify a set of metrics that successfully quantify the difficulty that a problem variation poses to a given model (Section~\ref{sec:quantifying_problem_variation_difficulty}). We show that as variation difficulty increases, model performance declines and errors in the models' reasoning chains become more pronounced (see Figure~\ref{fig:intro_figure}). Next, we introduce a discrete prompt-optimization procedure that efficiently searches the space of problem variations defined by a symbolic template to find those that maximize difficulty (Section~\ref{sec:lpds}). We demonstrate that, at each optimization step, the search procedure identifies progressively more difficult problem variations that cause models to fail. Notably, when evaluating models on a set of problem variations identified via LPDS, we observe performance drops relative to the original base problems that are up to five times larger than those obtained through random sampling from a template, underscoring both the effectiveness of our optimization procedure in uncovering difficult variations and the persistent brittleness of LLMs. Finally, when grouping problem variations by difficulty score, we show that fine-tuning models on more difficult samples yields more consistent robustness gains than fine-tuning on easier ones, further highlighting the usefulness of our difficulty estimate in characterizing problem variations from a given template (Section~\ref{sec:improving_robustness}). Overall, LPDS provides an efficient framework for evaluating LLM robustness by systematically identifying problem variations that reveal model failures.

\section{Preliminaries}\label{sec:preliminaries}

\subsection{Symbolic templates}
\label{subsec:symbolic_templates}
In natural language processing, a common method for generating synthetic data is through templates~\citep{mccoy-etal-2019-right, rottger-etal-2021-hatecheck, 10.5555/3666122.3669119, hupkes2018visualisation}. Templates are structured patterns that combine fixed text with variable slots or placeholders that can be filled with different values. While enabling rapid data generation, most templates focus on varying lexical choices~\citep{10.1145/3306618.3317950, ribeiro-etal-2020-beyond}. For instance, a template like ``I want to order \{\textsc{food}\} from \{\textsc{restaurant}\}'' can generate multiple prompts by substituting different food items and restaurant names.

\begin{figure*}[tbp]
\centering
\noindent
\begin{minipage}{0.49\textwidth}

\begin{tcolorbox}[
    title=\footnotesize{GSM-Symbolic Template},
    fontupper=\scriptsize,
    fontlower=\scriptsize,
    colback=metabg,
    colframe=metafg,
    coltitle=white,
    colbacktitle=metafg,
    coltext=metafg,
    boxsep=1pt,
    top=3pt,
    bottom=3pt,
    left=6pt,
    right=6pt,
    height=8.675cm,
    valign=top,
]
% Problem Statement
\textbf{\scriptsize\textcolor{metafg}{Problem}}
\vspace{2pt}
\hrule height 0.3pt
\vspace{4pt}

\highlight{cyan20}{\{\textsc{name}\}} can peel \highlight{purple30}{\{\textsc{n1}\}} \highlight{orange20}{\{\textsc{food}\}} a minute and saute \highlight{magenta20}{\{\textsc{n2}\}} \highlight{orange20}{\{\textsc{food}\}} in \highlight{yellow20}{\{\textsc{t}\}} minutes. How long will it take her to peel and saute \highlight{green20}{\{\textsc{total}\}} of \highlight{orange20}{\{\textsc{food}\}}?

\vspace{5pt}

% Variables
\textbf{\scriptsize\textcolor{metafg}{Variables}}
\vspace{2pt}
\hrule height 0.3pt
\vspace{4pt}

\begin{tabular}{@{}l@{\,}l@{}}
\highlight{cyan20}{\textsc{name}} & $\in$ \{Sophia, Claire, $\dots$, Sarah\}\\
\highlight{orange20}{\textsc{food}} & $\in$ \{shrimps, onions, $\dots$, clams\}\\
\highlight{purple30}{\textsc{n1}} & $\in$ \{4, 5, 6, $\dots$, 14, 15\}\\
\highlight{magenta20}{\textsc{n2}} & $\in$ \{20, 25, 30, $\dots$, 45, 50\}\\
\highlight{yellow20}{\textsc{t}} & $\in$ \{5, 6, 7, $\dots$, 19, 20\}\\
\highlight{green20}{\textsc{total}} & $\in$ \{60, 70, 80, $\dots$, 190, 200\}\\
\end{tabular}

\vspace{5pt}

% Conditions
\textbf{\scriptsize\textcolor{metafg}{Conditions}}
\vspace{2pt}
\hrule height 0.3pt
\vspace{4pt}

\begin{tabular}{@{}l@{\,}l@{}}
& divides(\highlight{green20}{\{\textsc{total}\}}, \highlight{purple30}{\{\textsc{n1}\}})\\
& divides(\highlight{green20}{\{\textsc{total}\}}, \highlight{magenta20}{\{\textsc{n2}\}})\\
\end{tabular}

\vspace{5pt}

% Ground Truth
\textbf{\scriptsize\textcolor{metafg}{Ground-Truth Reasoning}}
\vspace{2pt}
\hrule height 0.3pt
\vspace{4pt}

First, let's compute how long it takes \highlight{cyan20}{\{\textsc{name}\}} to peel the food: \highlight{green20}{\{\textsc{total}\}} of \highlight{orange20}{\{\textsc{food}\}} / \highlight{purple30}{\{\textsc{n1}\}} minutes.

\vspace{3pt}

Next, we calculate how many batches of \highlight{orange20}{\{\textsc{food}\}} she needs to cook: \highlight{green20}{\{\textsc{total}\}} / \highlight{magenta20}{\{\textsc{n2}\}} batches.

\vspace{3pt}

Now, we multiply the number of batches by the time per batch to find the total cooking time: \highlight{green20}{\{\textsc{total}\}} / \highlight{magenta20}{\{\textsc{n2}\}} $\times$ \highlight{yellow20}{\{\textsc{t}\}}.

\vspace{3pt}

To find the final time, we add both peeling and cooking time: \highlight{green20}{\{\textsc{total}\}} / \highlight{purple30}{\{\textsc{n1}\}} + \highlight{green20}{\{\textsc{total}\}} / \highlight{magenta20}{\{\textsc{n2}\}} $\times$ \highlight{yellow20}{\{\textsc{t}\}}.

%\#\#\#\# \highlight{green20}{\{\textsc{total}\}} / \highlight{purple30}{\{\textsc{n1}\}} + \highlight{green20}{\{\textsc{total}\}} / \highlight{magenta20}{\{\textsc{n2}\}} $\times$ \highlight{yellow20}{\{\textsc{t}\}}
\end{tcolorbox}

\end{minipage}
\hfill
\begin{minipage}{0.49\textwidth}

\begin{tcolorbox}[
    title=\footnotesize{FinChain Template},
    fontupper=\scriptsize,
    fontlower=\scriptsize,
    colback=metabg,
    colframe=metafg,
    coltitle=white,
    colbacktitle=metafg,
    coltext=metafg,
    boxsep=1pt,
    top=3pt,
    bottom=3pt,
    left=6pt,
    right=6pt,
    height=8.675cm,
    valign=top,
]
% Problem Statement
\textbf{\scriptsize\textcolor{metafg}{Problem}}
\vspace{2pt}
\hrule height 0.3pt
\vspace{4pt}

\highlight{cyan20}{\{\textsc{company}\}} reports an income tax expense of \$\highlight{orange20}{\{\textsc{tax expense}\}}. At the beginning of the year, the company had \$\highlight{purple30}{\{\textsc{start tax}\}} in taxes payable, and at the end of the year, taxes payable changed to \$\highlight{magenta20}{\{\textsc{end tax}\}}. Calculate \highlight{cyan20}{\{\textsc{company}\}}'s cash outflow for taxes.

\vspace{6pt}

% Variables
\textbf{\scriptsize\textcolor{metafg}{Variables}}
\vspace{2pt}
\hrule height 0.3pt
\vspace{4pt}

\begin{tabular}{@{}l@{\,}l@{}}
\highlight{cyan20}{\textsc{company}} & = \{Apple, Microsoft, $\dots$, Nike\}\\
\highlight{orange20}{\textsc{tax expense}} & = \{25000, 30000, $\dots$, 45000, 50000\}\\
\highlight{purple30}{\textsc{start tax}} & = \{5000, 5100, 5200, $\dots$, 9900, 10000\}\\
\highlight{magenta20}{\textsc{end tax}} & = \{8000, 8100, 8200, $\dots$, 15900, 16000\}\\
\end{tabular}

\vspace{6pt}

% Conditions
\textbf{\scriptsize\textcolor{metafg}{Conditions}}
\vspace{2pt}
\hrule height 0.3pt
\vspace{4pt}

\begin{tabular}{@{}l@{\,}l@{}}
\highlight{magenta20}{\{\textsc{end tax}\}} & $\geq$ \highlight{purple30}{\{\textsc{start tax}\}}\\
\end{tabular}

\vspace{6pt}

% Ground Truth
\textbf{\scriptsize\textcolor{metafg}{Ground-Truth Reasoning}}
\vspace{2pt}
\hrule height 0.3pt
\vspace{4pt}
We first calculate the change in taxes payable and then determine \highlight{cyan20}{\{\textsc{company}\}}'s cash outflow for taxes.

\vspace{5pt}

Step 1. Change in Taxes Payable = Closing Taxes Payable - Opening Taxes Payable = \$\highlight{magenta20}{\{\textsc{end tax}\}} - \$\highlight{purple30}{\{\textsc{start tax}\}}

\vspace{5pt}

Step 2. Tax Cash Outflow = Tax Expense - Change in Taxes Payable = \$\highlight{orange20}{\{\textsc{tax expense}\}} - (\$\highlight{magenta20}{\{\textsc{end tax}\}} - \$\highlight{purple30}{\{\textsc{start tax}\}})

\vspace{5pt}

Thus, \highlight{cyan20}{\{\textsc{company}\}}'s final cash outflow for taxes is:\\\$\highlight{orange20}{\{\textsc{tax expense}\}} - \$\highlight{magenta20}{\{\textsc{end tax}\}} + \$\highlight{purple30}{\{\textsc{start tax}\}}.

\end{tcolorbox}

\end{minipage}
\caption{\textbf{Example symbolic templates.} Example templates from GSM-Symbolic~(\citealt{mirzadeh2025gsmsymbolic}; left) and FinChain~(\citealt{xie2025finchain}; right). An example template from EngTrace~\citep{gull2025engtrace} is shown in Figure~\ref{fig:symb_template_example_engtrace} in the Appendix.}
\label{fig:symb_template_examples}
\end{figure*}

Symbolic templates, as used in GSM-Symbolic~\citep{mirzadeh2025gsmsymbolic}, FinChain~\citep{xie2025finchain}, and EngTrace~\citep{gull2025engtrace}, extend this concept to a deeper level of abstraction. Rather than varying surface tokens, symbolic templates encode the underlying logical structure of a problem. They define variables, constraints, and the complete reasoning procedure, while treating numbers, names, and other contextual details as parameters (see Figure~\ref{fig:symb_template_examples} for examples). Instantiating a symbolic template therefore yields many problem variations that differ in wording and values but share the same reasoning graph. For a more formal description of symbolic templates, please refer to Appendix~\ref{app:datasets}.

By generating variations of a problem that differ in surface form but share the same underlying logic, symbolic templates can reveal inflated model performance driven by data contamination and superficial pattern matching. However, existing work lacks a systematic method for selecting which variations to generate from a template\textemdash{}models are typically evaluated on a subset of variations sampled at random~\citep{mirzadeh2025gsmsymbolic}. Specifically, there is no principled approach for identifying the variations that most effectively expose model failures, leading to potential overestimation of robustness and unwarranted confidence in the model. In this work, we present a systematic framework that addresses this gap. Before introducing our approach, we first outline our evaluation setup.

\subsection{Evaluation setup}\label{subsec:eval_setup}
\paragraph{\textbf{Datasets.}} We use symbolic templates from three datasets: (i) GSM-Symbolic~\citep{mirzadeh2025gsmsymbolic}, derived from the math word problems of GSM8K~\citep{cobbe2021training} (left of Figure~\ref{fig:symb_template_examples}); (ii) FinChain~\citep{xie2025finchain}, which contains symbolic templates for financial multi-step reasoning problems (right of Figure~\ref{fig:symb_template_examples}); and (iii) EngTrace~\citep{gull2025engtrace}, which provides templates for multi-step reasoning problems in chemical, electrical, and mechanical engineering (Figure~\ref{fig:symb_template_example_engtrace} in the Appendix). From each dataset, we select 100 symbolic templates, each of which can generate a broad range of problem variations. Further details about each dataset are provided in Appendix~\ref{app:datasets}.

\paragraph{\textbf{Models.}} For our experiments, we evaluate Llama-3.2-3B-Instruct, Llama-3.1-8B/70B-Instruct~\citep{grattafiori2024llama}, Qwen-2.5-7B/32B/72B-Instruct~\citep{qwen2025qwen25technicalreport}, and Phi-4~\citep{abdin2024phi}. Further details about each model are provided in Appendix~\ref{app:models}.

\paragraph{\textbf{Evaluating responses.}} Following~\citet{mirzadeh2025gsmsymbolic}, we evaluate models using five-shot chain-of-thought (CoT) prompting~\citep{NEURIPS2022_9d560961} and greedy decoding. When necessary, we adapt templates so that each problem asks for a single numeric answer. Models are instructed to provide their final numeric answer after the $\#\#\#\#$ prefix, as illustrated by the CoT examples in each prompt (Figures~\ref{fig:gsms_prompt}--\ref{fig:engtrace_prompt} in Appendix~\ref{app:prompts}). We extract the model's final numeric answer by parsing the last decimal in the response with regular expressions. Since GSM-Symbolic answers are integers, we use exact match; for FinChain and EngTrace, a prediction $\hat{y}$ is correct if it falls within the tolerance bounds: $|\hat{y} - y| \leq 1 \cdot 10^{-6} + 1 \cdot 10^{-2}|y|$, where $y$ denotes the ground-truth.

\section{Quantifying problem variation difficulty}
\label{sec:quantifying_problem_variation_difficulty}
To enable efficient evaluation of model robustness, we seek to identify which problem variations defined by a symbolic template $\mathcal{T}$ are most likely to induce failure. To this end, we first assess the difficulty that a problem variation poses to a given model.

\subsection{Reference-based distance metrics}
\label{subsec:ref_based_distance}
Intuitively, we expect problem variations that are difficult for a model to elicit responses that deviate more from the correct solution path than responses to easier variations. Accordingly, we quantify the difficulty of a problem variation $p_i \in \mathcal{T}$ as the distance between the model's response $Y_i$ and the template's reasoning graph (see Figure~\ref{fig:symb_template_examples}). We represent this graph using a reference set $\mathcal{Y} = \{Y_k\}_{k=1}^N$ of responses to correctly answered variations $p_k \neq p_i$ from the same template. Larger distances indicate greater deviation from the correct solution trace and, consequently, a lower likelihood of correctness. In particular, for two incorrectly answered variations, the one closer to $\mathcal{Y}$ is more likely to contain partially correct reasoning, whereas the more distant one is more likely to reflect severe model failures (cf. Figure~\ref{fig:intro_figure}).

We consider two reference-based distance metrics: (i) \emph{Levenshtein distance} and (ii) \emph{Mahalanobis distance}. We show that both metrics effectively detect model failures and significantly outperform alternative measures in predicting answer correctness.

\paragraph{\textbf{Levenshtein distance.}} The Levenshtein distance is a simple metric for capturing lexical similarity between two strings. It quantifies the minimum number of single-token insertions, deletions, and substitutions required to transform one string into another. To measure the distance from $Y_i$ to the reference set $\mathcal{Y}$, we compute the (normalized) Levenshtein distance of $Y_i$ to each $Y_{k} \in \mathcal{Y}$. We then take the minimum of these distances, $\mathrm{LD}_{\min}$, to quantify the closest lexical match between the model's response and the reference set for a template. By construction, $\mathrm{LD}_{\min} \in [0,1]$, where $0$ indicates identical token sequences and values closer to $1$ indicate greater dissimilarity. A formal definition of $\mathrm{LD}_{\min}$ is provided in Appendix~\ref{app:formal_definitions}.

\paragraph{\textbf{Mahalanobis distance.}} While Levenshtein distance captures lexical similarity, a model's latent space often encodes higher-level semantic information~\citep{skean2025layer}. We therefore follow~\citet{ren2023outofdistribution} and compare the model's response to the reference set in latent space. For a response $Y_i = (y_1, y_2, \dots, y_M)$, we compute the average hidden-state vector at layer $l$ as $\mathbf{H}_i^{(l)} = \frac{1}{M} \sum_{t=1}^{M} \vh_t^{(l)},$ where $\vh_t^{(l)} \in \mathbb{R}^d$ is the hidden-state vector for output token $y_t$. We compute this quantity for each response in the reference set, yielding a set of average hidden-state vectors over correctly answered problem variations, $\mathcal{H}=\{\mathbf{H}_k^{(l)}\}_{k=1}^N$. We then measure how out-of-distribution $\mathbf{H}_i^{(l)}$ is with respect to $\mathcal{H}$ by fitting a Gaussian distribution $\mathcal{N}(\mu_{\mathcal{H}}, \Sigma_{\mathcal{H}})$ to $\mathcal{H}$ and computing the Mahalanobis distance:

\begin{equation}\label{eq:MD_response}
    \text{MD}_{\mathcal{H}} = \text{MD}(\mathbf{H}_i^{(l)}; \mu_{\mathcal{H}}, \Sigma_{\mathcal{H}}) = (\mathbf{H}_i^{(l)} - \mu_{\mathcal{H}})^T \Sigma_{\mathcal{H}}^{-1} (\mathbf{H}_i^{(l)} - \mu_{\mathcal{H}})
\end{equation}

where $\mu_{\mathcal{H}} \in \mathbb{R}^d$ and $\Sigma_{\mathcal{H}} \in \mathbb{R}^{d \times d}$. Based on initial experiments, we set $l$ to be roughly two-thirds through the model (e.g., $l=21$ for a 32-layer model) and $N=200$ (see Appendix~\ref{app:reference_based_metrics} for further details). Higher $\text{MD}_{\mathcal{H}}$ values indicate greater deviation from the reference distribution and thus a lower likelihood of following the template's reasoning graph.

\begin{figure*}[tbp]
     \centering
     \captionsetup[subfigure]{font=scriptsize}
     \begin{subfigure}[b]{0.245\textwidth}
         \centering
         \begin{tikzpicture}
        \node[inner sep=0] (img) at (0,0) {%
            \includegraphics[
                width=0.957\linewidth,
                trim={1.0cm 0.cm 0.cm 0.cm},
                clip
            ]{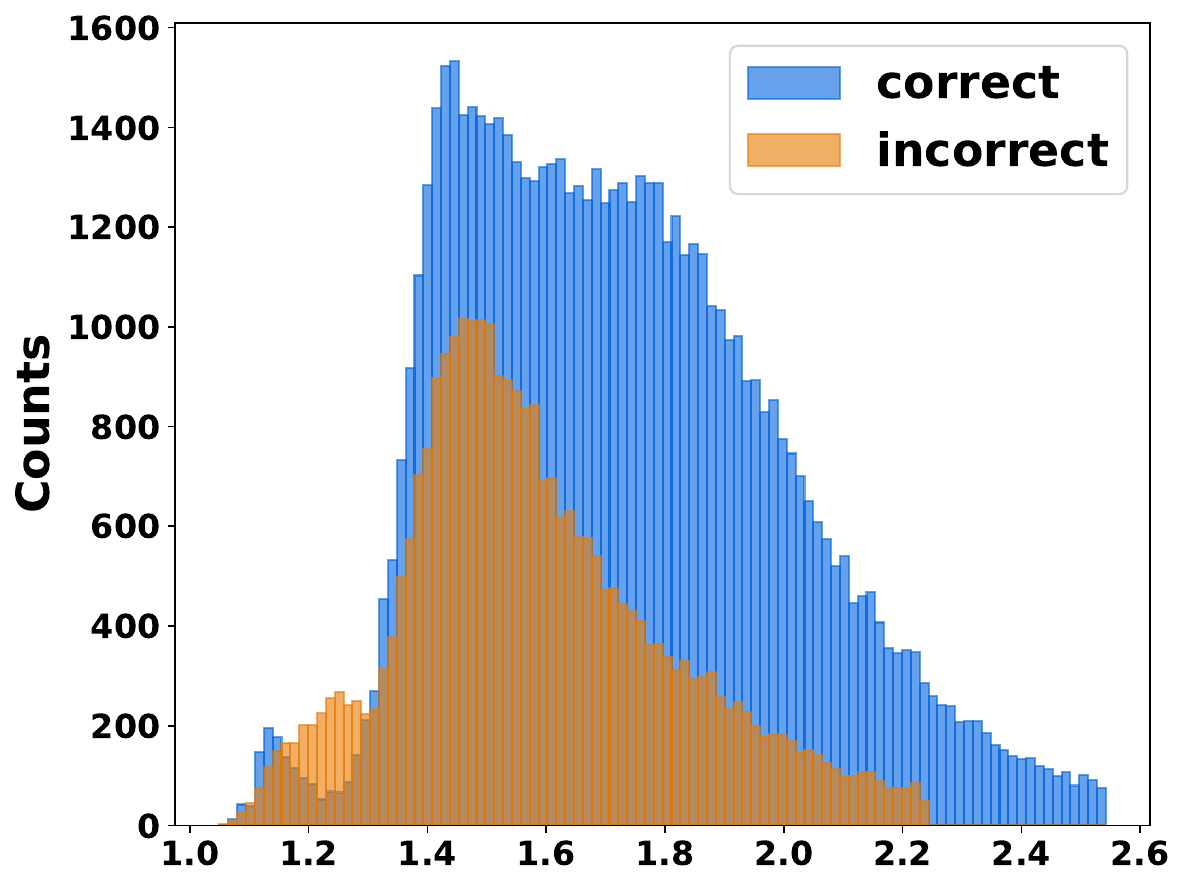}
         };
    \node[
            rotate=90,
            font=\fontsize{6pt}{7pt}\selectfont\sffamily,
            anchor=south
        ] at ([xshift=0.115cm]img.west) {Counts};
    \end{tikzpicture}
         \caption{Perplexity}
         \label{fig:plots:a}
     \end{subfigure}
     \begin{subfigure}[b]{0.245\textwidth}
         \centering
         \includegraphics[width=\linewidth]{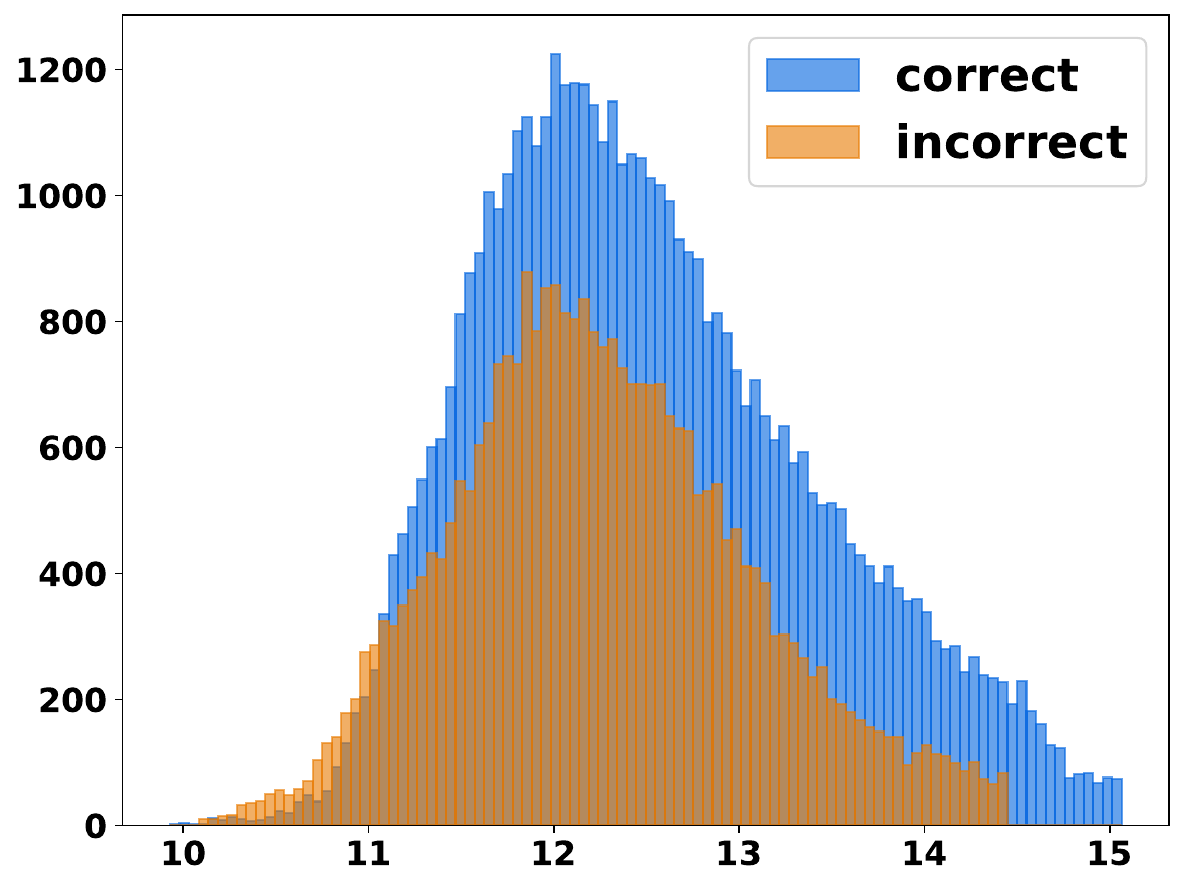}
         \caption{Self-Certainty}
         \label{fig:plots:b}
     \end{subfigure}
     \begin{subfigure}[b]{0.245\textwidth}
         \centering
         \includegraphics[width=\linewidth]{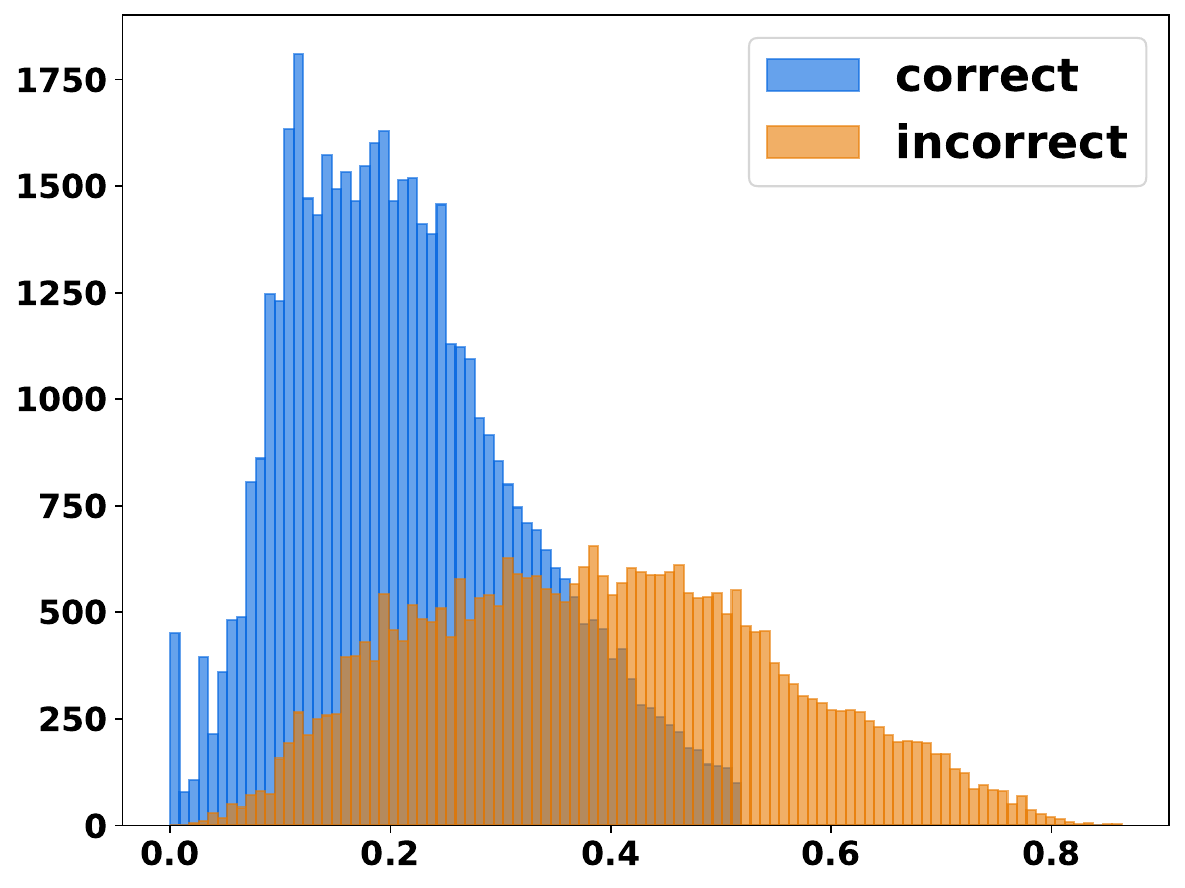}
         \caption{Levenshtein distance}
         \label{fig:plots:c}
     \end{subfigure}
     \begin{subfigure}[b]{0.245\textwidth}
         \centering
         \includegraphics[width=\linewidth]{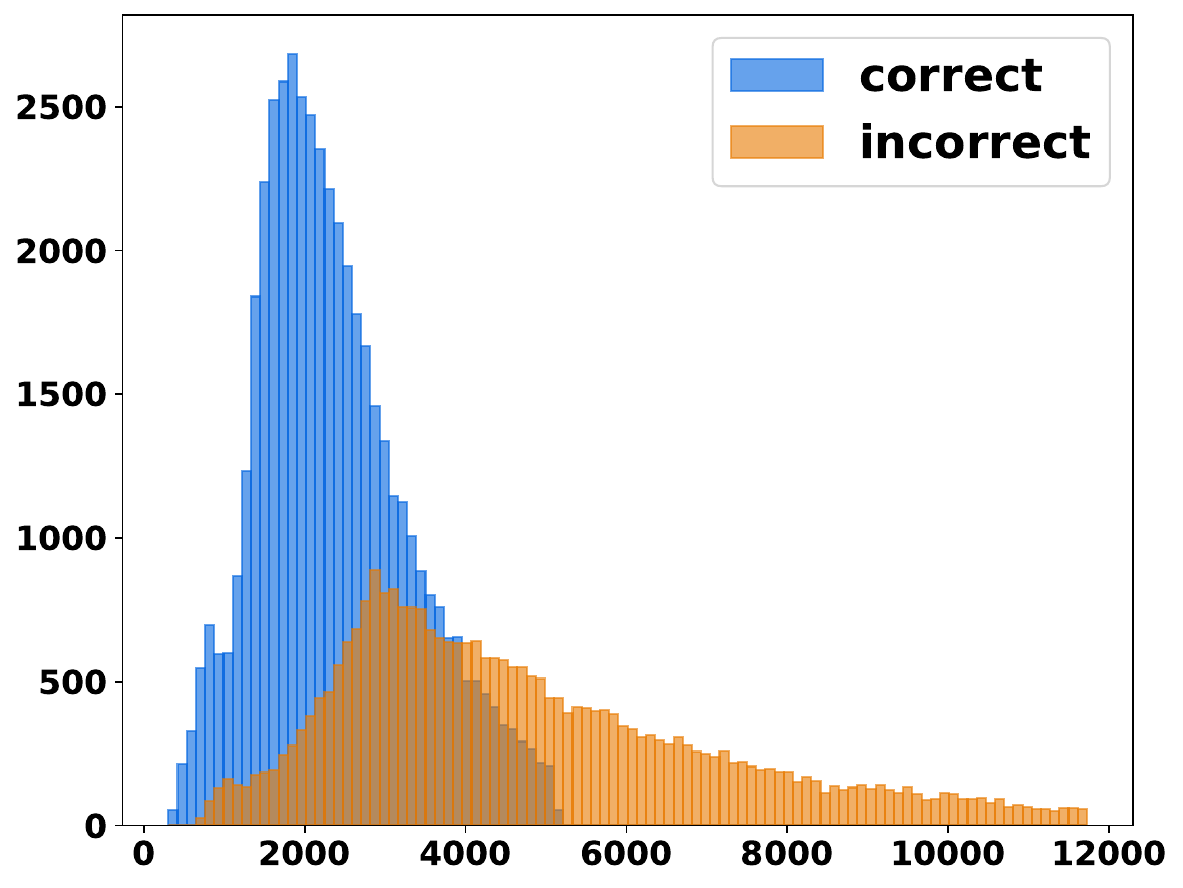}
         \caption{Mahalanobis distance}
         \label{fig:plots:d}
     \end{subfigure}
     \caption{\textbf{Levenshtein and Mahalanobis distances separate correct from incorrect responses better than other metrics.} Distribution of metric scores for responses from Llama-3.2-3B-Instruct across 1{,}000 samples per GSM-Symbolic template (100{,}000 samples in total). Corresponding results for other models and datasets, including FinChain and EngTrace, are shown in Figures~\ref{fig:predicting_answer_correctness_llama_8b_gsms}--\ref{fig:predicting_answer_correctness_qwen_32b_engtrace} in the Appendix.}
     \label{fig:predicting_answer_correctness_llama_3b_gsms}
\end{figure*}

\subsection{Capturing model failure}
\label{subsec:pred_answer_correctness}
We evaluate whether the proposed distance metrics successfully capture model failures. For each of the 100 templates from GSM-Symbolic, FinChain, and EngTrace (Section~\ref{subsec:eval_setup}), we generate 1{,}000 problem variations, yielding 100{,}000 problems per dataset. For each variation $p_i$, we generate the model response $Y_i$, record whether it is correct ($c_i=1$) or incorrect ($c_i=0$), and compute the difficulty metric $f_i$.

\begin{table}[b!]
\centering
\scriptsize
\begin{tabular}{
  lcl
  | S[table-format=1.2]
    S[table-format=-1.2]
    >{\hspace{-0.8em}}l
  | S[table-format=1.2]
    S[table-format=-1.2]
    >{\hspace{-0.8em}}l
  | S[table-format=1.2]
    S[table-format=-1.2]
    >{\hspace{-0.8em}}l
}
\toprule
& \multicolumn{1}{c}{\multirow{2}{*}{\textbf{Span}}}
& \multicolumn{1}{c}{\multirow{2}{*}{\textbf{Metric}}}
& \multicolumn{3}{c}{\textbf{GSM-Symbolic}}
& \multicolumn{3}{c}{\textbf{FinChain}}
& \multicolumn{3}{c}{\textbf{EngTrace}} \\
\cmidrule(lr){4-6}\cmidrule(lr){7-9}\cmidrule(lr){10-12}
& &
& \multicolumn{1}{c}{\textbf{AUC}}
& \multicolumn{2}{c}{\textbf{Odds Ratio}}
& \multicolumn{1}{c}{\textbf{AUC}}
& \multicolumn{2}{c}{\textbf{Odds Ratio}}
& \multicolumn{1}{c}{\textbf{AUC}}
& \multicolumn{2}{c}{\textbf{Odds Ratio}} \\
\toprule

\multirow{10}{*}{\rotatebox{90}{Llama-3.1-8B-Instruct}}
& \multirow{3}{*}{Input}
& Perplexity          & 0.53 & 0.84 & (0.82, 0.86) & 0.49 & 1.03 & (0.67, 1.61) & 0.54 &  0.31 & (0.30, 0.32) \\
& & Entropy            & 0.51 &  0.86 & (0.84, 0.88) & 0.50 & 0.99 & (0.97, 1.02) & 0.53 & 0.42 & (0.42, 0.44) \\
& & Self-Certainty         & 0.51 & 1.11 & (1.08, 1.13) & 0.50 & 0.93 & (0.90, 0.95) & 0.51 & 1.39 & (1.36, 1.43) \\

\cmidrule{2-12}

& \multirow{6}{*}{Output}
& Perplexity          & 0.44 &  0.96 & (0.94, 0.98) & 0.46 & 1.03 & (1.00, 1.05) & 0.48 & 1.08 & (1.05, 1.10) \\
& & Entropy            & 0.51 &  0.96 & (0.94, 0.99) & 0.53 & 0.84 & (0.82, 0.86) & 0.52 & 0.87 & (0.85, 0.89) \\
& & Self-Certainty         & 0.56 & 1.36 & (1.33, 1.39) & 0.54 & 1.14 & (1.11, 1.17) & 0.55 & 1.20 & (1.17, 1.22) \\
& & {\cellcolor{metabg}$\text{LD}_{\text{min}}$} & {\cellcolor{metabg}0.78} & {\cellcolor{metabg}0.18} & {\cellcolor{metabg}(0.18, 0.19)} & {\cellcolor{metabg}0.70} & {\cellcolor{metabg}\bfseries{0.16}} & {\cellcolor{metabg}(0.15, 0.17)} & {\cellcolor{metabg}0.76} & {\cellcolor{metabg}0.20} & {\cellcolor{metabg}(0.19, 0.21)} \\
& & {\cellcolor{metabg}$\text{MD}_\mathcal{H}$}   & {\cellcolor{metabg}\bfseries{0.80}} & {\cellcolor{metabg}\bfseries{0.14}} & {\cellcolor{metabg}(0.13, 0.14)} & {\cellcolor{metabg}\bfseries{0.74}} & {\cellcolor{metabg}0.49} & {\cellcolor{metabg}(0.47, 0.51)} & {\cellcolor{metabg}\bfseries{0.78}} & {\cellcolor{metabg}\bfseries{0.02}} & {\cellcolor{metabg}(0.02, 0.02)} \\
& & $\text{KNN}_\mathcal{H}$  & 0.71 & 0.40 & (0.38, 0.41) & 0.65 & 0.59 & (0.57, 0.61) & 0.67 & 0.28 & (0.22, 0.24) \\
\bottomrule
\end{tabular}
\caption{\textbf{Distance metrics outperform other metrics in predicting answer correctness}. AUC scores and odds ratios for predicting answer correctness across datasets. Among all metrics, $\mathrm{LD}_{\min}$ and $\mathrm{MD}_{\mathcal{H}}$ are the most predictive.}
\label{tab:correlation_llama_3.1_8B}
\end{table}

\paragraph{\textbf{Distance metrics successfully predict answer correctness.}} Figure~\ref{fig:predicting_answer_correctness_llama_3b_gsms} shows the distribution of $(f_i, c_i)$ pairs for Llama-3.2-3B-Instruct on GSM-Symbolic for different metrics $f$, including output perplexity, self-certainty~\citep[][definition in Appendix~\ref{app:formal_definitions}]{kang2025scalable}, $\mathrm{LD}_{\min}$, and $\mathrm{MD}_{\mathcal{H}}$. We find that both $\mathrm{LD}_{\min}$ (\ref{fig:plots:c}) and $\mathrm{MD}_{\mathcal{H}}$ (\ref{fig:plots:d}) visibly separate correctly answered variations from incorrectly answered ones, unlike perplexity (\ref{fig:plots:a}) and self-certainty (\ref{fig:plots:b}). Similar trends hold for other datasets and models, as shown in Figures~\ref{fig:predicting_answer_correctness_llama_8b_gsms}--\ref{fig:predicting_answer_correctness_qwen_32b_engtrace} in the Appendix.

Table~\ref{tab:correlation_llama_3.1_8B} further quantifies each metric's predictive power for answer correctness using two measures: the AUC, i.e., the probability that an incorrectly answered variation receives a higher difficulty score than a correctly answered one, and the odds ratio, i.e., the effect of a one-standard-deviation increase in the z-scored difficulty estimate on the odds of a correct answer (see Appendix~\ref{app:measuring_predictive_power} for details). Responses $Y_i$ are generated by Llama-3.1-8B-Instruct. We compare difficulty estimates $f$ computed over the input (i.e., using only $p_i$) with those computed over the output $Y_i$. In addition to the Mahalanobis distance, we compute $\mathrm{KNN}_{\mathcal{H}}$, the distance from $\mathbf{H}_i$ to its $k$-th nearest neighbor in $\mathcal{H}$ ($k=10$). Across all datasets, we find that both proposed distance metrics, $\mathrm{LD}_{\min}$ and $\mathrm{MD}_{\mathcal{H}}$, are highly predictive of answer correctness. $\mathrm{MD}_{\mathcal{H}}$ achieves the highest AUC scores, ranging from $0.74$ on FinChain to $0.80$ on GSM-Symbolic, and the strongest odds ratios on GSM-Symbolic $(\mathrm{OR}=0.14)$ and EngTrace $(\mathrm{OR}=0.02)$. Both $\mathrm{MD}_{\mathcal{H}}$ and $\mathrm{LD}_{\min}$ substantially outperform other metrics, such as output perplexity (AUC $0.44$--$0.48$) and self-certainty (AUC $0.54$--$0.56$). Input-based metrics generally show lower predictive power. Similar trends hold across other models and datasets. A more detailed overview, including additional metrics, is provided in Appendix~\ref{app:assessing_variation_difficulty}, specifically Tables~\ref{tab:correlation_llama_family}--\ref{tab:correlation_phi4_14b}.

\paragraph{\textbf{Errors become more pronounced as distance increases.}} In addition to our quantitative evaluation, we qualitatively analyze how model responses change as the difficulty estimate $\mathrm{MD}_{\mathcal{H}}$ increases. Figure~\ref{fig:qualitative_examples_template_55_qwen_7b_instruct} in the Appendix shows problem variations from template 55 in GSM-Symbolic together with the responses of Qwen-2.5-7B-Instruct. We observe that, as $\mathrm{MD}_{\mathcal{H}}$ increases, the model's verbalized reasoning deviates more strongly from the correct solution path. Specifically, low-$\mathrm{MD}_{\mathcal{H}}$ instances are mostly solved correctly, medium-$\mathrm{MD}_{\mathcal{H}}$ instances typically contain minor arithmetic errors, and high-$\mathrm{MD}_{\mathcal{H}}$ instances often exhibit severe failures. This pattern is consistent across templates (Figures~\ref{fig:qualitative_examples_template_74_qwen_7b_instruct}--\ref{fig:qualitative_examples_template_79_qwen_7b_instruct} in the Appendix) and aligns with the strong predictive performance of $\mathrm{LD}_{\min}$ and $\mathrm{MD}_{\mathcal{H}}$ presented in Table~\ref{tab:correlation_llama_3.1_8B}. Together, $\mathrm{LD}_{\min}$ and $\mathrm{MD}_{\mathcal{H}}$ not only predict whether an answer is correct, but also help rank incorrectly answered problem variations by the severity of the corresponding model failure.

\section{Logic-Preserving Difficulty Scaling}
\label{sec:lpds}
To find problem variations from a symbolic template that expose failures, we next design an optimization procedure that maximizes the difficulty estimates introduced in Section~\ref{sec:quantifying_problem_variation_difficulty}.

\subsection{Scaling difficulty via beam search}
\label{subsec:scaling_difficulty}
As described in Section~\ref{subsec:symbolic_templates}, a symbolic template defines the space of logic-preserving problem variations $p_i \in \mathcal{T}$ through a set of variable slots $S=\{s_1,\dots,s_K\}$. For each slot $s \in S$, the variable $v_s$ is drawn from a set of allowed substitutions $V_s$ (see Figure~\ref{fig:symb_template_examples}). When the number of slots $K$ or the sizes of the substitution sets $\lvert V_s \rvert$ are large, exhaustive search becomes infeasible. We therefore devise an optimization procedure that efficiently navigates the space of possible variations to find difficult ones using beam search.

The key idea is to explore variable substitutions one slot at a time and to assess difficulty in two stages: (i) a cheap approximation of difficulty, $\tilde f$, for filtering candidates, and (ii) a more accurate but expensive estimate, $f$ (such as $\mathrm{MD}_{\mathcal{H}}$ introduced in Section~\ref{sec:quantifying_problem_variation_difficulty}) for beam selection. Starting from a base problem or random variation $p_0 \in \mathcal{T}$, we compute its difficulty $f(p_0)$ for a model and initialize the beam as $\mathcal{B}=\{(p_0, f(p_0))\}$. We then run beam search for $T$ iterations with branching factor $b$ and beam width $w$. Each iteration proceeds as follows:

\begin{enumerate}[leftmargin=*, itemindent=0pt]
    \item \textbf{Candidate generation.} For each variation $p$ in the current beam, generate ``neighboring'' variations $p'$ by replacing one slot $s$ at a time with substitutions $v_s \in V_{s}$. Keep only valid candidates not already in the beam. Also explore non-neighboring variations by sampling a small fraction $\rho_{\text{expl}}$ of candidates directly from $\mathcal{T}$ at random.
    
    \item \textbf{Cheap scoring.} Compute $\tilde f(p')$ for all candidates $p'$ generated in Step~1.

    \item \textbf{Pruning.} Add $b$ candidates to the pool $\mathcal{C}$ for exact scoring. Specifically, choose the top-$\left((1-\rho_{\text{sel}})\cdot b\right)$ candidates according to $\tilde f$ and the remaining $\rho_{\text{sel}}\cdot b$ uniformly at random to encourage exploration and to mitigate potential misrankings from $\tilde f$.
            
    \item \textbf{Exact scoring.} Compute the more accurate difficulty estimate $f(p')$ for each $p'$ in $\mathcal{C}$.
        
    \item \textbf{Beam update.} Add the scored variations from Step~4 to the beam and keep the top-$w$ according to $f$.
    
    \item \textbf{Best-so-far update / termination.} Update $(p^{*}, f^{*})$ if the updated beam contains a higher-scoring variation. After $T$ iterations, return $p^{*}$ as the optimized (most difficult) variation.
\end{enumerate}

The pseudocode for the search procedure is presented in Algorithm~\ref{alg:beam_search} in the Appendix. For our experiments, we use $b = 16$, $w = 16$, $\rho_{\mathrm{expl}} = 0.2$, $\rho_{\text{sel}} = 0.4$ (details in Appendix~\ref{app:beam_search}).

A key component of our algorithm is its two-stage scoring procedure. When the number of candidates generated during Step~1 is large, computing the output-based difficulty estimates proposed in Section~\ref{subsec:ref_based_distance} for all candidates becomes expensive. We therefore use a cheaper approximation $\tilde f$ for pre-filtering. For a variation $p_i=(x_1,\ldots,x_L)$, we compute the average input embedding $\mathbf{E}_i=\frac{1}{L}\sum_{t=1}^{L}\mW_{\mathrm{emb}}[x_t]$, where $\mathbf{W}_{\mathrm{emb}}$ is the model's embedding matrix. Analogous to Equation~\ref{eq:MD_response}, we then compare $\mathbf{E}_i$ to a reference set $\mathcal{E}=\{\mathbf{E}_k\}_{k=1}^{N}$ built from correctly answered variations by computing the Mahalanobis distance $\mathrm{MD}_{\mathcal{E}}$ from $\mathbf{E}_i$ to the Gaussian distribution $\mathcal{N}(\mu_{\mathcal{E}}, \Sigma_{\mathcal{E}})$. This input-based score is efficient to compute and scales to large variation sets. Its predictive power is summarized in Tables~\ref{tab:correlation_llama_family}--\ref{tab:correlation_phi4_14b} in the Appendix.

\begin{figure*}[tbp]
  \centering
  \begin{subfigure}{0.485\textwidth}
    \centering
    \includegraphics[width=0.85\linewidth]{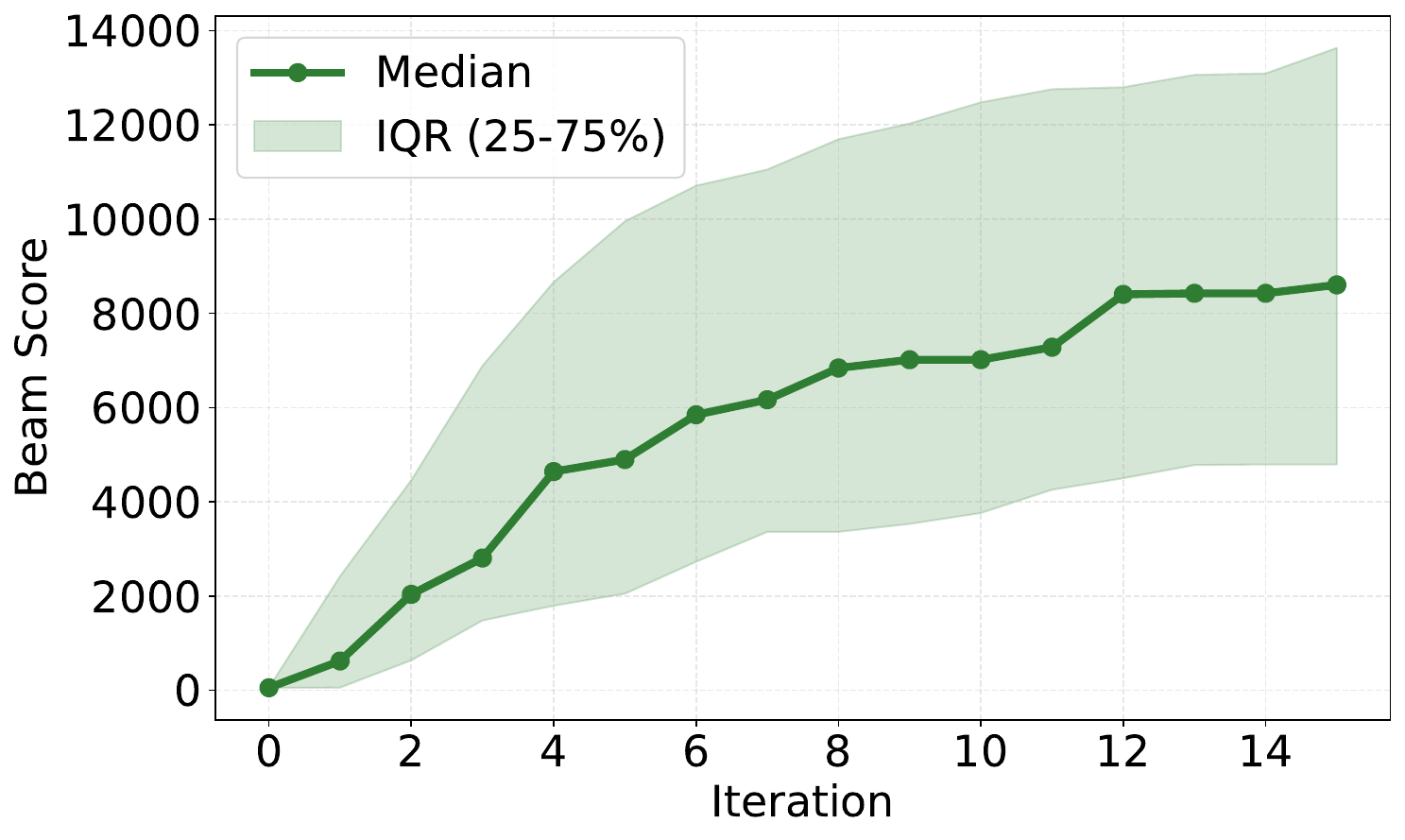}
    \caption{Top score $f^* = \text{MD}_{\mathcal{H}}^*$ in the current beam.}
    \label{fig:beam_score_history}
  \end{subfigure}
  \hspace{0.01\textwidth}
  \begin{subfigure}{0.485\textwidth}
    \centering
    \includegraphics[width=0.85\linewidth]{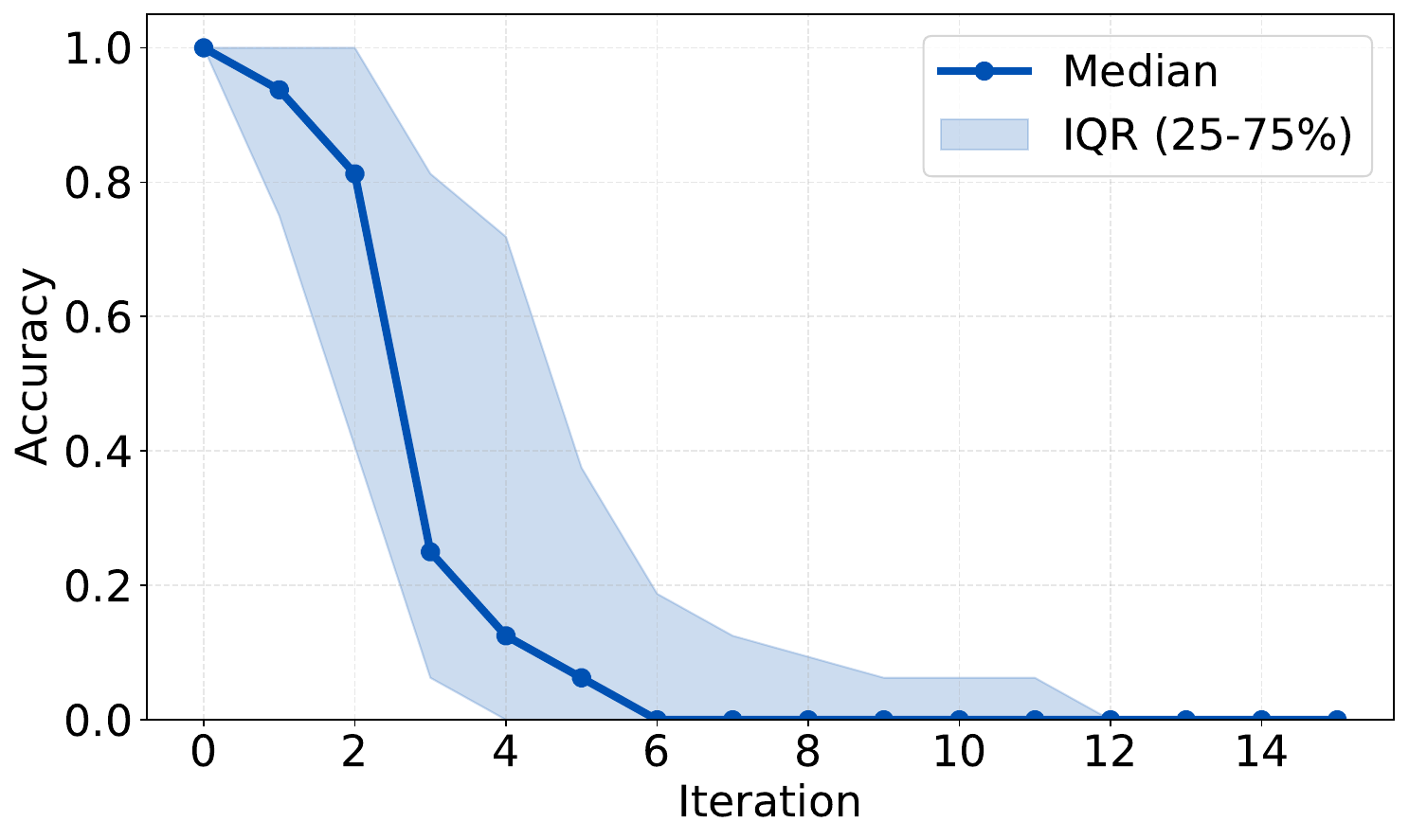}
    \caption{The model's average beam accuracy.}
    \label{fig:beam_acc_history}
  \end{subfigure}
  \caption{\textbf{Beam search finds increasingly difficult problem variations.} (a) Across iterations, variation difficulty increases, while (b) Llama-3.1-8B-Instruct’s beam accuracy decreases. Lines show medians, and shaded regions show interquartile ranges, across runs over all GSM-Symbolic templates. Similar trends for other models and datasets, including FinChain and EngTrace, are shown in Figures~\ref{fig:beam_search_gsms_qwen_7b}--\ref{fig:beam_search_engtrace_qwen_7b} in the Appendix.}
  \label{fig:beam_search_gsms_llama_8b}
\end{figure*}

\begin{figure*}[b!]
  \centering
  \begin{subfigure}[b]{0.355\textwidth}
    \centering
    \includegraphics[width=\linewidth]{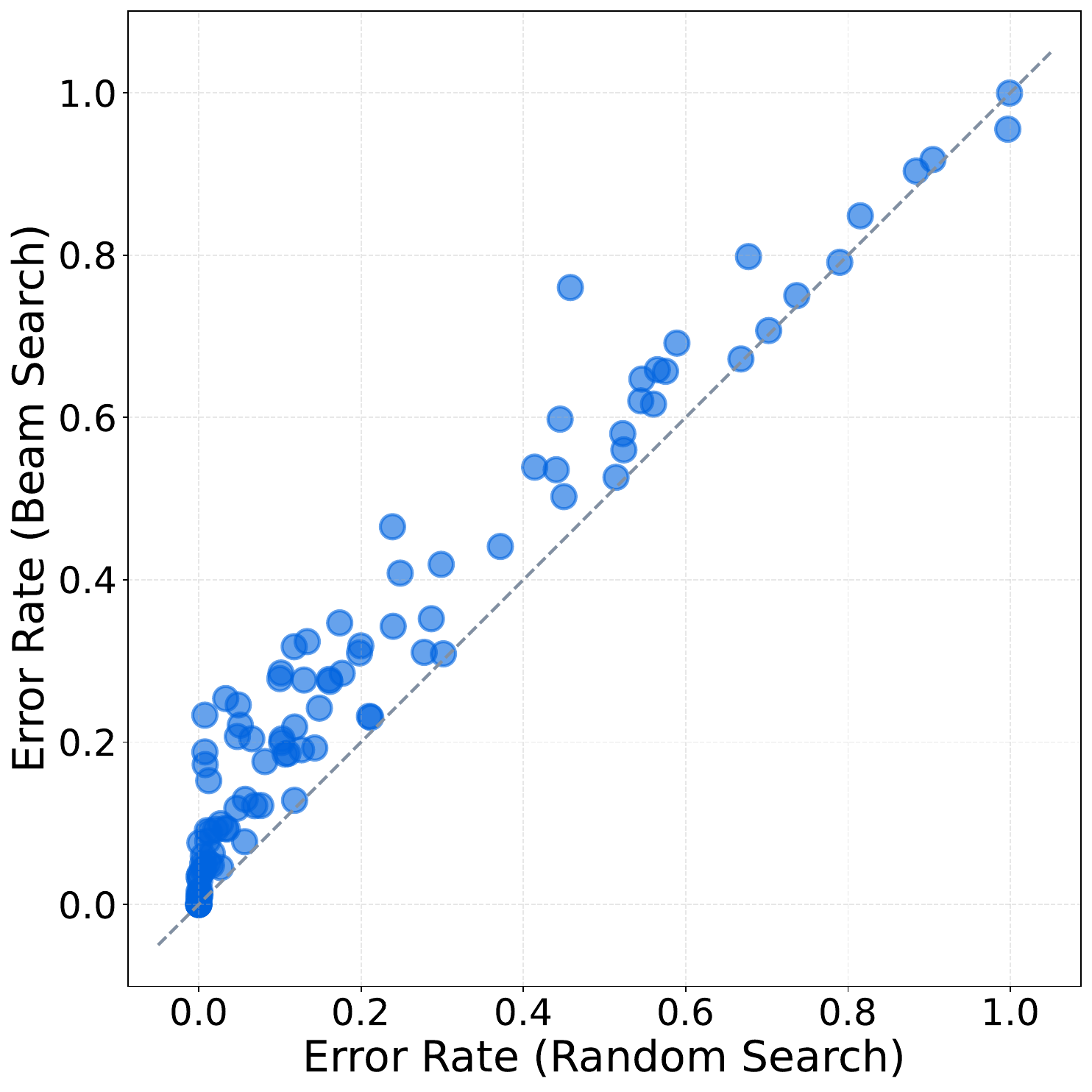}
    \caption{\small{Error rate comparison.}}
    \label{fig:gsms_llama8b_error_rate}
  \end{subfigure}
  \hspace{0.065\textwidth}
  \begin{subfigure}[b]{0.533\textwidth}
    \centering
    \includegraphics[width=\linewidth]{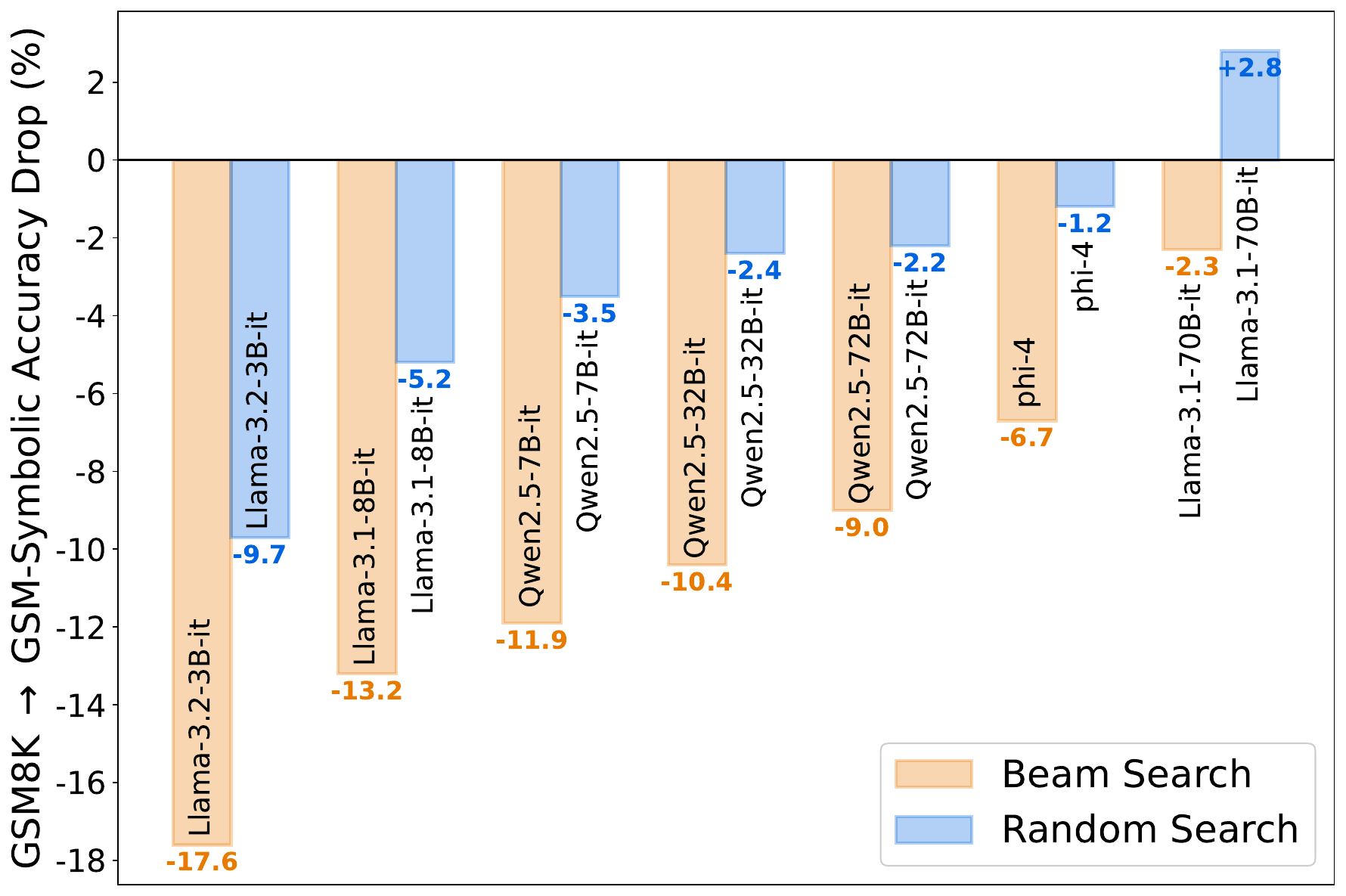}
    \caption{\small{Accuracy drop (\%) with respect to GSM8K.}}
    \label{fig:gsms_gsm8k_performance_drop}
  \end{subfigure}
  \caption{\textbf{LPDS captures brittleness more accurately than random search.} (a) Per-template error rates of Llama-3.1-8B-it on variations from GSM-Symbolic identified via beam search vs.\ random search. (b) Performance drop on GSM-Symbolic relative to GSM8K. Similar trends for other models and datasets are shown in Figures~\ref{fig:error_rate_comparison_gsms}--\ref{fig:error_rate_comparison_finchain} in the Appendix.}
  \label{fig:beam_search_vs_random_search_gsms}
\end{figure*}

\subsection{Exposing model failures}
\label{subsec:lpds_expose_failure}
We run the beam search algorithm from Section~\ref{subsec:scaling_difficulty} on each of the 100 GSM-Symbolic templates for $T = 15$ iterations to find difficult problem variations that cause models to fail.

\paragraph{\textbf{LPDS efficiently finds difficult problem variations.}} Figure~\ref{fig:beam_search_gsms_llama_8b} shows that, as search progresses, the maximum difficulty $f^* = \mathrm{MD}_{\mathcal{H}}^*$ among the problem variations in the beam increases (left), while Llama-3.1-8B-Instruct's average accuracy on those variations decreases (right). Notably, after only six iterations, the model's average beam accuracy drops to $0\%$ (median over 100 GSM-Symbolic templates). Similar trends hold across other models and datasets (Figures~\ref{fig:beam_search_gsms_qwen_7b}--\ref{fig:beam_search_engtrace_qwen_7b} in the Appendix), highlighting the efficiency of our method in identifying difficult problem variations that expose failures.

\paragraph{\textbf{LPDS captures robustness more accurately than random sampling.}} After beam search, up to $T \times w \times b$ variations per template have been explored and assigned a difficulty estimate $\mathrm{MD}_{\mathcal{H}}$. Figure~\ref{fig:gsms_gsm8k_performance_drop} shows the accuracy drop for each model on these variations relative to its baseline performance on the respective GSM8K problems. Across all models, performance degrades substantially: Llama-3.2-3B-Instruct's accuracy drops by 17.6\% despite the variations and base problems sharing the same logic, suggesting reliance on superficial cues and possible data contamination. Larger models also degrade (11.9\% for Qwen-2.5-32B-it and 9.0\% for Qwen-2.5-72B-it), while Llama-3.1-70B-Instruct is most robust, with a drop of 2.3\%. We demonstrate the effectiveness of LPDS in identifying challenging problem variations\textemdash{}and thereby assessing robustness\textemdash{}by comparing it with variations sampled at random from a template, following GSM-Symbolic~\citep{mirzadeh2025gsmsymbolic}. For a fair comparison, we sample the same number of variations per template as in beam search. As shown in Figure~\ref{fig:gsms_gsm8k_performance_drop}, the performance drop is notably smaller: e.g., 9.7\% (vs. 17.6\%) for Llama-3.2-3B-Instruct and 2.2\% (vs. 9.0\%) for Qwen-2.5-72B-Instruct. We further compare Llama-3.1-8B-Instruct's error rates per template and find that beam search identifies more difficult variations for nearly all templates (points above the diagonal in Figure~\ref{fig:gsms_llama8b_error_rate}). Similar trends hold across models and datasets, as shown in Figures~\ref{fig:error_rates},~\ref{fig:error_rate_comparison_gsms}, and~\ref{fig:error_rate_comparison_finchain} in Appendix~\ref{app:evaluating_robustness}.

\begin{wrapfigure}{r}{0.4575\textwidth}
  \vspace{-12.5pt}
  \centering
  \begin{tikzpicture}
        \node[inner sep=0] (img) at (0,0) {%
            \includegraphics[
                width=\dimexpr\linewidth-0.45cm\relax,
                trim={1.cm 1.cm 0.cm 0.cm},
                clip
            ]{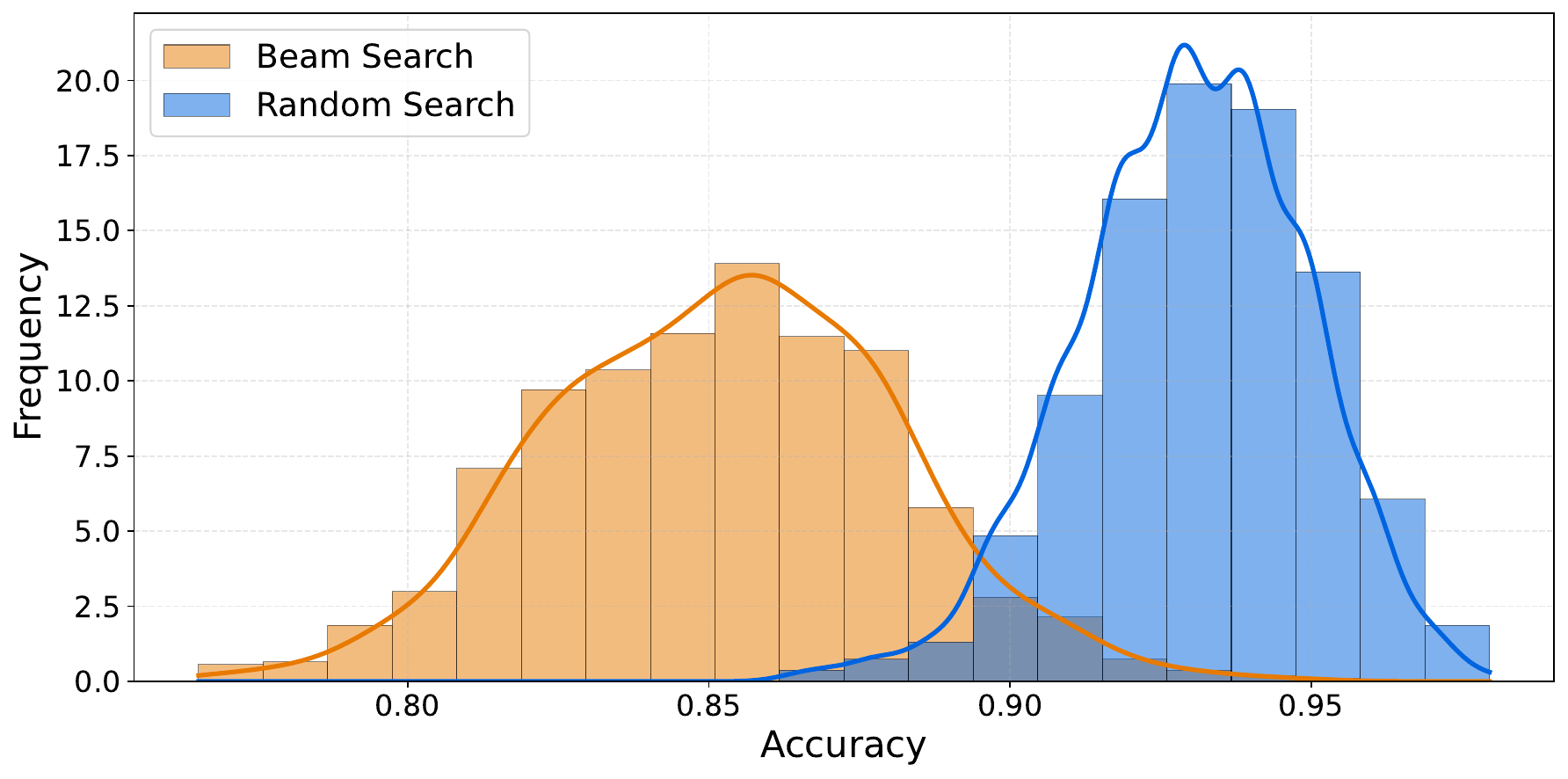}
    };
    \node[
            rotate=90,
            font=\fontsize{6pt}{7pt}\selectfont\sffamily,
            anchor=south
        ] at ([xshift=0.05cm]img.west) {Relative Frequency (\%)};
        \node[
            font=\fontsize{6pt}{7pt}\selectfont\sffamily,
            anchor=north,
            overlay
        ] at ([yshift=0.075cm]img.south) {Accuracy};
    \end{tikzpicture}
  \caption{\textbf{LPDS reveals a lower, less consistent performance profile.} Empirical accuracy distribution of Qwen-2.5-32B-Instruct on GSM-Symbolic.}
  \label{fig:performance_variation_gsms_qwen_32b}
  \vspace{-15pt}
\end{wrapfigure}
Figure~\ref{fig:performance_variation_gsms_qwen_32b} shows the variability of Qwen-2.5-32B-Instruct's performance on samples from GSM-Symbolic identified via beam search and random sampling. For each template, we randomly select one variation explored during search and compute the average accuracy across templates. Repeating this process 1{,}000 times yields an empirical accuracy distribution for each search strategy. Beam search produces a broader, left-shifted distribution relative to random sampling, indicating lower and less robust performance than random sampling would suggest. Similar trends for other datasets and models are shown in Figures~\ref{fig:performance_variance_finchain}--\ref{fig:performance_variance_engtrace} in the Appendix.

\begin{figure*}[b!]
  \centering
  \begin{subfigure}[t]{0.31\textwidth}
    \centering
    \includegraphics[width=\linewidth]{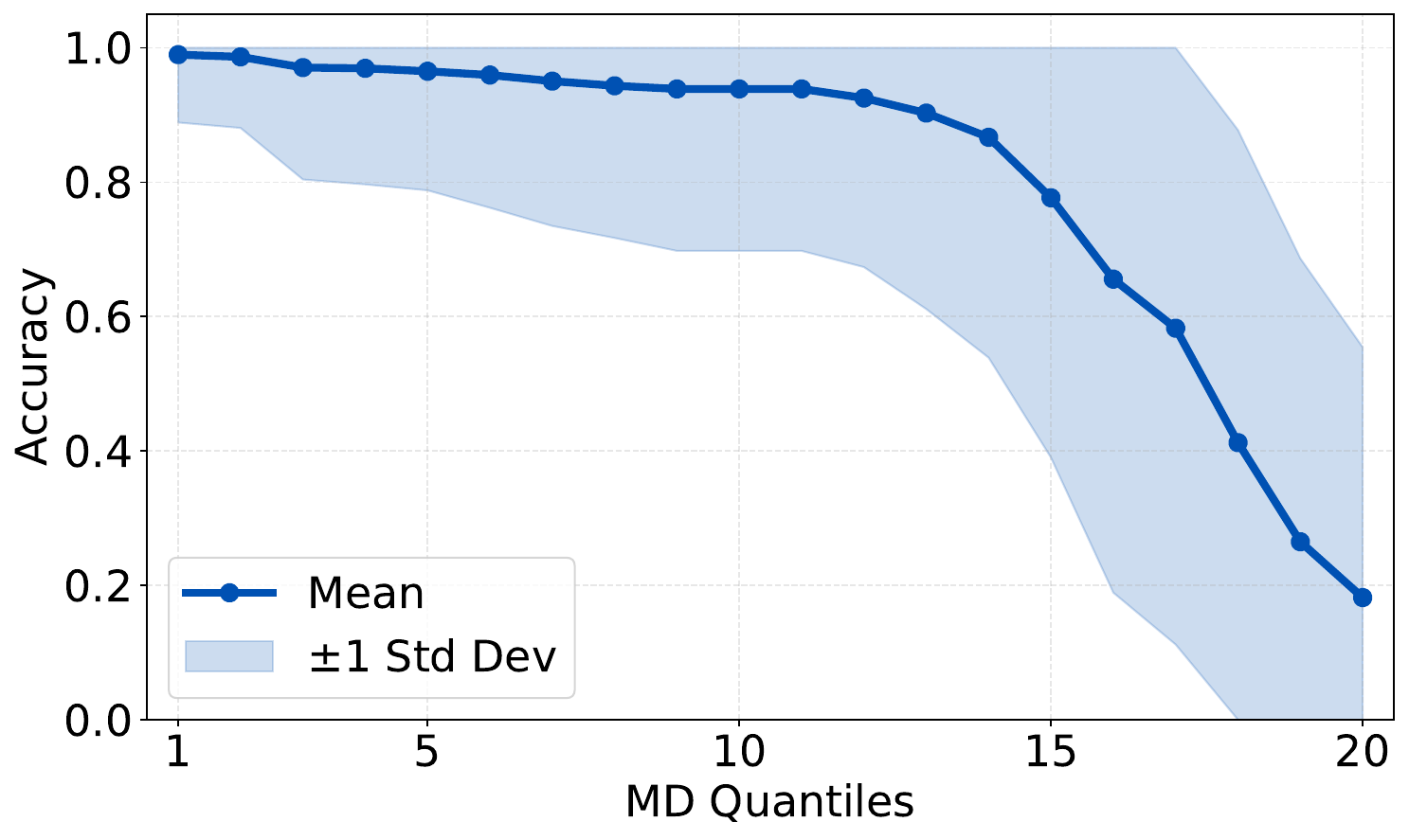}
    \caption{Qwen-2.5-7B-it}
    \label{fig:reliability_gsms_qwen_7b}
  \end{subfigure}
  \hspace{0.01\textwidth}
  \begin{subfigure}[t]{0.31\textwidth}
    \centering
    \includegraphics[width=\linewidth]{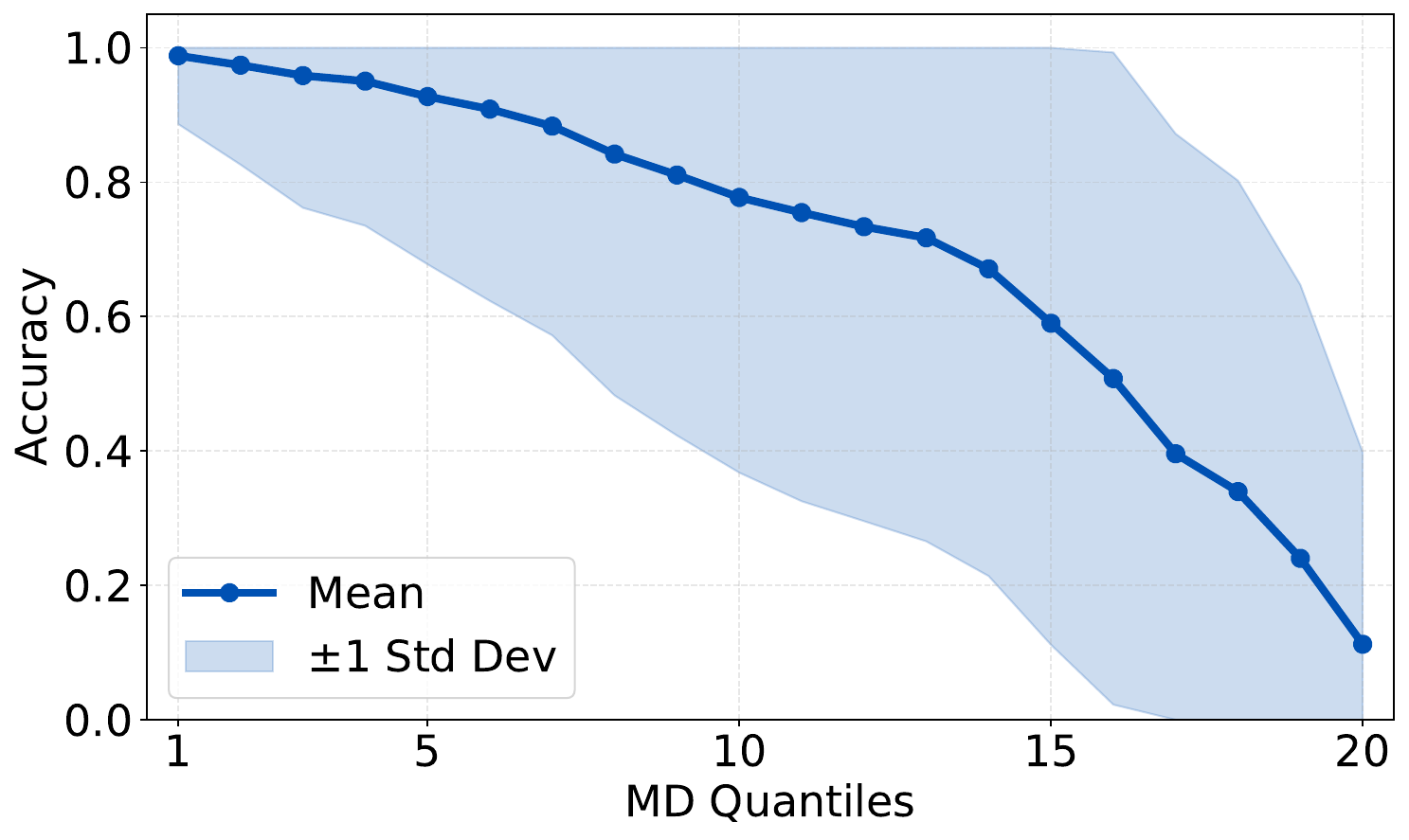}
    \caption{Llama-3.1-8B-it}
    \label{fig:reliability_gsms_llama_8b}
  \end{subfigure}
  \hspace{0.01\textwidth}
  \begin{subfigure}[t]{0.31\textwidth}
    \centering
    \includegraphics[width=\linewidth]{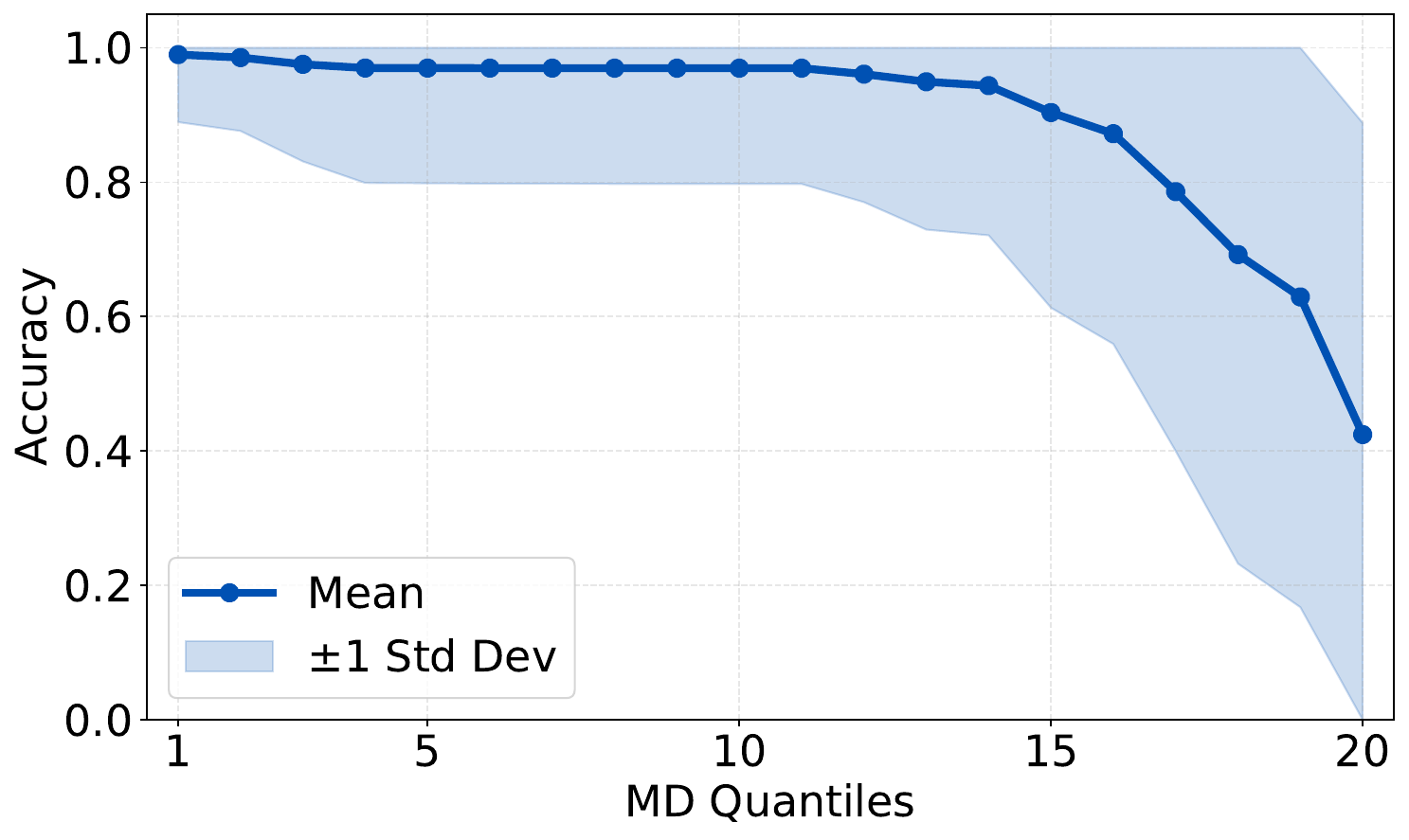}
    \caption{Llama-3.1-70B-it}
    \label{fig:reliability_gsms_llama_70b}
  \end{subfigure}
  \caption{\textbf{Accuracy decreases on subsets of increasing difficulty.} Variations are found via beam search; curves show means across GSM-Symbolic templates, with standard deviation. Corresponding results for other models and datasets, including FinChain and EngTrace, are shown in Figures~\ref{fig:reliability_plot_finchain}--\ref{fig:reliability_plot_engtrace} in the Appendix.}
  \label{fig:reliability_gsms}
\end{figure*}

\paragraph{\textbf{Brittleness becomes more pronounced with increasing difficulty.}} To study performance as difficulty scales, we group the problem variations identified via beam search into 20 quantile-based subsets of increasing difficulty for each template. We then compute accuracy for each subset and average it across templates. Figure~\ref{fig:reliability_gsms} shows the resulting curves for Qwen-2.5-7B-Instruct, Llama-3.1-8B-Instruct, and Llama-3.1-70B-Instruct. As difficulty $\mathrm{MD}_{\mathcal{H}}$ increases, accuracy generally declines. However, the models exhibit distinct robustness profiles: Qwen-2.5-7B-Instruct remains relatively stable through quantile 14, then drops noticeably. In comparison, Llama-3.1-8B-Instruct begins to degrade earlier, falling below 80\% at quantile 10, suggesting worse robustness to moderate increases in difficulty. Both models show pronounced drops in the hardest bins, reaching accuracies of 18.19\% and 11.25\% at quantile 20, respectively. Llama-3.1-70B-Instruct is the most robust, with accuracy remaining above 80\% up to quantile 16 and reaching 42.45\% at quantile 20. Similar trends for other models and datasets are presented in Appendix~\ref{app:evaluating_robustness}, specifically Figures~\ref{fig:reliability_plot_finchain}--\ref{fig:reliability_plot_engtrace}.

\section{Fine-tuning on difficult problem variations}
\label{sec:improving_robustness}
We further analyze how fine-tuning models on problem variations of different difficulty levels affects their robustness to logic-preserving variations.

\paragraph{\textbf{Training setup.}} We fine-tune Llama-3.1-8B-Instruct on problem variations from GSM-Symbolic that were explored during beam search (Section~\ref{sec:lpds}), which provides a large set of difficult samples. We split the data by template into training (70\%), validation (15\%), and test (15\%) sets. For validation and test, we subsample 10 variations per template, yielding up to 1{,}000 samples per split. To fairly compare the effect of training on easier and more difficult samples, we restrict the training set to variations the model initially answers incorrectly. From this pool, we randomly subsample up to 100 variations per template, yielding up to 10{,}000 training samples. We rank these samples by difficulty, $\mathrm{MD}_{\mathcal{H}}$, and divide them into three equally sized subsets of increasing difficulty level ($\mathrm{Q}_{\mathrm{low}}$, $\mathrm{Q}_{\mathrm{mid}}$, $\mathrm{Q}_{\mathrm{high}}$). We then train a separate model on each subset via supervised fine-tuning using ground-truth reasoning traces (Figure~\ref{fig:symb_template_examples}). Hyperparameters are tuned via Bayesian optimization on the validation split. Besides evaluating models on the full test set (GSM-S), we also filter the test set to samples that were originally answered incorrectly and split the resulting data into three equal-sized subsets of increasing difficulty (GSM-S$_{\mathrm{low}}$, GSM-S$_{\mathrm{mid}}$, GSM-S$_{\mathrm{high}}$). Details about the dataset statistics and training procedure are provided in Appendix~\ref{app:training_details}.

\begin{table*}[tbp]
    \centering
    \footnotesize
    \begin{tabular}{l|c|ccc}
        \toprule
        \textbf{Train set} & \textbf{GSM-S} & \textbf{GSM-S$_{\mathrm{\textbf{low}}}$} & \textbf{GSM-S$_{\mathrm{\textbf{mid}}}$} & \textbf{GSM-S$_{\mathrm{\textbf{high}}}$} \\ 
        \midrule
        -           & 69.29 & 0.00 & 0.00 & 0.00 \\
        \midrule
        $\mathrm{Q_{\mathrm{low}}}$ (100\%)  & 72.93 & \textbf{78.20} & 60.15 & 36.84 \\
        $\mathrm{Q_{\mathrm{mid}}}$ (100\%)  & 81.11 & 75.94 & \textbf{85.71} & 53.38 \\
        $\mathrm{Q_{\mathrm{high}}}$ (100\%)  & \textbf{83.03} & 73.68 & \textbf{85.71} & \textbf{85.71} \\
        \bottomrule
    \end{tabular}
    \caption{\textbf{Fine-tuning on difficult problem variations yields more consistent robustness gains than training on easier ones.}
    Accuracy after fine-tuning Llama-3.1-8B-Instruct on subsets of varying difficulty. Results are shown for problem variations from GSM-Symbolic. Corresponding results for FinChain are reported in Table~\ref{tab:full_sft_per_template_val_filter_finchain} in Appendix~\ref{app:improving_robustness}.}
    \label{tab:full_sft_gsms_llama_8b_main}
\end{table*}

\paragraph{\textbf{Training on difficult variations yields more consistent robustness gains.}} Table~\ref{tab:full_sft_gsms_llama_8b_main} shows performance after fine-tuning on different subsets. We make the following observations: first, fine-tuning on any subset ($\mathrm{Q}_{\mathrm{low}}$--$\mathrm{Q}_{\mathrm{high}}$) improves accuracy on the full test set (GSM-S) relative to the base model. Second, the gains vary: training on $\mathrm{Q}_{\mathrm{low}}$ increases accuracy to 72.93\%, whereas training on $\mathrm{Q}_{\mathrm{mid}}$ and $\mathrm{Q}_{\mathrm{high}}$ raise it to 81.11\% and 83.03\%, respectively. Third, performance on the difficulty-specific subsets helps explain this pattern. Training on $\mathrm{Q}_{\mathrm{low}}$ yields strong gains on GSM-S$_{\mathrm{low}}$ (78.20\%), but smaller improvements on GSM-S$_{\mathrm{mid}}$ (60.15\%) and GSM-S$_{\mathrm{high}}$ (36.84\%). In contrast, training on $\mathrm{Q}_{\mathrm{mid}}$ improves performance on both GSM-S$_{\mathrm{low}}$ (75.94\%) and GSM-S$_{\mathrm{mid}}$ (85.71\%), with more moderate gains on GSM-S$_{\mathrm{high}}$ (53.38\%). Training on $\mathrm{Q}_{\mathrm{high}}$ yields strong performance across all difficulty subsets, reaching 73.68\% on GSM-S$_{\mathrm{low}}$, 85.71\% on GSM-S$_{\mathrm{mid}}$, and 85.71\% on GSM-S$_{\mathrm{high}}$. These results suggest that training on more difficult problem variations leads to more consistent robustness gains across difficulty levels than training on easier ones. Evaluations of models trained on mixtures of $\mathrm{Q}_{\mathrm{low}}$--$\mathrm{Q}_{\mathrm{high}}$ data, as well as results on FinChain, are provided in Appendix~\ref{app:improving_robustness}.

\section{Conclusion}\label{sec:conclusion}
In this work, we introduced \emph{logic-preserving difficulty scaling} (LPDS), a systematic framework for identifying problem variations defined by a symbolic template that cause models to fail. Across datasets and models, we showed that LPDS efficiently finds difficult problem variations and captures model brittleness more reliably than random sampling. Overall, LPDS enables more accurate robustness estimates, identifies severe failure modes, and can help improve robustness via difficulty-aware fine-tuning.

\paragraph{\textbf{Limitations \& future directions.}} We applied LPDS to symbolic templates to find difficult variations that reveal model failures. While the algorithm in Section~\ref{sec:lpds} is general and can be extended to search spaces beyond symbolic templates, it remains an open question how our findings transfer to such settings. Extending LPDS to other types of variations is a promising direction for future work, for both robustness evaluation and training data selection.

\section*{Reproducibility statement}\label{app:reproducibility}
To facilitate reproducibility, we report details about the datasets, models, and training/evaluation procedures used in this study. Datasets are documented in Section~\ref{sec:preliminaries} and Appendix~\ref{app:datasets}. Model and prompt specifications appear in Section~\ref{subsec:eval_setup} (see Figures~\ref{fig:gsms_prompt}--\ref{fig:engtrace_prompt} for exact prompts), with further details in Appendix~\ref{app:models} and Appendix~\ref{app:prompts}. The evaluation protocol is described in Section~\ref{subsec:eval_setup}, with experiment-specific information in Sections~\ref{sec:quantifying_problem_variation_difficulty}--\ref{sec:improving_robustness} and Appendix~\ref{app:exp_details}. Training details\textemdash{}including dataset statistics, hyperparameters, and the training procedure\textemdash{}are reported in Section~\ref{sec:improving_robustness} and Appendix~\ref{app:training_details}.

%\section*{Author Contributions}
%If you'd like to, you may include  a section for author contributions as is done
%in many journals. This is optional and at the discretion of the authors.

% Uncomment for research review
\section*{Acknowledgments}
%Use unnumbered first level headings for the acknowledgments. All
%acknowledgments, including those to funding agencies, go at the end of the paper.
We express our gratitude to Maryam Fazel-Zarandi for her support throughout the project. We also thank Timon Willi, Evangelia Spiliopoulou, Mark Ibrahim, Daniil Cherniavskii, Angelo Ortiz, Sophia Houhamdi, and Lars Quaedvlieg for insightful discussions. Finally, we thank the anonymous reviewers for their valuable comments and suggestions.

%\clearpage
%\newpage
\bibliographystyle{assets/plainnat}
\bibliography{paper}

%\clearpage
%\newpage
\beginappendix

\section{Extended literature review}\label{sec:related_work}
\paragraph{\textbf{Evaluating robustness of LLM reasoning.}} Several studies assess the robustness of LLM reasoning by applying logic-preserving perturbations to inputs and examining whether responses remain correct when a problem’s surface form changes while its underlying solution remains the same~\citep{kumar2025robustness, zhang2025evaluating, mondorf2024beyond}. Common perturbations include non-semantic noise (e.g., spelling errors, random casing, extra whitespace, or punctuation)~\citep{gan-etal-2024-reasoning, wang-etal-2024-resilience, singh2024robustness}, reordering statements without altering logical dependencies~\citep{10.5555/3692070.3692325, zheng2024large, pezeshkpour-hruschka-2024-large}, inserting distractor or irrelevant sentences~\citep{pmlr-v202-shi23a, NEURIPS2024_dfaa29ed, mirzadeh2025gsmsymbolic}, semantic paraphrasing~\citep{10.1162/tacl_a_00681, sun2024evaluating, cao2024on}, and alternative prompt formatting~\citep{su2025single, sclar2024quantifying, he2024does}. Across benchmarks, these perturbations often reveal sensitivity to superficial cues and substantial performance variance despite problems having identical ground-truth solutions. Symbolic templates go beyond superficial perturbations by defining variables, constraints, and the complete reasoning procedure of a problem, while parameterizing numbers, names, item types, and other contextual details~\citep{mirzadeh2025gsmsymbolic, xie2025finchain, gull2025engtrace, srivastava2024functional, gulati2024putnamaxiom}. In this work, we use LPDS to identify those instantiations of a symbolic template that are most likely to induce model failure.

\paragraph{\textbf{Automatic prompt optimization.}} The goal of automatic prompt optimization (APO) is to adapt a model’s input prompt to optimize a metric of interest (e.g., task performance)~\citep{ramnath2025systematic}. APO methods can be categorized by their search space~\citep{li2025survey}. Continuous prompt optimization strategies optimize continuous variables, such as learnable embedding vectors that are either added to the prompt~\citep[often referred to as soft prompt tuning;][]{lester-etal-2021-power, wu-shi-2022-adversarial, li2025a} or projected back into the model’s discrete vocabulary~\citep{shi-etal-2023-toward}. Discrete approaches search directly over the model's vocabulary~\citep{li2025survey}, using gradient-based approximations~\citep{shin-etal-2020-autoprompt, zou2023universal, NEURIPS2024_608fe7e3} or combinatorial search~\citep{guo2024connecting, NEURIPS2024_b46bc144, hsieh-etal-2024-automatic}. For example,~\citet{guo2024connecting} propose EvoPrompt, an optimization procedure that evolves a population of prompts via iterative evolutionary operators.~\citet{NEURIPS2024_b46bc144} use fixed-budget best-arm identification to select promising prompt candidates.~\citet{hsieh-etal-2024-automatic} show that greedy search with beam search can outperform other purely greedy and genetic baselines. Similar to~\citet{hsieh-etal-2024-automatic}, LPDS is a discrete prompt optimization algorithm based on beam search (Section~\ref{subsec:scaling_difficulty}). However, unlike~\citet{hsieh-etal-2024-automatic}, LPDS searches over the space of problems defined by a symbolic template to identify variants that maximize difficulty.

\section{Datasets}\label{app:datasets}
As described in Section~\ref{sec:preliminaries}, we generate data via symbolic templates, which can yield many variations of a problem that differ in wording and values but share the same reasoning graph. Specifically, we consider symbolic templates from three datasets: GSM-Symbolic~\citep{mirzadeh2025gsmsymbolic}, Finchain~\citep{xie2025finchain}, and EngTrace~\citep{gull2025engtrace}. From each dataset, we select 100 symbolic templates, allowing us to generate logic-preserving variations of 100 different problems per dataset. Below is a formalization of how problem variations are generated from a symbolic template, followed by an overview of each dataset.

\paragraph{\textbf{Formalization.}} Let us consider a prompt $p_0 = (x_1, x_2, \ldots, x_L)$ consisting of $L$ tokens $x_i \in \mathcal{V}$, representing a reasoning problem of interest (e.g., sample $j$ from GSM8K~\citep{cobbe2021training}). A symbolic template for this problem can be used to generate logically equivalent variations of $p_0$ by replacing certain words or n-grams with semantically meaningful substitutions while preserving the underlying structure of the problem. Specifically, let $S$ denote the set of slot positions in the template, where each slot $s \in S$ corresponds to a span of one or more consecutive token positions in $[L] := \{1, \ldots, L\}$. For each slot $s \in S$, the corresponding variable $v_s$ (which may be a single word like ``shrimps'', a number, or a multi-word entity like ``hot chocolate'', as shown in Figure~\ref{fig:symb_template_examples}) is drawn from a predefined set of suitable substitutions $V_s \subset \mathcal{V}^*$, where $\mathcal{V}^*$ denotes the set of all possible sequences over vocabulary $\mathcal{V}$. Token positions not covered by any slot in $S$ remain fixed across all variations. We denote by $\mathcal{T}$ the set of all logically equivalent variations of prompt $p_0$ that can be generated from this template, where each $p_i \in \mathcal{T}$ represents one specific variation. More generally, we consider $m$ different templates $\{\mathcal{T}^{(1)}, \mathcal{T}^{(2)}, \ldots, \mathcal{T}^{(m)}\}$, one for each base problem in $\mathcal{D}_0 = \{p_0^{(1)}, p_0^{(2)}, \ldots, p_0^{(m)}\}$.\footnote{Note that a base prompt $p_0$ is not required to define a template. While GSM-Symbolic provides symbolic templates for problems from GSM8K~\citep{cobbe2021training}, FinChain and EngTrace do not derive their templates from a fixed base benchmark.}

\subsection{GSM-Symbolic}\label{app:gsms_details}
Introduced by~\citet{mirzadeh2025gsmsymbolic}, GSM-Symbolic is a collection of symbolic templates derived from grade school math word problems of GSM8K (Grade School Math 8K)~\citep{cobbe2021training}. Each template specifies variables, constraints, and a step-by-step solution procedure of a problem, while treating numbers, names, and other contextual details as parameters (see the left side of Figure~\ref{fig:symb_template_examples}). GSM-Symbolic contains 100 templates corresponding to 100 different GSM8K problems. While~\citet{mirzadeh2025gsmsymbolic} propose variants such as GSM-M1\textemdash{}where one sentence of the problem statement is removed\textemdash{}and GSM-Symbolic-P1/P2\textemdash{}where one or two irrelevant sentences are added\textemdash{}we use the vanilla GSM-Symbolic templates in this study.\footnote{Obtained from: \href{https://github.com/apple/ml-gsm-symbolic/tree/main/templates/symbolic}{https://github.com/apple/ml-gsm-symbolic/tree/main/templates/symbolic}.} To increase the range of possible variations, we manually extend the lists of possible variable values (e.g., names, objects, etc.) while keeping number ranges as proposed by~\citet{mirzadeh2025gsmsymbolic}. For further details about GSM-Symbolic, we refer to the original study by~\citet{mirzadeh2025gsmsymbolic}.

\subsection{FinChain}\label{app:finchain_details}
Recently introduced by~\citet{xie2025finchain}, FinChain provides symbolic templates for multi-step reasoning problems in the financial domain. It spans 58 topics across 12 financial domains, such as personal finance, risk management, and financial reporting. For each topic, FinChain presents five symbolic templates spanning three difficulty levels (two basic, two intermediate, and one advanced), where difficulty is defined by the number and complexity of the reasoning steps required to solve the underlying problem. Similar to GSM-Symbolic, each symbolic template specifies variables, constraints, and a step-by-step solution procedure, while treating numbers, names, and other contextual details as parameters (see the right side of Figure~\ref{fig:symb_template_examples}). Across topics and domains, we select 100 FinChain templates with varying difficulty levels\footnote{Obtained from: \href{https://github.com/mbzuai-nlp/finchain/tree/main/data/templates}{https://github.com/mbzuai-nlp/finchain/tree/main/data/templates}.} and verify that each template asks for a single final numeric answer. Because some templates define floating-point variables over very fine-grained numerical ranges, we further ensure that the space of possible problem variations does not grow excessively by adjusting the ranges accordingly (see right side of Figure~\ref{fig:symb_template_examples} for an example). For further details about FinChain, we refer to the original study by~\citet{xie2025finchain}.

\subsection{EngTrace}\label{app:engtrace_details}
EngTrace~\citep{gull2025engtrace} introduces symbolic templates for multi-step reasoning problems in engineering science (see Figure~\ref{fig:symb_template_example_engtrace} for an example). It spans three major branches: chemical, electrical, and mechanical engineering. For each branch, EngTrace provides templates across three core domains. For example, in chemical engineering, it presents templates covering thermodynamics, reaction kinetics, and transport phenomena. Similar to FinChain, the templates in EngTrace are characterized by difficulty: easy and intermediate templates are designed to test core engineering laws and procedural workflows, whereas advanced templates involve high conceptual and mathematical complexity. EngTrace introduces 90 symbolic templates, 30 for each engineering branch.\footnote{Obtained from: \href{https://github.com/usmansafdarktk/EngTrace}{https://github.com/usmansafdarktk/EngTrace}.} However, we noticed that some templates include variables that, depending on their values, substantially alter the solution structure. For instance, based on the value of a binary variable, a template might either ask for the maximum axial load of a rod given an allowable elongation or ask for the total elongation given an axial tensile load. To ensure that problem variations generated by a template have a consistent solution structure, we split such cases into separate templates, yielding more than 90 symbolic templates in total. We further ensure that each template asks for a single final numeric answer and that the space of possible problem variations remains tractable by limiting the ranges and granularity of allowed numerical values. From the resulting set, we select 100 symbolic templates. For further details about EngTrace, we refer to the original study by~\citet{gull2025engtrace}.

\section{Experimental details}\label{app:exp_details}

\subsection{Models}\label{app:models}
As described in Section~\ref{subsec:eval_setup}, we evaluate large language models from different model families and across various sizes. All models are instruction-tuned with parameter sizes ranging from 3 billion and 72 billion. Table~\ref{tab:model_comparison} summarizes important characteristics of each model.

\begin{table*}[tbp]
    \centering
    \small
    \begin{tabular}{lcccc}
        \toprule
        Model & Parameters & Layers & Hidden Dim & License \\
        \midrule
        {\small\href{https://huggingface.co/meta-llama/Llama-3.2-3B-Instruct}{Llama-3.2-3B-Instruct}} & 3.21B & 28 & $3{,}072$ & \href{https://huggingface.co/meta-llama/Llama-3.2-3B-Instruct/blob/main/LICENSE.txt}{{\small Llama 3.2}}\\
        {\small\href{https://huggingface.co/meta-llama/Llama-3.1-8B-Instruct}{Llama-3.1-8B-Instruct}} & 8.03B & 32 & $4{,}096$ & \href{https://huggingface.co/meta-llama/Llama-3.1-70B-Instruct/blob/main/LICENSE}{{\small Llama 3.1}}\\
        {\small\href{https://huggingface.co/meta-llama/Llama-3.1-70B-Instruct}{Llama-3.1-70B-Instruct}} & 70.60B & 80 & $8{,}192$ & \href{https://huggingface.co/meta-llama/Llama-3.1-70B-Instruct/blob/main/LICENSE}{{\small Llama 3.1}}\\
        \cmidrule{1-5}
        {\small\href{https://huggingface.co/Qwen/Qwen2.5-7B-Instruct}{Qwen-2.5-7B-Instruct}} & 7.61B & 28 & $3{,}584$ & \href{https://huggingface.co/Qwen/Qwen2.5-7B-Instruct/blob/main/LICENSE}{{\small Apache 2.0}}\\
        {\small\href{https://huggingface.co/Qwen/Qwen2.5-32B-Instruct}{Qwen-2.5-32B-Instruct}} & 32.50B & 64 & $5{,}120$ & \href{https://huggingface.co/Qwen/Qwen2.5-32B-Instruct/blob/main/LICENSE}{{\small Apache 2.0}}\\
        {\small\href{https://huggingface.co/Qwen/Qwen2.5-72B-Instruct}{Qwen-2.5-72B-Instruct}} & 72.70B & 80 & $8{,}192$ & \href{https://huggingface.co/Qwen/Qwen2.5-72B-Instruct/blob/main/LICENSE}{{\small Qwen}}\\
        \cmidrule{1-5}
        {\small\href{https://huggingface.co/microsoft/phi-4}{Phi-4}} & 14.00B & 40 & $5{,}120$ & \href{https://huggingface.co/microsoft/phi-4/blob/main/LICENSE}{{\small MIT}}\\
        \bottomrule
    \end{tabular}
    \caption{\textbf{Properties of the models studied in this work.} Provided are details on the number of parameters, layers, and the hidden dimension size of the residual stream. Model weights are obtained from their respective Hugging Face repositories.}
    \label{tab:model_comparison}
\end{table*}

\subsection{Prompts}\label{app:prompts}
We prompt models via chain-of-thought (CoT) prompting~\citep{NEURIPS2022_9d560961}, following the evaluation setup of~\citet{mirzadeh2025gsmsymbolic}, as outlined in Section~\ref{subsec:eval_setup}. The exact prompts for samples from GSM-Symbolic, FinChain, and EngTrace are provided in Figures~\ref{fig:gsms_prompt},~\ref{fig:finchain_prompt}, and~\ref{fig:engtrace_prompt}, respectively. Within each prompt, we present five CoT examples to guide the models' reasoning and response format. For GSM-Symbolic, we take the same first five CoT examples as used by~\citet{mirzadeh2025gsmsymbolic}, which are also used in the common evaluation setup on GSM8K.\footnote{These examples are also used in Eleuther AI's LM evaluation harness: \href{https://github.com/EleutherAI/lm-evaluation-harness/blob/main/lm_eval/tasks/gsm8k/gsm8k-cot.yaml}{https://github.com/EleutherAI/lm-evaluation-harness/blob/main/lm\_eval/tasks/gsm8k/gsm8k-cot.yaml}.} For FinChain and EngTrace, we provide five CoT examples by instantiating a problem from each of five different symbolic templates per dataset, which we then exclude from the evaluation. All prompts further ask the model to reason step-by-step and to provide a final numeric answer after the \#\#\#\# prefix, following the standard format of GSM8K~\citep{cobbe2021training}.

\subsection{Assessing variation difficulty}\label{app:difficulty_assessment}
As outlined in Section~\ref{sec:quantifying_problem_variation_difficulty}, we compare different metrics to assess the difficulty of a problem variation $p_i \in \mathcal{T}$ for a given model $\mathcal{M}$. Below, we formally define all metrics and describe how we evaluate their predictive power for answer correctness by computing the micro-averaged AUC score and odds ratio, as presented in Section~\ref{subsec:pred_answer_correctness}.

\subsubsection{Formal definitions}\label{app:formal_definitions}
Consider a language model $\mathcal{M}$ and a token sequence $s = (s_1, s_2, \ldots, s_L)$ of length $L$. The sequence $s$ may represent a problem variation of interest $p_i \in \mathcal{T}$ (see the ``input'' rows in Table~\ref{tab:correlation_llama_3.1_8B}) or the model's corresponding response $Y_i$ (see the ``output'' rows in Table~\ref{tab:correlation_llama_3.1_8B}). Based on this sequence, we compute a range of metrics intended to capture variation difficulty.

\paragraph{\textbf{Perplexity.}} Formally, the perplexity of a model over the token sequence $s$ is defined as the exponentiated average negative log-likelihood:

\begin{equation}
\mathrm{PPL}_{\mathcal{M}}(s) =
\exp\!\left(-\frac{1}{L}\sum_{t=1}^{L}\log P_{\mathcal{M}}(s_t \mid s_{<t})\right)
\end{equation}

where $\log P_{\mathcal{M}}(s_t \mid s_{<t})$ is the log-likelihood of token $s_t$ conditioned on the preceding tokens $s_{<t}$. Intuitively, perplexity can be understood as a measure of average uncertainty or ``surprise'' the model assigns to the observed token sequence $s$, where higher values indicate greater surprisal~\citep{huyen2019evaluation-metrics}. Thus, it is commonly used to assess the model's uncertainty over some token sequence $s$~\citep{gonen-etal-2023-demystifying, mora-cross-calderon-ramirez-2024-uncertainty}.

\paragraph{\textbf{Entropy}.} Another metric that is commonly used to quantify the uncertainty of a language model with respect to a given token sequence is entropy~\citep{huang2024survey}. Formally, the model's entropy over the token sequence $s$ is defined as:

\begin{equation}
\mathrm{H}_{\mathcal{M}}(s) =  -\frac{1}{L}\sum_{t=1}^{L}\sum_{v \in \mathcal{V}} P_{\mathcal{M}}(v \mid s_{<t}) \log P_{\mathcal{M}}(v \mid s_{<t})
\end{equation}

where $\mathcal{V}$ denotes the model's vocabulary and $P_{\mathcal{M}}(v \mid s_{<t})$ is the model's predictive probability of token $v$ at position $t$ conditioned on the preceding tokens $s_{<t}$. Intuitively, the entropy of a model over the sequence $s$ measures the average uncertainty of its next-token predictive distribution along the sequence, where higher values indicate less confident predictions. In contrast to perplexity\textemdash{}which measures how much probability the model assigned to the actually observed next tokens along $s$\textemdash{}entropy measures how spread out the model's full next-token predictive distributions are along $s$.

\paragraph{\textbf{Self-certainty}.} To quantify the quality of a model's response,~\citet{kang2025scalable} introduce self-certainty, a confidence metric that measures how concentrated the model's next-token distribution is on average across a token sequence. Let $\mathcal{V}^{(k)}_t \subseteq \mathcal{V}$ denote the set of the $k$ highest-probability tokens under $P_{\mathcal{M}}(\cdot \mid s_{<t})$. Formally, self-certainty over a token sequence $s$ is defined as:

\begin{equation}
\text{Self-Certainty}_{\mathcal{M}}(s)= -\frac{1}{L\cdot k} \sum_{t=1}^{L} \sum_{v \in \mathcal{V}^{(k)}_t} \log \bigl(k \cdot P_{\mathcal{M}} (v \mid s_{<t}) \bigr).
\end{equation}

Intuitively, larger values indicate that the probability mass is more concentrated on a small set of tokens (i.e., higher confidence), whereas smaller values correspond to more diffuse next-token distributions and thus greater uncertainty.~\citet{kang2025scalable} find that self-certainty outperforms other metrics such as perplexity and entropy in distinguishing correctly answered from incorrectly answered model responses. In this study, we compute self-certainty using $k = 50$.

\paragraph{\textbf{Levenshtein distance}.} As described in Section~\ref{subsec:ref_based_distance}, the Levenshtein distance measures the minimum number of single-token insertions, deletions, and substitutions required to transform one string into the other. Let $s = Y_i$ denote the model response for problem variation $p_i$ from a symbolic template, and let $\mathcal{Y}=\{Y_k\}_{k=1}^N$ be a reference set of $N$ correct responses to problem variations $p_k \neq p_i$ from the same template. To measure the distance from $Y_i$ to the reference set $\mathcal{Y}$, we compare $Y_i$ to each $Y_k \in \mathcal{Y}$. Specifically, we first pre-process each string\footnote{To handle (mathematical) reasoning traces, we normalize Unicode (NFKC), case-folds, and canonicalize common typographic/math variants (e.g., $\times\mapsto *$, $\div\mapsto /$).} and apply a math-aware tokenizer $\tau(\cdot)$ that extracts tokens corresponding to words, numbers, and operators. Let $u=\tau(Y_i)$ and $v=\tau(Y_k)$ denote the resulting token sequences. We compute the minimum (normalized) Levenshtein distance as:

\begin{equation}\label{eq:LD_min}
\text{LD}_{\min}\!\left(Y_i,\mathcal{Y}\right) = \min_{Y_k\in\mathcal{Y}} \tilde d_L\!\left(Y_i,Y_k\right), \qquad \tilde d_L\!\left(Y_i,Y_k\right) = \frac{d_L(u,v)}{\max\{|u|,|v|\}}
\end{equation}

where we aggregate over the reference set using minimum pooling and $d_L(u,v)$ is the minimum number of single-token insertions, deletions, and substitutions required to transform $u$ into $v$. Intuitively, $\text{LD}_{\min}$ measures the closest lexical match between the model's response and the reference set of correct reasoning traces for the given symbolic template. Instead of the minimum distance, it is possible to apply other pooling operators. For instance, we can compute the mean Levenshtein distance as:

\begin{equation}
    \text{LD}_{\text{mean}}(Y_i, \mathcal{Y}) = \frac{1}{|\mathcal{Y}|} \sum_{k=1}^{|\mathcal{Y}|} \tilde{d}_L(Y_i, Y_k)
\end{equation}

Similarly, we can compute the the maximum Levenshtein distance $\text{LD}_{\max}$ or the median Levenshtein distance $\text{LD}_{\text{median}}$ over the reference set.

\paragraph{\textbf{Input-based Mahalanobis distance.}} A key aspect of the algorithm presented in Section~\ref{subsec:scaling_difficulty} is its two-stage scoring procedure. As described in Section~\ref{subsec:scaling_difficulty}, we use a cheap approximation of problem difficulty, denoted by $\tilde{f}$, to pre-filter candidates generated during Step~1. Although this approximation is less predictive than the exact score, it is much faster to compute. Concretely, for a variation $p_i=(x_1, x_2, \ldots, x_L)$, we compute the average input embedding $\mathbf{E}_i=\frac{1}{L}\sum_{t=1}^{L}\ve_t$, where $\ve_t=\mW_{\mathrm{emb}}[x_t]\in\mathbb{R}^d$ and $\mW_{\mathrm{emb}}$ denotes the model's input embedding matrix. We then compare $\mathbf{E}_i$ to a reference set $\mathcal{E}=\{\mathbf{E}_k\}_{k=1}^{N}$ constructed from variations $p_k \neq p_i$ that the model answered correctly. Specifically, we compute the Mahalanobis distance from $\mathbf{E}_i$ to a Gaussian distribution $\mathcal{N}(\mu_{\mathcal{E}}, \Sigma_{\mathcal{E}})$ fitted to $\mathcal{E}$, analogous to Equation~\ref{eq:MD_response}:

\begin{equation}\label{eq:MD_input}
    \text{MD}_{\mathcal{E}} = \text{MD}(\mathbf{E}_i; \mu_{\mathcal{E}}, \Sigma_{\mathcal{E}}) = (\mathbf{E}_i - \mu_{\mathcal{E}})^T \Sigma^{-1}_{\mathcal{E}} (\mathbf{E}_i - \mu_{\mathcal{E}})
\end{equation}

\subsubsection{Computing reference-based metrics}\label{app:reference_based_metrics}
As described in Section~\ref{subsec:ref_based_distance}, we represent a template’s underlying reasoning graph using a reference set $\mathcal{Y} = \{Y_k\}_{k=1}^N$ of model responses to correctly answered problem variations $p_k$. In all experiments, we set $N = 200$ and exclude the corresponding problem variations $p_k$ from subsequent analyses. For templates where the model answers fewer than 200 variations correctly, we use all available samples. If the model answers none of the problem variations correctly for a given template, an alternative reference set can be used\textemdash{}for example, one derived from a different reference model or from ground-truth solution traces\textemdash{}as discussed in Appendix~\ref{app:assessing_variation_difficulty}.

When computing the Mahalanobis distance $\text{MD}_{\mathcal{H}}$ between the hidden representation $\mathbf{H}_i^{(l)}$ of the model's response $Y_i$ and the Gaussian distribution $\mathcal{N}(\mu_{\mathcal{H}}, \Sigma_{\mathcal{H}})$, the layer $l$ at which the average hidden state $\mathbf{H}_i^{(l)}$ is computed must be specified. Based on initial experiments, we select $l$ to be roughly two-thirds of the way through the model. Specifically, this means $l = 19$ for Llama-3.2-3B-Instruct, $l = 21$ for Llama-3.1-8B-Instruct, $l = 52$ for Llama-3.1-70B-Instruct, $l = 19$ for Qwen-2.5-7B-Instruct, $l = 43$ for Qwen-2.5-32B-Instruct, $l = 53$ for Qwen-2.5-72B-Instruct, and $l = 27$ for Phi-4 (cf. Table~\ref{tab:model_comparison}).

\subsubsection{Measuring predictive power}\label{app:measuring_predictive_power}
As described in Section~\ref{subsec:pred_answer_correctness}, we quantify the predictive power of a metric $f$ for answer correctness ($c = 1$ for correct, $c = 0$ for incorrect) using its AUC score and the change in the odds of a correct answer associated with a one-standard-deviation increase in the z-scored version of $f$.

\paragraph{\textbf{AUC score.}} For each template, we compute an AUC score from its $(f_i, c_i)$ pairs. For binary outcomes $c_i$, the AUC equals the probability that a randomly chosen incorrectly answered variation receives a higher difficulty score $f_i$ than a randomly chosen correctly answered variation. An AUC of $0.5$ corresponds to random prediction, whereas values closer to $0$ or $1$ indicate stronger separation. We first compute AUC scores separately for each template to account for template-specific differences such as overall difficulty and error rates. We then aggregate the per-template scores into a micro-averaged AUC by weighting each template $j$ by its number of incorrect--correct pairs, $n_{\text{pairs}}^{(j)} = n_{0}^{(j)} n_{1}^{(j)}$:

\begin{equation}
\label{eq:auc_score}
\begin{aligned}
\mathrm{AUC}_{\text{micro}}
&=
\frac{\sum_j n_{\text{pairs}}^{(j)}\,\mathrm{AUC}^{(j)}}
     {\sum_j n_{\text{pairs}}^{(j)}}, \\
\mathrm{AUC}^{(j)}
&=
\frac{1}{n_{0}^{(j)}\,n_{1}^{(j)}}
\sum_{i:\,c_i^{(j)}=0}^{n_{0}^{(j)}}
\sum_{k:\,c_k^{(j)}=1}^{n_{1}^{(j)}}
\Bigl[
\mathds{1}\!\bigl(f_i^{(j)} > f_k^{(j)}\bigr)
+ \tfrac{1}{2}\,\mathds{1}\!\bigl(f_i^{(j)} = f_k^{(j)}\bigr)
\Bigr].
\end{aligned}
\end{equation}

where $n_{0}^{(j)}$ and $n_{1}^{(j)}$ are the numbers of incorrectly and correctly answered examples for template $j$, respectively; $f_i^{(j)}$ and $f_k^{(j)}$ are the corresponding difficulty scores; and $\mathds{1}(\cdot)$ is the indicator function. Because each per-template AUC score $\mathrm{AUC}^{(j)}$ is normalized by the number of incorrect--correct pairs, $n_{\text{pairs}}^{(j)} = n_0^{(j)} n_1^{(j)}$, we exclude any templates with $n_{\text{pairs}}^{(j)} = 0$ from the micro-averaged AUC; i.e., templates whose problem variations were all answered either correctly or incorrectly by the model.

\begin{algorithm}[t!]
\caption{Scaling problem difficulty via beam search.}
\label{alg:beam_search}
\small
\begin{algorithmic}[1]
\REQUIRE variation $p_0$, beam size $b$, width $w$, exploration ratio $\rho_{\mathrm{expl}}$,
selection ratio $\rho_{\mathrm{sel}}$, iterations $T$
\ENSURE optimized variation $p^*$ with score $f^*$

\STATE Initialize beam: $\mathcal{B}_0 \gets \{(p_0, f(p_0))\}$
\STATE $p^* \gets p_0$, $f^* \gets f(p_0)$

\FOR{$t=1$ to $T$}
    \STATE $\mathcal{C} \gets \emptyset$
    \FORALL{$(p,f(p)) \in \mathcal{B}_{t-1}$}
        \STATE $\mathcal{P} \gets \emptyset$
        \STATE \textbf{Step 1: Generate candidates}
        \FORALL{slot $s_i \in \{s_1,\dots,s_K\}$}
            \FORALL{$v_{s_i} \in V_{s_i}$}
                \STATE $p' \gets p$ with $s_i$ replaced by $v_{s_i}$
                \IF{$p'$ satisfies constraints and $p' \notin \mathcal{B}_{t-1}$}
                    \STATE $\mathcal{P} \gets \mathcal{P} \cup \{p'\}$
                \ENDIF
            \ENDFOR
        \ENDFOR
        \STATE $\mathcal{P} \gets \mathcal{P} \cup \{p_i \sim \mathcal{T}\}_{i=1}^{\rho_{\mathrm{expl}}\cdot \lvert\mathcal{P}\rvert}$ (random sampling)

        \STATE \textbf{Step 2: Cheap scoring}
        \FORALL{$p' \in \mathcal{P}$}
            \STATE compute $\tilde f(p')$
        \ENDFOR
        \STATE \textbf{Step 3: Local pruning}
        \STATE $\mathcal{P}_b \gets$ top-$((1-\rho_{\mathrm{sel}}) \cdot b)$ variations from $\mathcal{P}$ by $\tilde f$
        \STATE $\mathcal{P}_r \gets \rho_{\mathrm{sel}} \cdot b$ variations sampled randomly from $\mathcal{P} \setminus \mathcal{P}_b$
        \STATE $\mathcal{C} \gets \mathcal{C} \cup \mathcal{P}_b \cup \mathcal{P}_r$
    \ENDFOR

    \STATE \textbf{Step 4: Exact scoring}
    \FORALL{$p' \in \mathcal{C}$}
        \STATE compute $f(p')$
    \ENDFOR
    \STATE \textbf{Step 5: Beam update}
    \STATE $\mathcal{B}_{t-1} \gets \mathcal{B}_{t-1} \cup \{(p',f(p')) : p' \in \mathcal{C}\}$
    \STATE $\mathcal{B}_t \gets$ top-$w$ variations from $\mathcal{B}_{t-1}$ by $f$

    \STATE \textbf{Step 6: Global best}
    \STATE $(p_{\mathrm{best}}, f_{\mathrm{best}}) \gets \arg\max_{(p,f)\in \mathcal{B}_t} f$
    \IF{$f_{\mathrm{best}} > f^*$}
        \STATE $p^* \gets p_{\mathrm{best}}$, $f^* \gets f_{\mathrm{best}}$
    \ENDIF
\ENDFOR

\STATE \RETURN$p^*, f^*$
\end{algorithmic}
\end{algorithm}

\paragraph{\textbf{Odds ratio.}} In addition to computing the micro-averaged AUC, we estimate how the probability of a correct answer varies with $f$ by fitting a logistic generalized linear mixed model. Concretely, let $c_i^{(j)} \in \{0,1\}$ denote correctness for problem variation $p_i^{(j)}$ from template $j$. We model the conditional probability of correctness as:

\begin{equation}
\text{logit}\!\left(\Pr\!\left(c_i^{(j)}=1 \mid u^{(j)}\right) \right) = \beta_0 + \beta_1\, z\!\left(f_i^{(j)}\right) + u^{(j)},
\end{equation}

where $z(f_i^{(j)})$ is the $z$-scored metric value, and $u^{(j)}$ is a template-specific random intercept that captures differences in overall problem difficulty across templates. Parameters are estimated via variational Bayes. We report the per-standard-deviation odds ratio as \(\exp(\beta_1)\), together with an approximate 95\% confidence interval $\exp(\beta_1 \pm 1.96\,\text{SE}(\beta_1))$. To avoid near-degenerate templates (i.e., quasi-separation), we exclude any highly imbalanced template for which more than 99\% of problem variations (more than 990 out of 1{,}000) were answered either correctly or incorrectly. An odds ratio of 1 indicates that answer correctness is invariant to $f$, whereas values farther from 1 indicate stronger associations. In particular, values below 1 indicate that higher values of $f$ reduce the odds of a correct answer, whereas values above 1 indicate the opposite. Because the z-scored difficulty metric $f$ is standardized, the per-SD odds ratios are directly comparable across models and datasets.

\subsection{Identifying problem variations via beam search}\label{app:beam_search}
As described in Section~\ref{subsec:scaling_difficulty}, we navigate the space of possible problem variations $p_i \in \mathcal{T}$ via beam search to identify those variations that maximize difficulty $f$. As illustrated in Algorithm~\ref{alg:beam_search}, we start from an initial problem variation $p_0 \in \mathcal{T}$, compute its difficulty score $f(p_0)$, and initialize the beam as $\mathcal{B} = {(p_0, f(p_0))}$. We then run beam search for $T$ iterations with branching factor $b$ and beam width $w$. At each iteration $t$, each active node (problem variation) $(p, \cdot) \in \mathcal{B}_{t}$ in the current beam is expanded into $b$ candidate variations. To choose which $b$ problem variations to consider from $\mathcal{T}$, we explore the effect of variable substitutions one slot at a time, assess the difficulty of the resulting variations via a cheap approximation $\tilde{f}$, and select the variations with the highest estimates $\tilde{f}$. To encourage exploration, we additionally consider non-neighboring variations by sampling a small fraction $\rho_{\text{expl}}$ of the final candidate variations randomly from the template $\mathcal{T}$. Furthermore, to mitigate potential ranking errors from the approximate metric $\tilde{f}$, we choose a small fraction $\rho_{\text{sel}}$ of the candidate set uniformly at random from the set of potential candidates. After each node has been expanded into $b$ candidate variations, we form the updated beam $\mathcal{B}_{t+1}$ from the top-$w$ candidates with the highest difficulty estimates $f$. As the search progresses, the difficulty of the problem variations in the current beam increases while the model's corresponding accuracy decreases, as shown in Figures~\ref{fig:beam_search_gsms_llama_8b},~\ref{fig:beam_search_gsms_qwen_7b},~\ref{fig:beam_search_gsms_qwen_32b}, and~\ref{fig:beam_search_finchain_llama_70b}.

To balance computational cost with sufficient exploration of the search space, we run beam search for $T = 15$ iterations with beam width $w = 16$ and branching factor $b = 16$. Furthermore, Table~\ref{tab:rho_expl} analyzes the impact of $\rho_{\text{expl}}$ on the models' error rates when evaluated on variations from GSM-Symbolic explored during search. We choose $\rho_{\text{expl}} = 0.2$ as it yields the highest error rates across models. For $\rho_{\text{sel}}$, we select a moderate value of $0.4$.

\begin{wraptable}{r}{0.48\textwidth}
\centering
\small
\vspace{-10pt}
\begin{tabular}{
  l|ccc
}
    \toprule
    \multicolumn{1}{c}{\multirow{2}{*}{\textbf{Model}}}
    & \multicolumn{3}{c}{$\bm{\rho_{\text{expl}}}$} \\
    \cmidrule(lr){2-4}
    \multicolumn{1}{c}{} & $0.2$ & $0.4$ & $0.6$ \\
    \toprule
    Llama-3.2-3B-Instruct & 44.18 & 43.87 & 42.99 \\
    Llama-3.1-8B-Instruct & 28.36 & 27.26 & 26.09 \\
    Qwen-2.5-7B-Instruct & 19.04 & 17.43 & 15.52 \\
    Qwen-2.5-32B-Instruct & 14.45 & 12.48 & 10.57 \\
    \bottomrule
\end{tabular}
\caption{\textbf{Impact of $\rho_{\text{expl}}$ on the models' error rates (\%).} Evaluations on problem variations from GSM-Symbolic explored during beam search. Other parameters remain fixed: $T = 15$, $w = 16$, $b = 16$, $\rho_{\text{sel}} = 0.4$.}
\label{tab:rho_expl}
\vspace{-10pt}
\end{wraptable}

\paragraph{\textbf{Moving goalpost.}} As the search progresses, an increasing number of problem variations are explored and assigned a difficulty estimate $f = \text{MD}_{\mathcal{H}}$. We can use the variations answered correctly by the model during the search procedure to update the reference distributions $\mathcal{N}(\mu_{\mathcal{H}}, \Sigma_{\mathcal{H}})$ (Equation~\ref{eq:MD_response}) and $\mathcal{N}(\mu_{\mathcal{E}}, \Sigma_{\mathcal{E}})$ (Equation~\ref{eq:MD_input}), respectively. This yields a better estimate of the template's underlying reasoning graph, as described in Section~\ref{subsec:ref_based_distance}. We update both reference distributions after every 50 newly correctly answered problem variations.

\subsection{Training on problem variations}\label{app:training_details}
As outlined in Section~\ref{sec:improving_robustness}, we fine-tune Llama-3.1-8B-Instruct on problems identified via beam search (Section~\ref{sec:lpds}) to improve the model's robustness to logic-preserving variations. In this Section, we report statistics for the training, validation and test sets and provide additional details about the training procedure and hyperparameters.

\begin{figure*}[tbp]
\centering
\includegraphics[width=\linewidth]{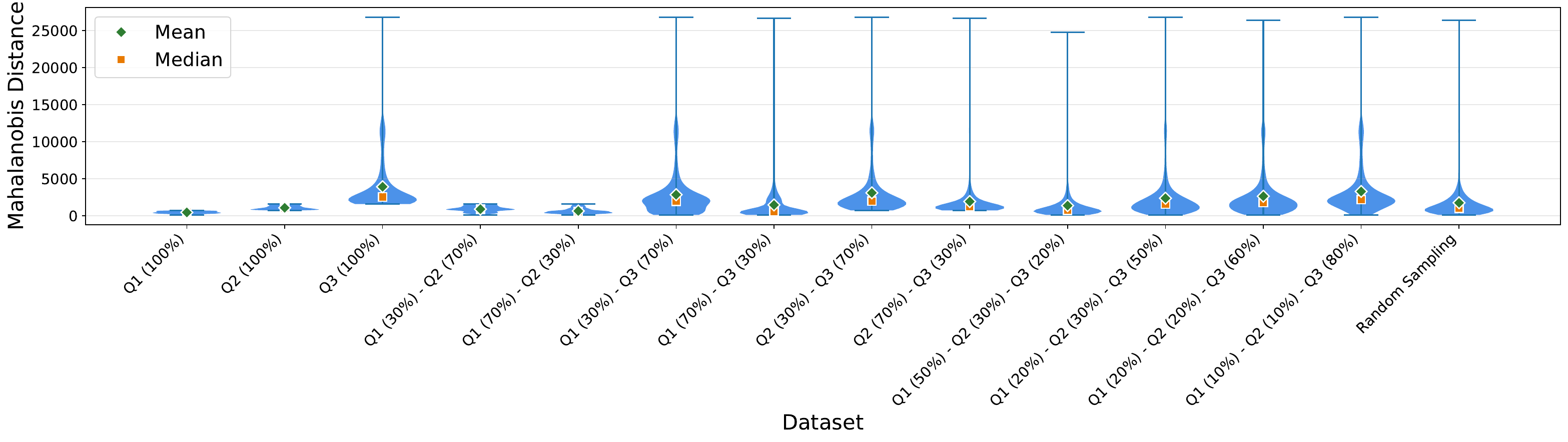}
\caption{\textbf{Distribution of difficulty scores.} Violin plots illustrating the distribution of difficulty scores $\text{MD}_{\mathcal{H}}$ per training data subset for samples from GSM-Symbolic.}
\label{fig:violin_difficulty_scores}
\end{figure*}

\subsubsection{Dataset statistics}\label{app:dataset_stats}
As described in Section~\ref{sec:improving_robustness}, we evaluate whether fine-tuning on difficult problem variations improves robustness more than training on easier ones. To separate robustness from basic task competence, we discard any template for which the initial model answers \emph{all} variations incorrectly. For these templates, the baseline accuracy is $0\%$, so robustness to logic-preserving variations is not well-defined; any improvement would mainly reflect learning the underlying task rather than increased invariance to its variations. Such cases are rare: for GSM-Symbolic, there is only one such template, and for FinChain, there are 10 templates for which the model’s baseline accuracy is $0\%$. We next split the data by template into training (70\%), validation (15\%), and test (15\%) sets. To fairly compare training on difficult versus easy variations, we restrict the training pool to variations that the model initially answers incorrectly. From this filtered set, we randomly subsample up to 100 variations per template. For templates with fewer than 100 initially incorrect variations, we include all available variations, yielding up to (but not necessarily) 100 variations per template. Overall, the resulting training set contains 8{,}694 samples for GSM-Symbolic and 7{,}356 samples for FinChain. Next, we rank training samples by the difficulty estimate $\text{MD}_{\mathcal{H}}$ and partition them into three equally sized subsets of increasing difficulty ($\mathrm{Q_{\mathrm{low}}}$, $\mathrm{Q_{\mathrm{mid}}}$, $\mathrm{Q_{\mathrm{high}}}$). Each subset contains 2{,}898 samples for GSM-Symbolic and 2{,}452 samples for FinChain. An overview of the distribution of difficulty scores $\text{MD}_{\mathcal{H}}$ for each training subset for samples from GSM-Symbolic is provided in Figure~\ref{fig:violin_difficulty_scores}. For validation and testing, we subsample 10 variations per template, resulting in 990 variations (from 99 templates) per evaluation set for GSM-Symbolic and 900 variations (from 90 templates) per set for FinChain. Finally, we further split the test set into initially incorrectly answered subsets of increasing difficulty (GSM-S$_{\mathrm{low}}$/FinChain$_{\mathrm{low}}$, GSM-S$_{\mathrm{mid}}$/FinChain$_{\mathrm{mid}}$, GSM-S$_{\mathrm{high}}$/FinChain$_{\mathrm{high}}$). Each subset contains 133 samples for GSM-Symbolic and 162 samples for FinChain.

\subsubsection{Hyperparameters}\label{app:hyperparameters}
As described in Section~\ref{sec:improving_robustness}, we fine-tune the model via SFT on the ground-truth solution traces for each problem variation. The model is trained for 300 steps with a batch size of 16. To identify suitable hyperparameters, we conduct Bayesian optimization over a predefined set of values: learning rate $\eta \in \{1 \times 10^{-3}, 1 \times 10^{-4}, 5 \times 10^{-5}, 1 \times 10^{-5}\}$, number of gradient accumulation steps $k \in \{4, 8, 12\}$, warm-up ratio $\alpha_{\text{warmup}} \in \{0.0, 0.01, 0.03, 0.05\}$, and weight decay $\lambda \in \{0.0, 0.05, 0.1, 0.3\}$. Across more than 75 independent training runs for each training split, we select the hyperparameter configuration that achieves the highest validation accuracy. The final hyperparameters for each training split are presented in Table~\ref{tab:full_sft_hyperparam_gsms_across_templates}. After training the model with the selected configuration, we evaluate its performance on the test split(s), as described in Section~\ref{sec:improving_robustness} and Appendix~\ref{app:dataset_stats}.

\begin{table*}[tbp]
    \centering
    \scriptsize
    \begin{tabular}{l|cccc|cccc}
        \toprule
        \multicolumn{1}{c}{\multirow{2}{*}{\textbf{Train set}}}
        & \multicolumn{4}{c}{\textbf{GSM-Symbolic}}
        & \multicolumn{4}{c}{\textbf{FinChain}}\\
        \cmidrule(lr){2-5}\cmidrule(lr){6-9}
        \multicolumn{1}{c}{} & $\bm{\eta}$ & $\bm{k}$ & $\bm{\alpha_{\text{warmup}}}$ & \multicolumn{1}{c}{$\bm{\lambda}$} 
        & $\bm{\eta}$ & $\bm{k}$ & $\bm{\alpha_{\text{warmup}}}$ & $\bm{\lambda}$ \\ 
        \midrule
        $\mathrm{Q_{\mathrm{low}}}$ (100\%)  & $1 \times 10^{-5}$ & 8 & 0.03 & 0.10 & $1 \times 10^{-5}$ & 12 & 0.03 & 0.00  \\
        $\mathrm{Q_{\mathrm{mid}}}$ (100\%)  & $1 \times 10^{-5}$ & 8 & 0.03 & 0.05 & $1 \times 10^{-5}$ & 12 & 0.03 & 0.00 \\
        $\mathrm{Q_{\mathrm{high}}}$ (100\%)  & $1 \times 10^{-5}$ & 8 & 0.05 & 0.00 & $1 \times 10^{-5}$ & 12 & 0.03 & 0.10 \\
        \midrule
        $\mathrm{Q_{\mathrm{low}}}$ (30\%) - $\mathrm{Q_{\mathrm{mid}}}$ (70\%)  & $1 \times 10^{-5}$ & 12 & 0.03 & 0.30 & $1 \times 10^{-5}$ & 12 & 0.03 & 0.00 \\
        $\mathrm{Q_{\mathrm{low}}}$ (70\%) - $\mathrm{Q_{\mathrm{mid}}}$ (30\%)  & $1 \times 10^{-5}$ & 8 & 0.00 & 0.00 & $1 \times 10^{-5}$ & 4 & 0.00 & 0.05 \\
        $\mathrm{Q_{\mathrm{low}}}$ (30\%) - $\mathrm{Q_{\mathrm{high}}}$ (70\%)  & $1 \times 10^{-5}$ & 4 & 0.05 & 0.00 & $1 \times 10^{-5}$ & 12 & 0.03 & 0.30 \\
        $\mathrm{Q_{\mathrm{low}}}$ (70\%) - $\mathrm{Q_{\mathrm{high}}}$ (30\%)  & $1 \times 10^{-5}$ & 8 & 0.03 & 0.00 & $1 \times 10^{-5}$ & 12 & 0.00 & 0.30 \\
        $\mathrm{Q_{\mathrm{mid}}}$ (30\%) - $\mathrm{Q_{\mathrm{high}}}$ (70\%)  & $1 \times 10^{-5}$ & 4 & 0.00 & 0.05 & $1 \times 10^{-5}$ & 12 & 0.00 & 0.10 \\
        $\mathrm{Q_{\mathrm{mid}}}$ (70\%) - $\mathrm{Q_{\mathrm{high}}}$ (30\%)  & $1 \times 10^{-5}$ & 4 & 0.03 & 0.05 & $1 \times 10^{-5}$ & 4 & 0.01 & 0.05 \\
        $\mathrm{Q_{\mathrm{low}}}$ (50\%) - $\mathrm{Q_{\mathrm{mid}}}$ (30\%) - $\mathrm{Q_{\mathrm{high}}}$ (20\%)  & $1 \times 10^{-5}$ & 4 & 0.03 & 0.30 & $1 \times 10^{-5}$ & 12 & 0.03 & 0.00 \\
        $\mathrm{Q_{\mathrm{low}}}$ (20\%) - $\mathrm{Q_{\mathrm{mid}}}$ (30\%) - $\mathrm{Q_{\mathrm{high}}}$ (50\%) & $1 \times 10^{-5}$ & 8 & 0.05 & 0.30 & $1 \times 10^{-5}$ & 12 & 0.05 & 0.10 \\
        $\mathrm{Q_{\mathrm{low}}}$ (20\%) - $\mathrm{Q_{\mathrm{mid}}}$ (20\%) - $\mathrm{Q_{\mathrm{high}}}$ (60\%) & $1 \times 10^{-5}$ & 4 & 0.03 & 0.00 & $1 \times 10^{-5}$ & 8 & 0.00 & 0.30 \\
        $\mathrm{Q_{\mathrm{low}}}$ (10\%) - $\mathrm{Q_{\mathrm{mid}}}$ (10\%) - $\mathrm{Q_{\mathrm{high}}}$ (80\%) & $1 \times 10^{-5}$ & 4 & 0.01 & 0.05 & $1 \times 10^{-5}$ & 12 & 0.01 & 0.30 \\
        Random sampling  & $1 \times 10^{-5}$ & 8 & 0.05 & 0.30 & $1 \times 10^{-5}$ & 12 & 0.03 & 0.30 \\
        \bottomrule
    \end{tabular}
    \caption{\textbf{Final hyperparameter configuration.} Hyperparameter achieving the highest validation accuracy for each training set, selected via Bayesian optimization.}
    \label{tab:full_sft_hyperparam_gsms_across_templates}
\end{table*}

\section{Additional results}\label{app:additional_results}
In this section, we present additional experiments and results for other datasets and models.

\subsection{Quantifying variation difficulty}\label{app:assessing_variation_difficulty}
In Section~\ref{subsec:pred_answer_correctness}, we analyze how well different metrics $f$\textemdash{}such as perplexity, entropy, self-certainty, $\text{LD}_{\min}$, and $\text{MD}_{\mathcal{H}}$\textemdash{}capture variation difficulty by evaluating their ability to predict answer correctness. Specifically, we compare (i) each metric's micro-averaged AUC, as defined in Equation~\ref{eq:auc_score}, and (ii) the change in the odds of a correct answer associated with a one-standard-deviation increase in the z-scored version of $f$ (see Section~\ref{subsec:pred_answer_correctness} and Appendix~\ref{app:measuring_predictive_power}).

\begin{figure*}[b]
     \centering
     \captionsetup[subfigure]{font=scriptsize}
     \begin{subfigure}[b]{0.245\textwidth}
         \centering
         \includegraphics[width=\linewidth]{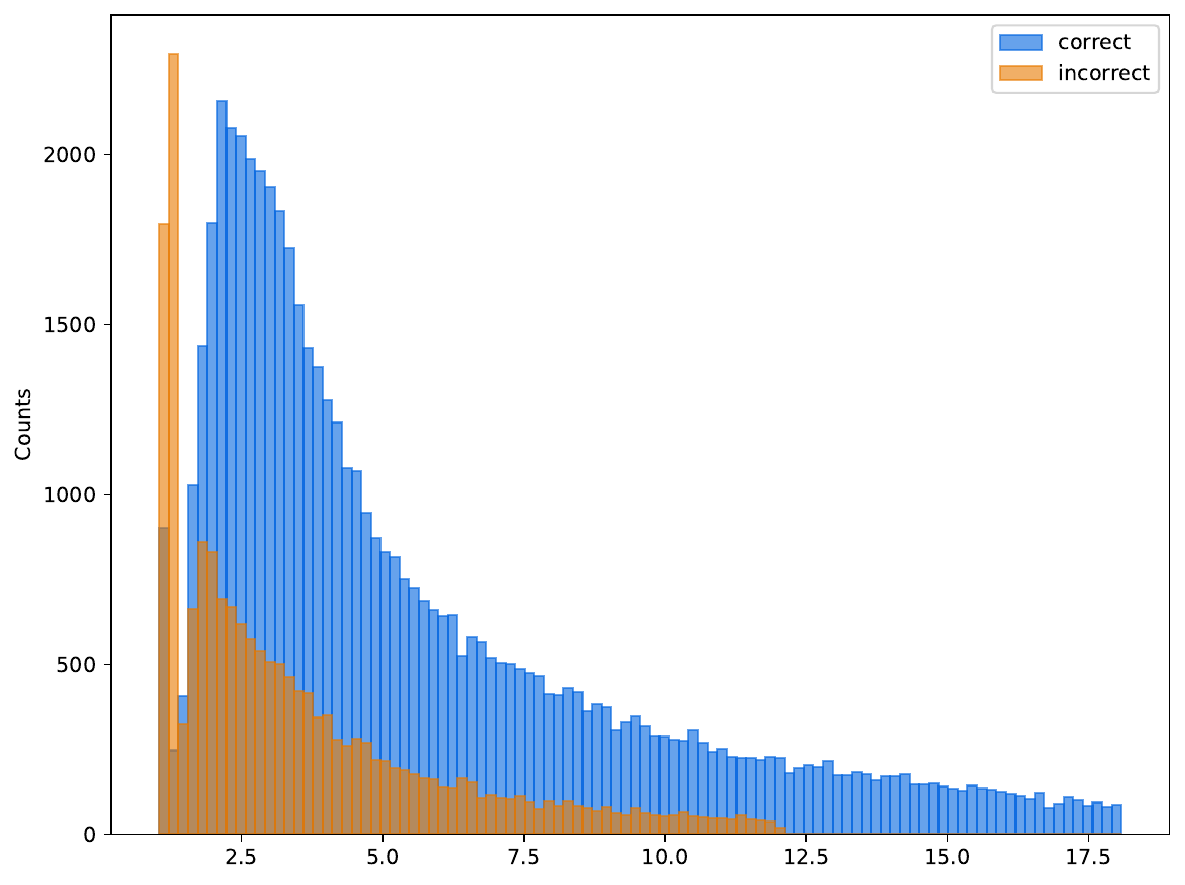}
         \caption{Perplexity}
         \label{fig:perplexity_gsms_llama_8b}
     \end{subfigure}
     \begin{subfigure}[b]{0.245\textwidth}
         \centering
         \includegraphics[width=\linewidth]{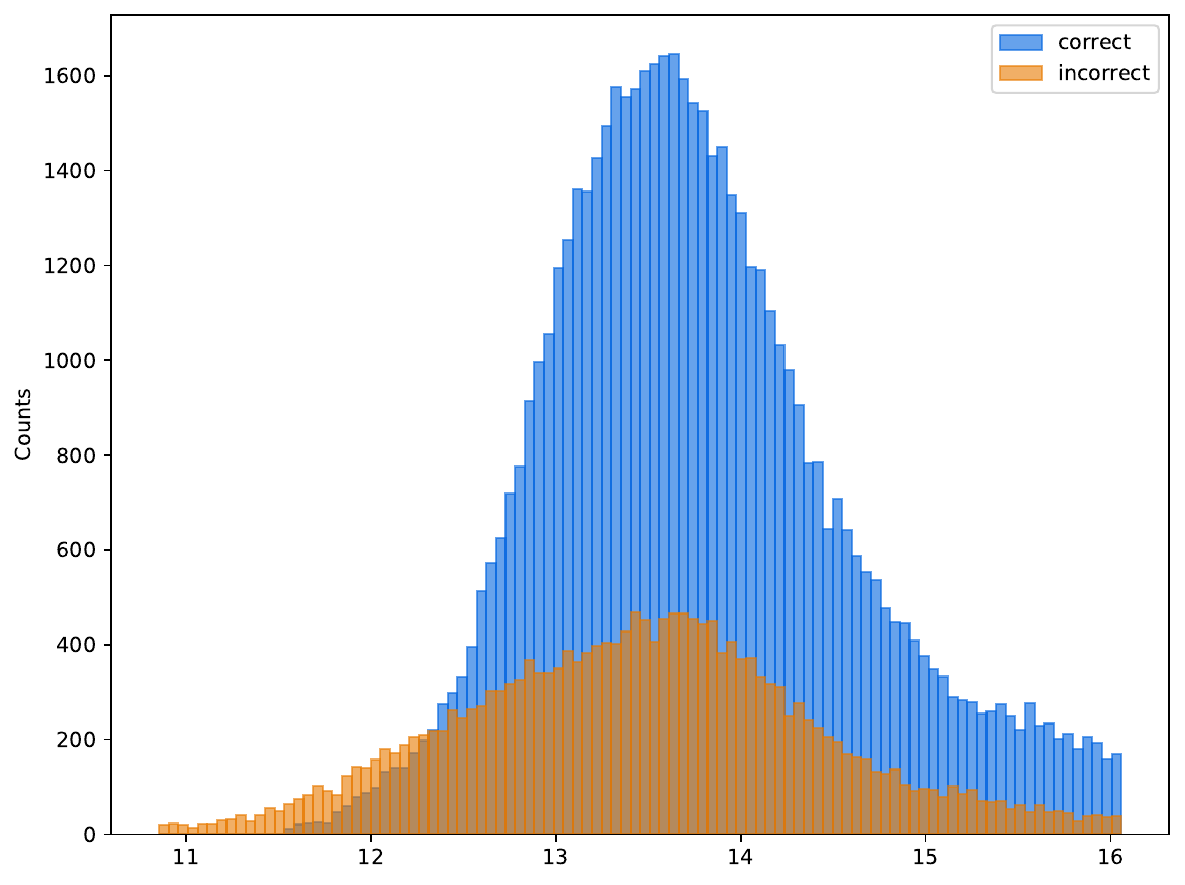}
         \caption{Self-Certainty}
         \label{fig:certainty_gsms_llama_8b}
     \end{subfigure}
     \begin{subfigure}[b]{0.245\textwidth}
         \centering
         \includegraphics[width=\linewidth]{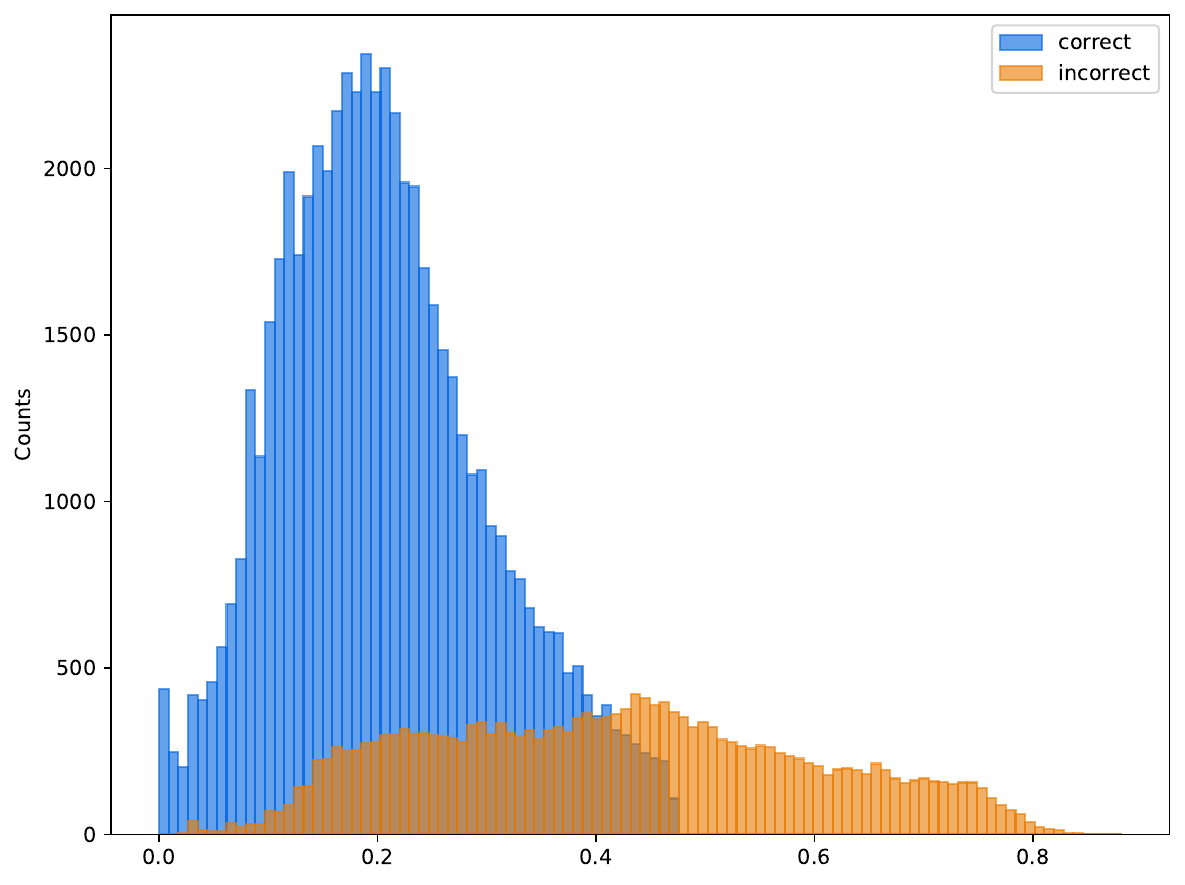}
         \caption{Levenshtein distance}
         \label{fig:levenshtein_gsms_llama_8b}
     \end{subfigure}
     \begin{subfigure}[b]{0.245\textwidth}
         \centering
         \includegraphics[width=\linewidth]{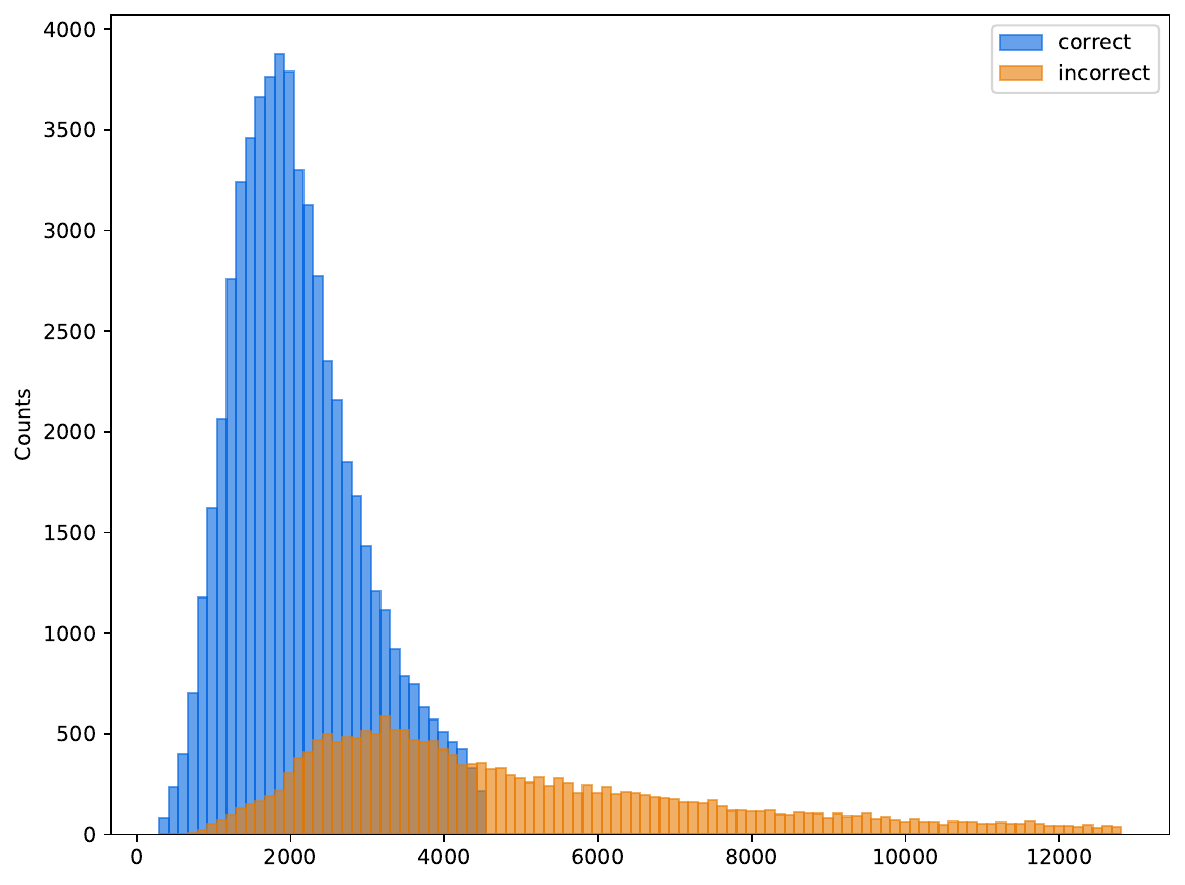}
         \caption{Mahalanobis distance}
         \label{fig:md_gsms_llama_8b}
     \end{subfigure}
     \caption{\textbf{Distribution of metric scores for correct and incorrect responses.} Answers are obtained from \emph{Llama-3.1-8B-Instruct} across 1,000 samples per \emph{GSM-Symbolic} template.}
     \label{fig:predicting_answer_correctness_llama_8b_gsms}
\end{figure*}

\begin{figure*}[b]
     \centering
     \captionsetup[subfigure]{font=scriptsize}
     \begin{subfigure}[b]{0.245\textwidth}
         \centering
         \includegraphics[width=\linewidth]{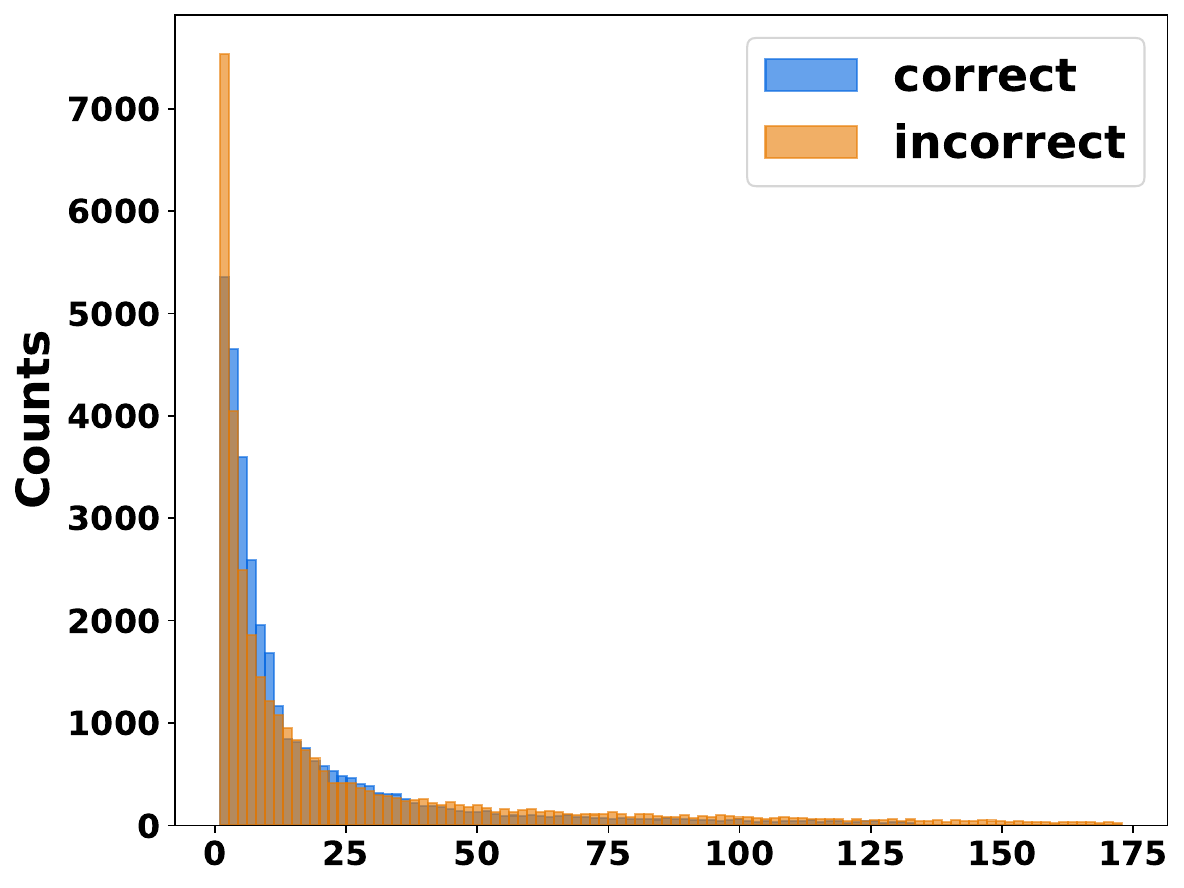}
         \caption{Perplexity}
         \label{fig:perplexity_finchain_llama_8b}
     \end{subfigure}
     \begin{subfigure}[b]{0.245\textwidth}
         \centering
         \includegraphics[width=\linewidth]{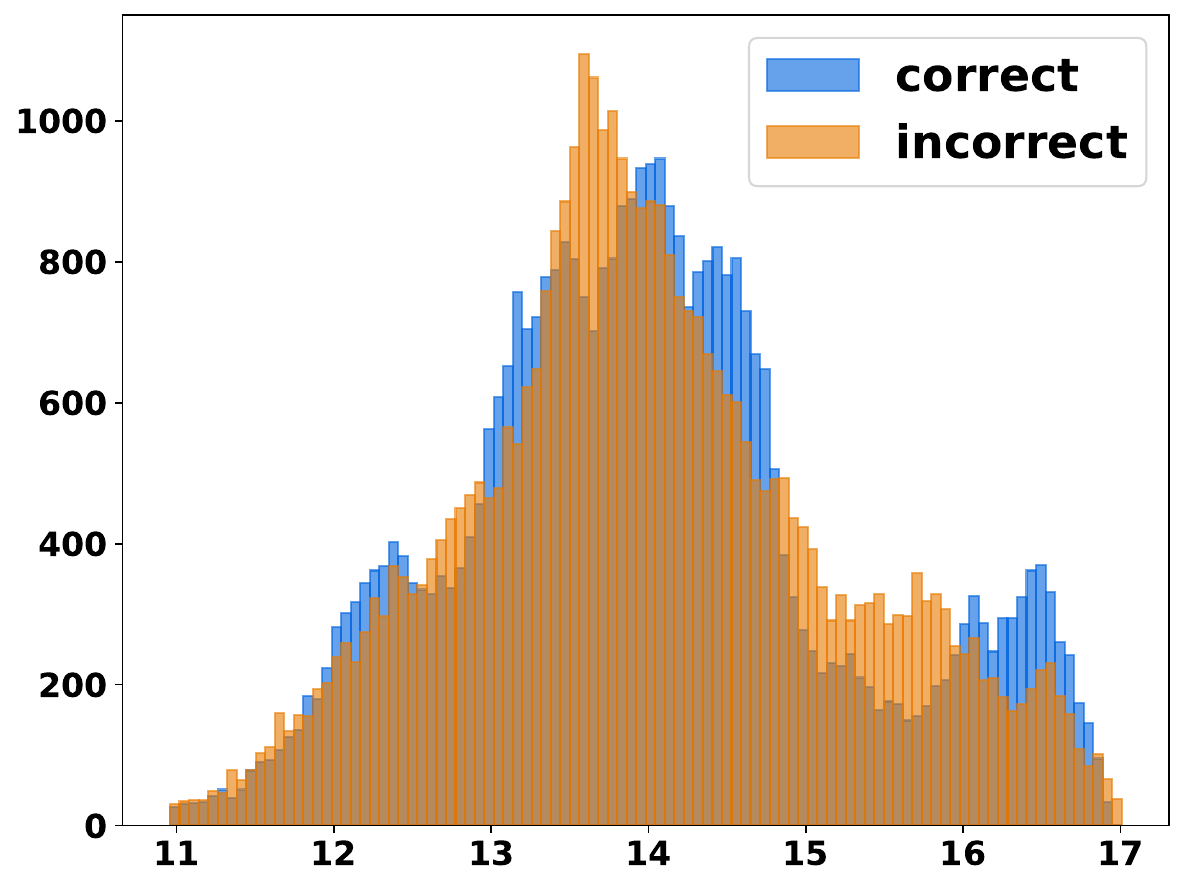}
         \caption{Self-Certainty}
         \label{fig:certainty_finchain_llama_8b}
     \end{subfigure}
     \begin{subfigure}[b]{0.245\textwidth}
         \centering
         \includegraphics[width=\linewidth]{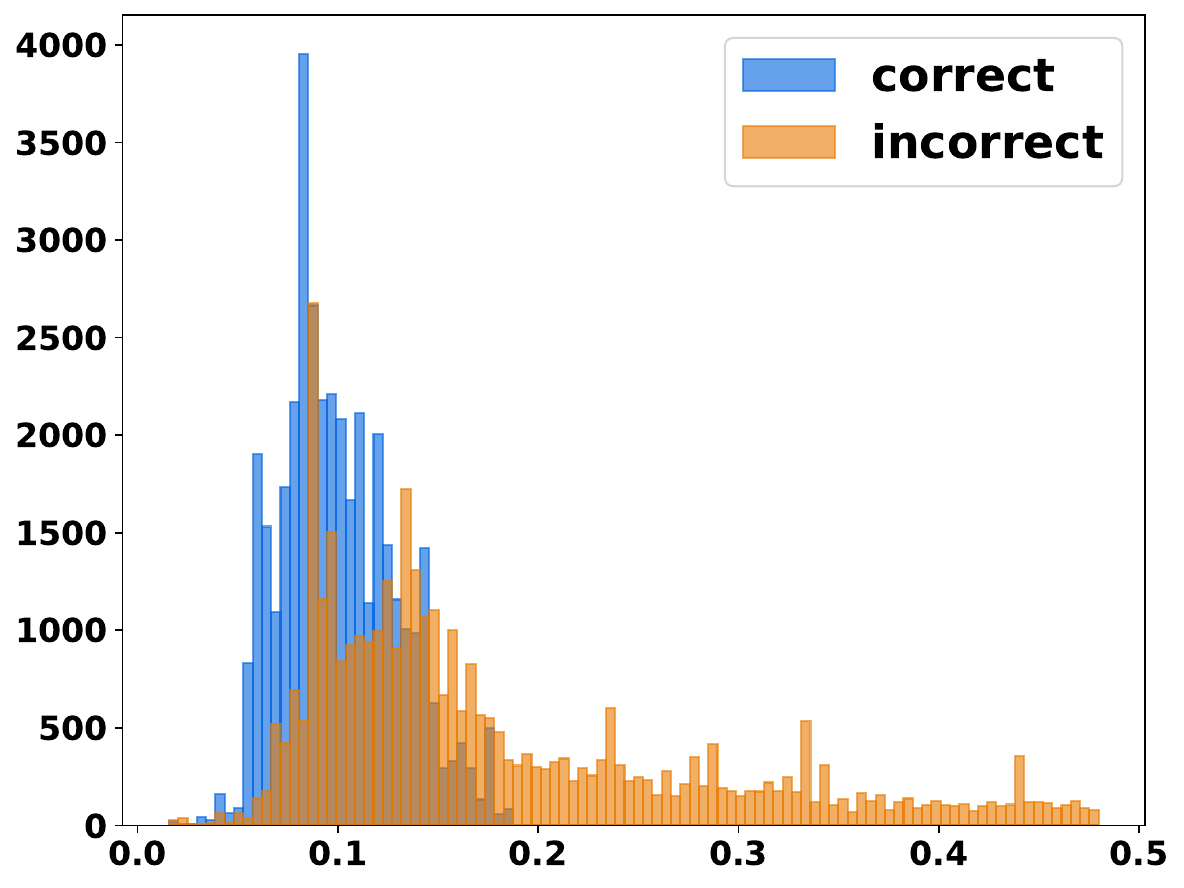}
         \caption{Levenshtein distance}
         \label{fig:levenshtein_finchain_llama_8b}
     \end{subfigure}
     \begin{subfigure}[b]{0.245\textwidth}
         \centering
         \includegraphics[width=\linewidth]{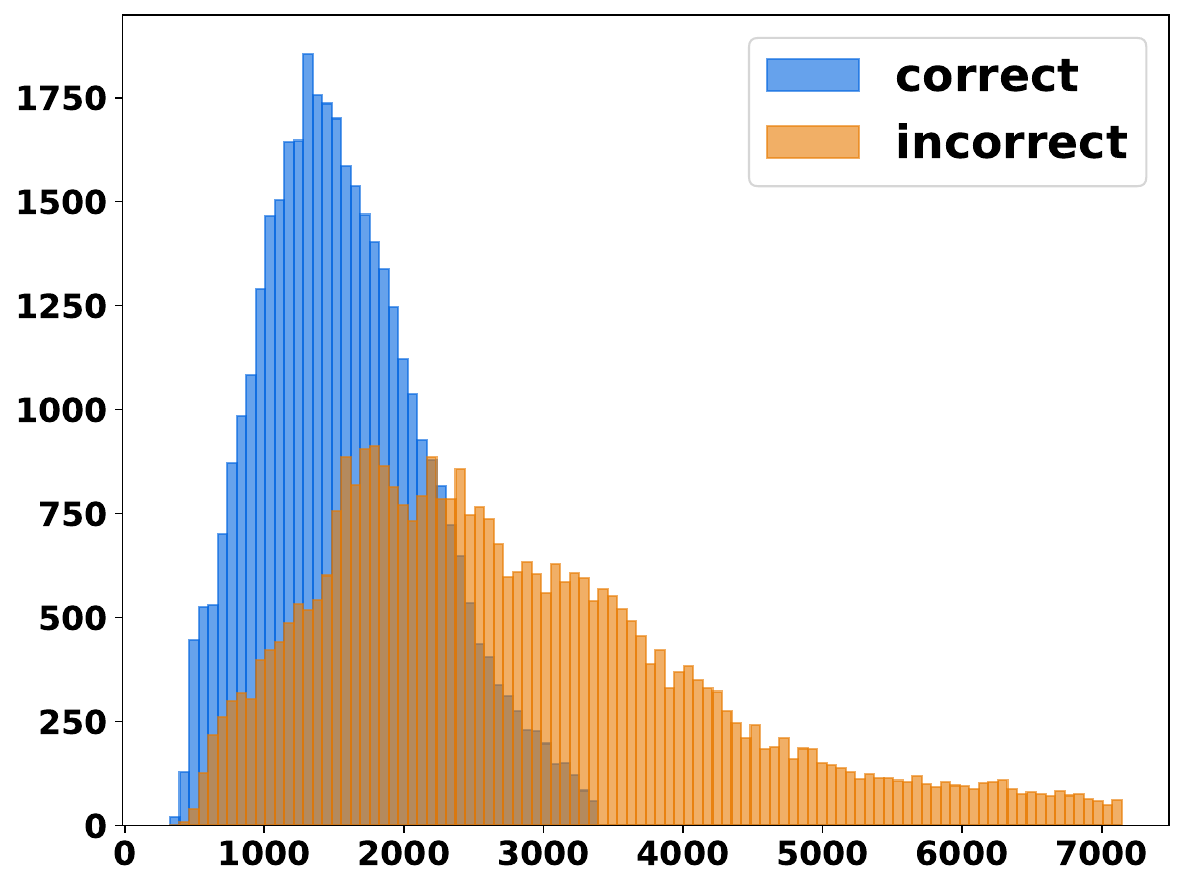}
         \caption{Mahalanobis distance}
         \label{fig:md_finchain_llama_8b}
     \end{subfigure}
     \caption{\textbf{Distribution of metric scores for correct and incorrect responses.} Answers are obtained from \emph{Llama-3.1-8B-Instruct} across 1,000 samples per \emph{FinChain} template.}
     \label{fig:predicting_answer_correctness_llama_8b_finchain}
\end{figure*}

\begin{figure*}[b]
     \centering
     \captionsetup[subfigure]{font=scriptsize}
     \begin{subfigure}[b]{0.245\textwidth}
         \centering
         \includegraphics[width=\linewidth]{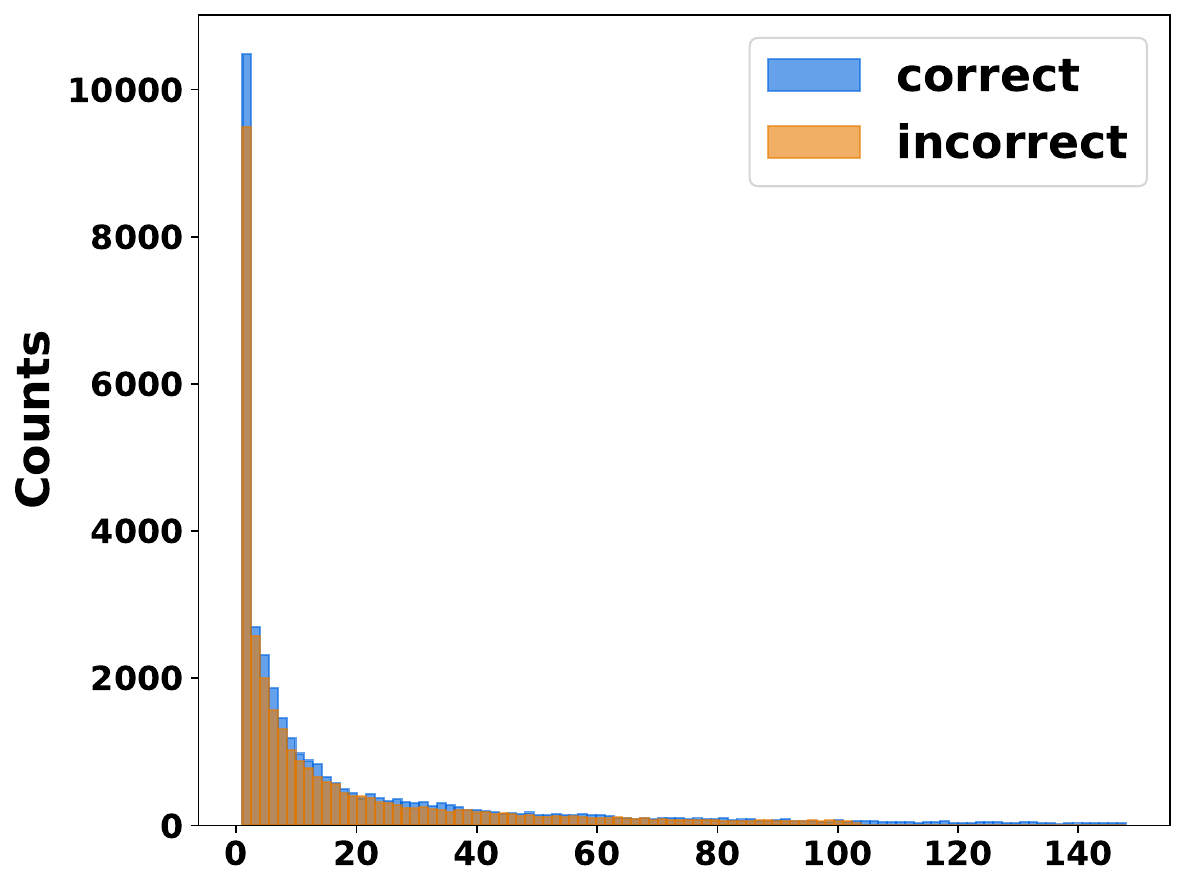}
         \caption{Perplexity}
         \label{fig:perplexity_engtrace_qwen_32b}
     \end{subfigure}
     \begin{subfigure}[b]{0.245\textwidth}
         \centering
         \includegraphics[width=\linewidth]{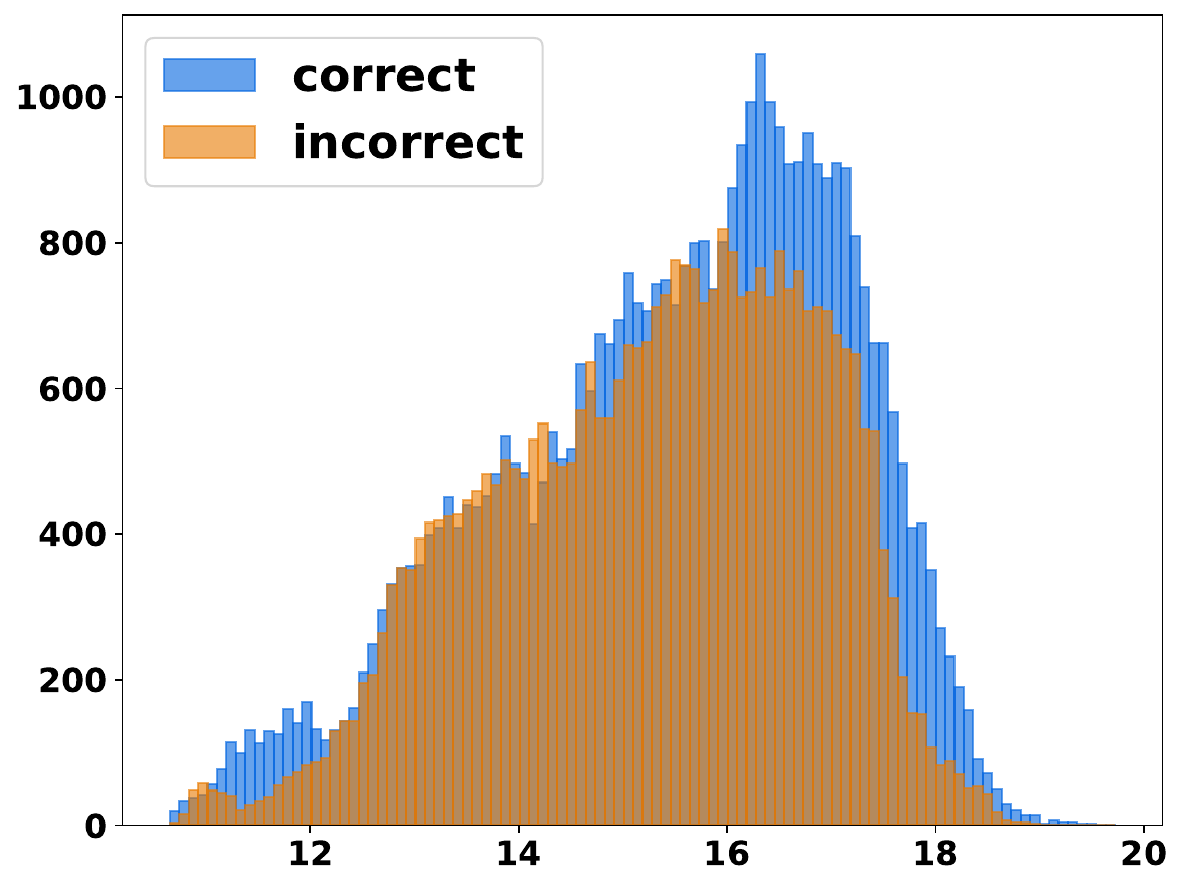}
         \caption{Self-Certainty}
         \label{fig:certainty_engtrace_qwen_32b}
     \end{subfigure}
     \begin{subfigure}[b]{0.245\textwidth}
         \centering
         \includegraphics[width=\linewidth]{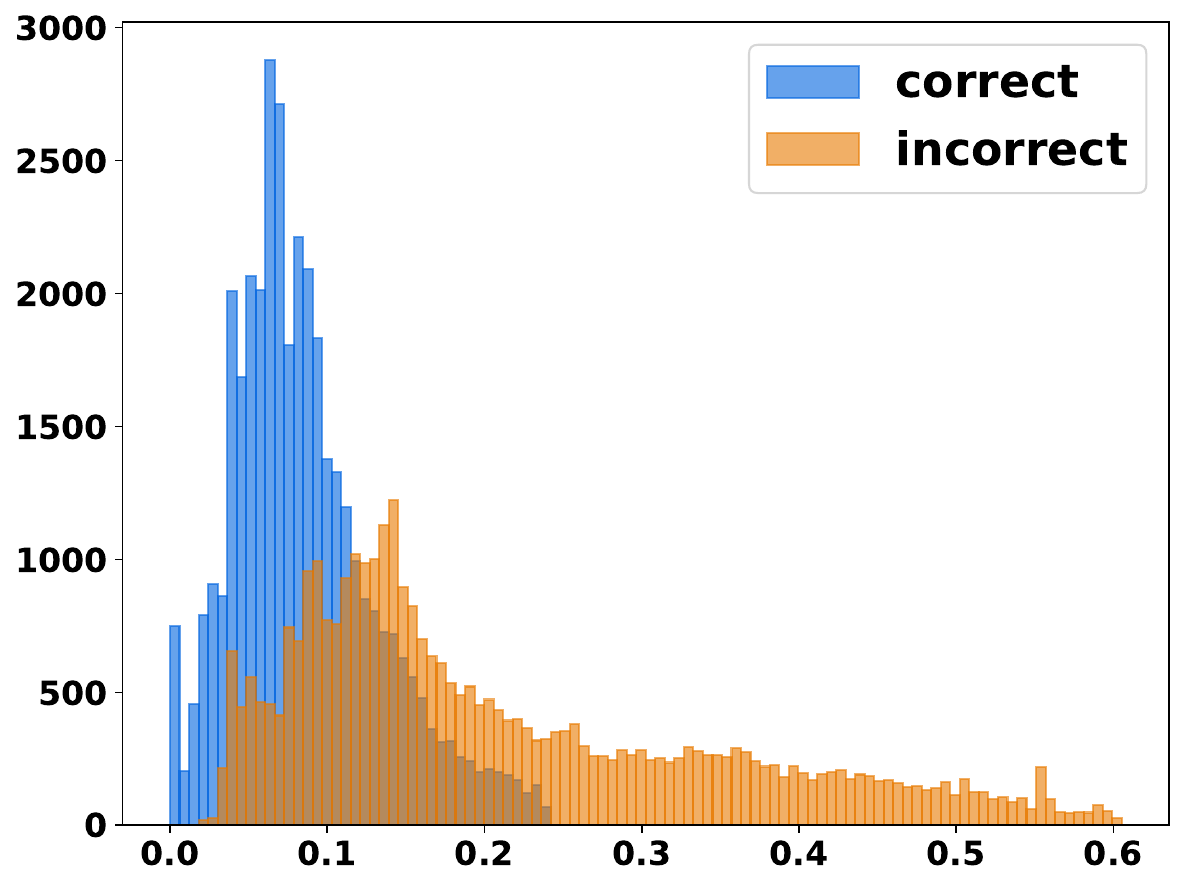}
         \caption{Levenshtein distance}
         \label{fig:levenshtein_engtrace_qwen_32b}
     \end{subfigure}
     \begin{subfigure}[b]{0.245\textwidth}
         \centering
         \includegraphics[width=\linewidth]{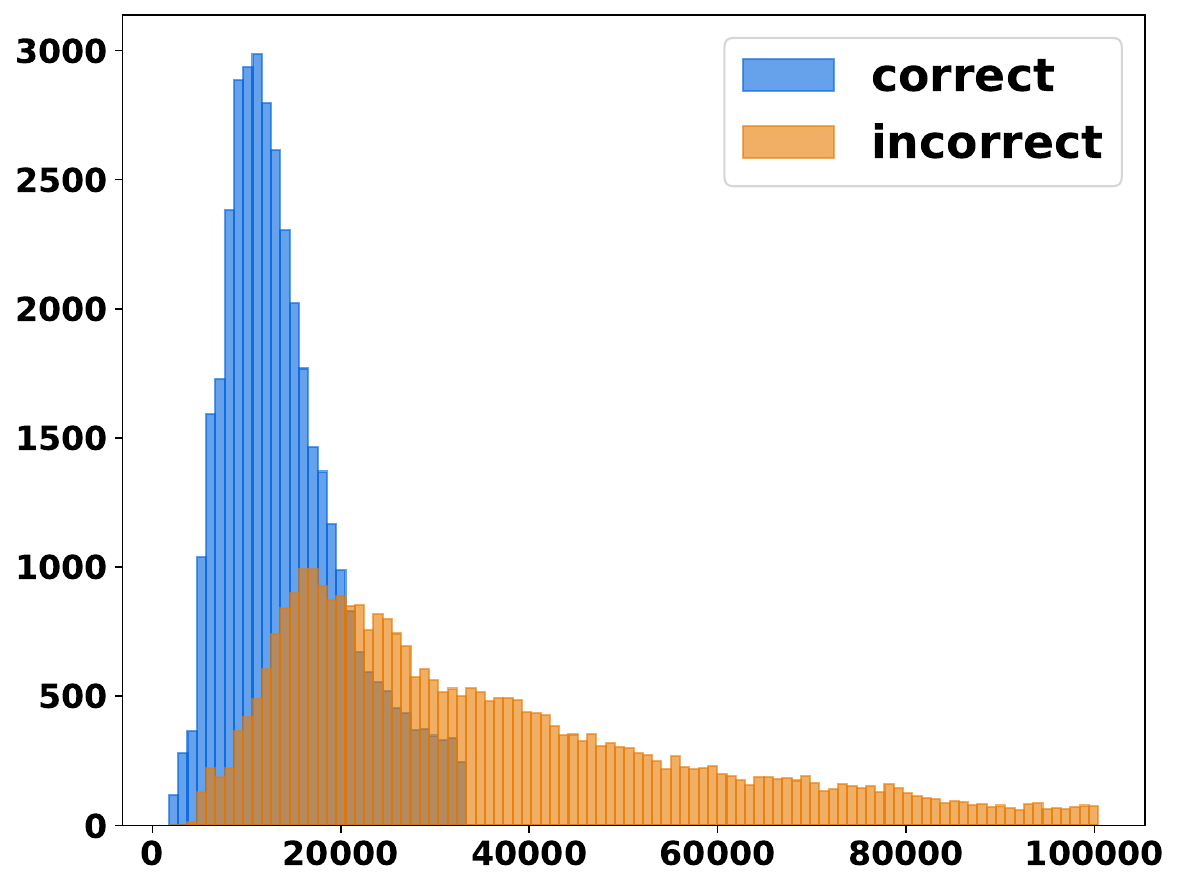}
         \caption{Mahalanobis distance}
         \label{fig:md_engtrace_qwen_32b}
     \end{subfigure}
     \caption{\textbf{Distribution of metric scores for correct and incorrect responses.} Answers are obtained from \emph{Qwen2.5-32B-Instruct} across 1,000 samples per \emph{EngTrace} template.}
     \label{fig:predicting_answer_correctness_qwen_32b_engtrace}
\end{figure*}

While Table~\ref{tab:correlation_llama_3.1_8B} reports AUC scores and odds ratios for Llama-3.1-8B-Instruct across GSM-Symbolic, FinChain, and EngTrace, Table~\ref{tab:correlation_llama_family} summarizes the corresponding results for all models from the Llama family\textemdash{}Llama-3.2-3B-Instruct, Llama-3.1-8B-Instruct, and Llama-3.1-70B-Instruct. Results for models from the Qwen family\textemdash{}Qwen-2.5-7B-Instruct, Qwen-2.5-32B-Instruct, and Qwen-2.5-72B-Instruct\textemdash{}are shown in Tables~\ref{tab:correlation_qwen_family} and~\ref{tab:correlation_qwen_72b}, and findings for Phi-4 are presented in Table~\ref{tab:correlation_phi4_14b}. Following Section~\ref{subsec:pred_answer_correctness}, we compare metrics computed from the models' input (i.e., using only $p_i$) with metrics computed over the models’ output $Y_i$. Across all models and datasets, input-only metrics exhibit substantially lower predictive power than output-based metrics. Moreover, reference-based metrics such as the Levenshtein and Mahalanobis distances introduced in Section~\ref{subsec:ref_based_distance} significantly outperform metrics derived from the models' internal probability distribution, such as perplexity, entropy, and self-certainty. In almost all settings, $\text{MD}_{\mathcal{H}}$ achieves the highest micro-averaged AUC and is often paired with the strongest odds ratios. Among Levenshtein pooling operations (min, max, mean, median), the minimum Levenshtein distance $\text{LD}_{\min}$ typically performs best. Figure~\ref{fig:predicting_answer_correctness_llama_8b_gsms} shows metric-score distributions by answer correctness for Llama-3.1-8B-Instruct on GSM-Symbolic, Figure~\ref{fig:predicting_answer_correctness_llama_8b_finchain} for FinChain, and Figure~\ref{fig:predicting_answer_correctness_qwen_32b_engtrace} for Qwen-2.5-32B-Instruct on EngTrace. Across models and datasets, $\text{LD}_{\min}$ and $\text{MD}_{\mathcal{H}}$ more clearly separate correctly answered from incorrectly answered variations than other metrics, underscoring the strong predictive power of the proposed reference-based metrics.

\paragraph{\textbf{Predictive power generalizes across reference distributions.}} Both $\text{LD}_{\min}$ and $\text{MD}_{\mathcal{H}}$ depend on a reference set $\mathcal{Y} = \{Y_k\}_{k=1}^N$ of correct model responses to problem variations instantiated from a symbolic template. In Table~\ref{tab:cross_model_edit_score}, we analyze how the predictive power of $\text{LD}_{\min}$ changes when the reference set $\mathcal{Y}$ is obtained not from the model itself, but from another LLM, or when ground-truth reasoning traces are used as reference samples instead. This allows us to assess the likelihood that a model response is correct even when a reference set of correct responses cannot be obtained from that model (e.g., due to computational cost or poor model performance).

We find that using either a different reference model or ground-truth solution traces as the reference set reduces the predictive power of $\text{LD}_{\min}$ for answer correctness on GSM-Symbolic samples for all models. We hypothesize that this drop arises because the solution strategies for a given template can be model-specific: reference responses generated by the model under evaluation may better capture its characteristic approaches\textemdash{}and, therefore, how strongly it deviates from these approaches\textemdash{}than responses from another model or the ground-truth traces. Nevertheless, the resulting reductions in AUC and odds ratios are moderate, with ground-truth traces slightly outperforming the use of another reference model. Thus, computing $\text{LD}_{\min}$ using ground-truth traces or a different reference model appears to be a viable alternative when reference responses from the model of interest are unavailable.

\subsection{Benchmarking robustness via LPDS}\label{app:benchmarking_robustness}
In Section~\ref{subsec:lpds_expose_failure}, we show how to assess the robustness of LLMs to logic-preserving variations by evaluating models on problem variations identified via LPDS (see Algorithm~\ref{alg:beam_search}). In this section, we present similar results for additional datasets and models.

\subsubsection{Beam search finds increasingly difficult problem variations}\label{app:identifying_difficult_variations}
\begin{figure*}[tbp]
  \centering
  \begin{subfigure}{0.45\textwidth}
    \centering
    \includegraphics[width=\linewidth]{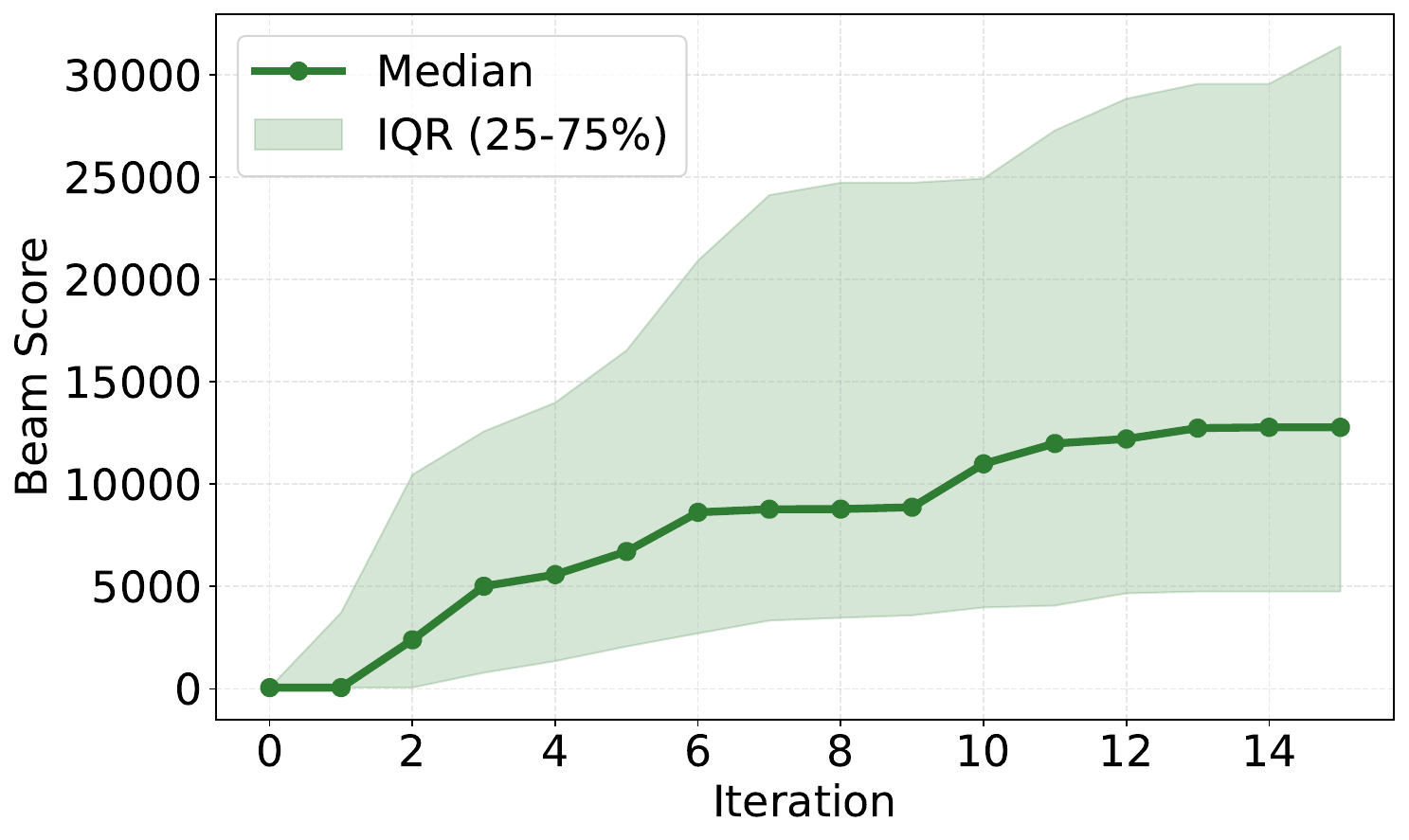}
    \caption{Top score $f^* = \text{MD}_{\mathcal{H}}^*$ in the current beam.}
    \label{fig:beam_score_history_gsms_qwen_7b}
  \end{subfigure}
  \hspace{0.05\textwidth}
  \begin{subfigure}{0.45\textwidth}
    \centering
    \includegraphics[width=\linewidth]{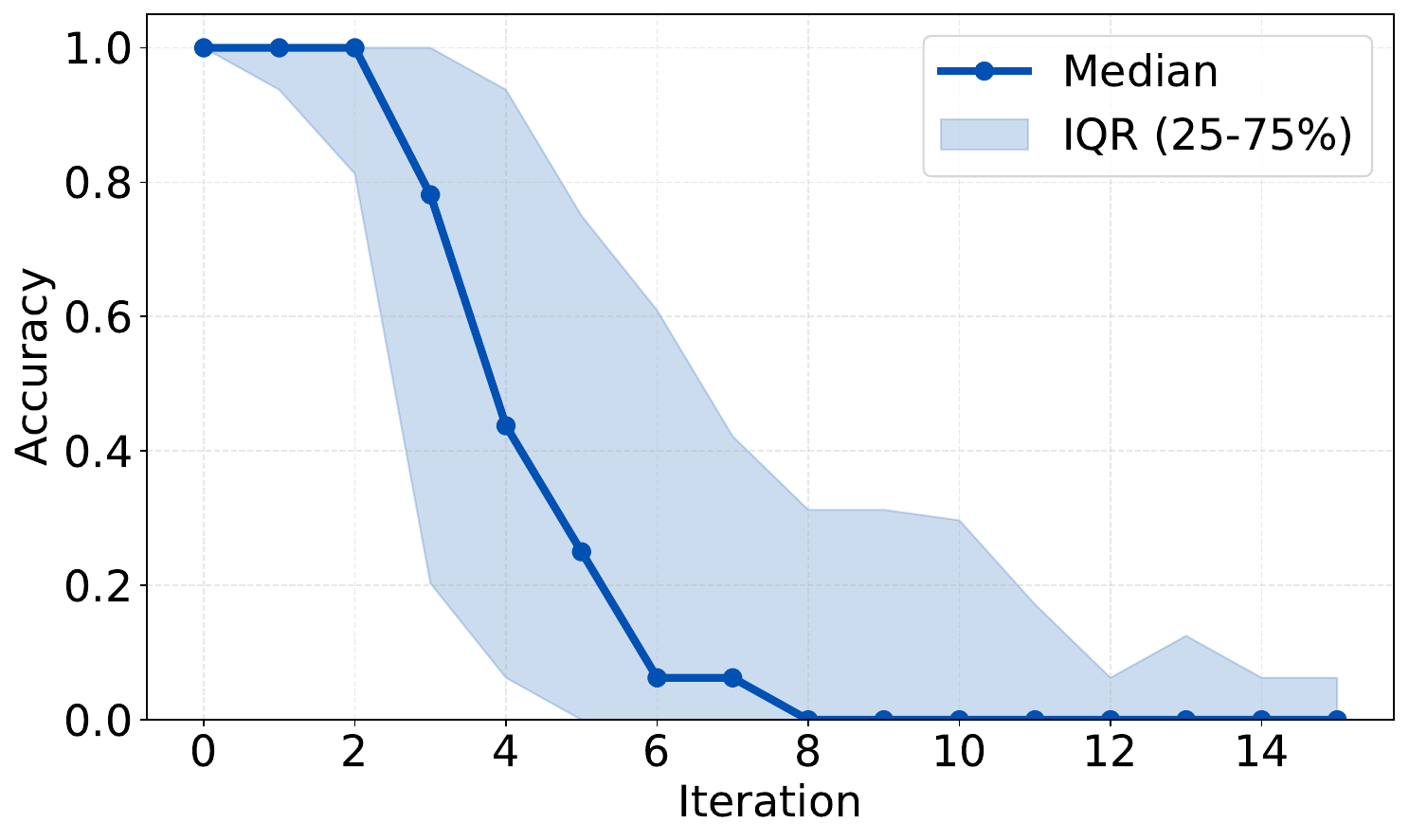}
    \caption{The model's average beam accuracy.}
    \label{fig:beam_acc_history_gsms_qwen_7b}
  \end{subfigure}
  \caption{\textbf{Beam search finds increasingly difficult problem variations.} Across iterations, variation difficulty increases (a), while \emph{Qwen-2.5-7B-Instruct’s} beam accuracy decreases (b). Lines show medians and interquartile ranges across runs over all \emph{GSM-Symbolic} templates.}
  \label{fig:beam_search_gsms_qwen_7b}
\end{figure*}

\begin{figure*}[tbp]
  \centering
  \begin{subfigure}{0.45\textwidth}
    \centering
    \includegraphics[width=\linewidth]{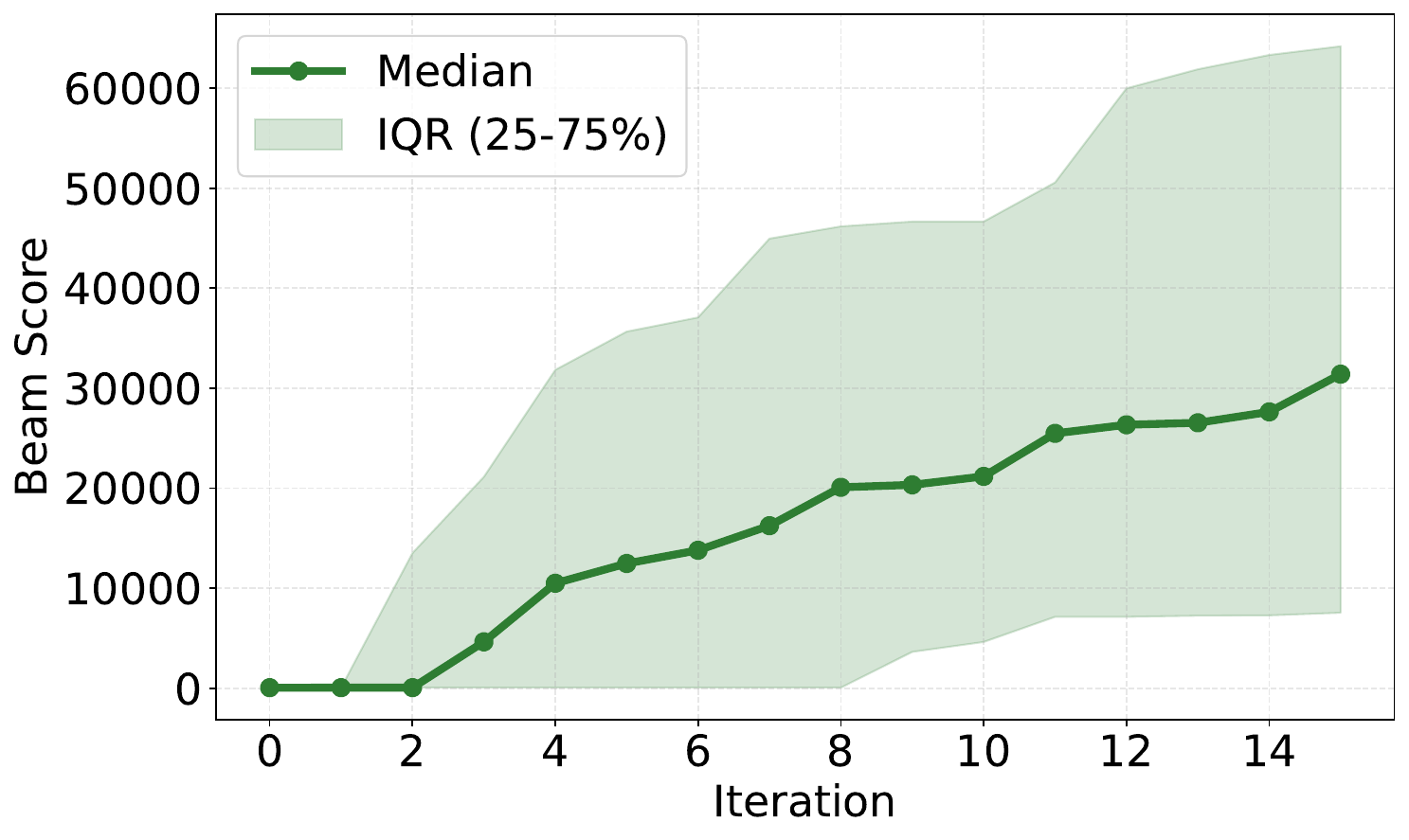}
    \caption{Top score $f^* = \text{MD}_{\mathcal{H}}^*$ in the current beam.}
    \label{fig:beam_score_history_gsms_qwen_32b}
  \end{subfigure}
  \hspace{0.05\textwidth}
  \begin{subfigure}{0.45\textwidth}
    \centering
    \includegraphics[width=\linewidth]{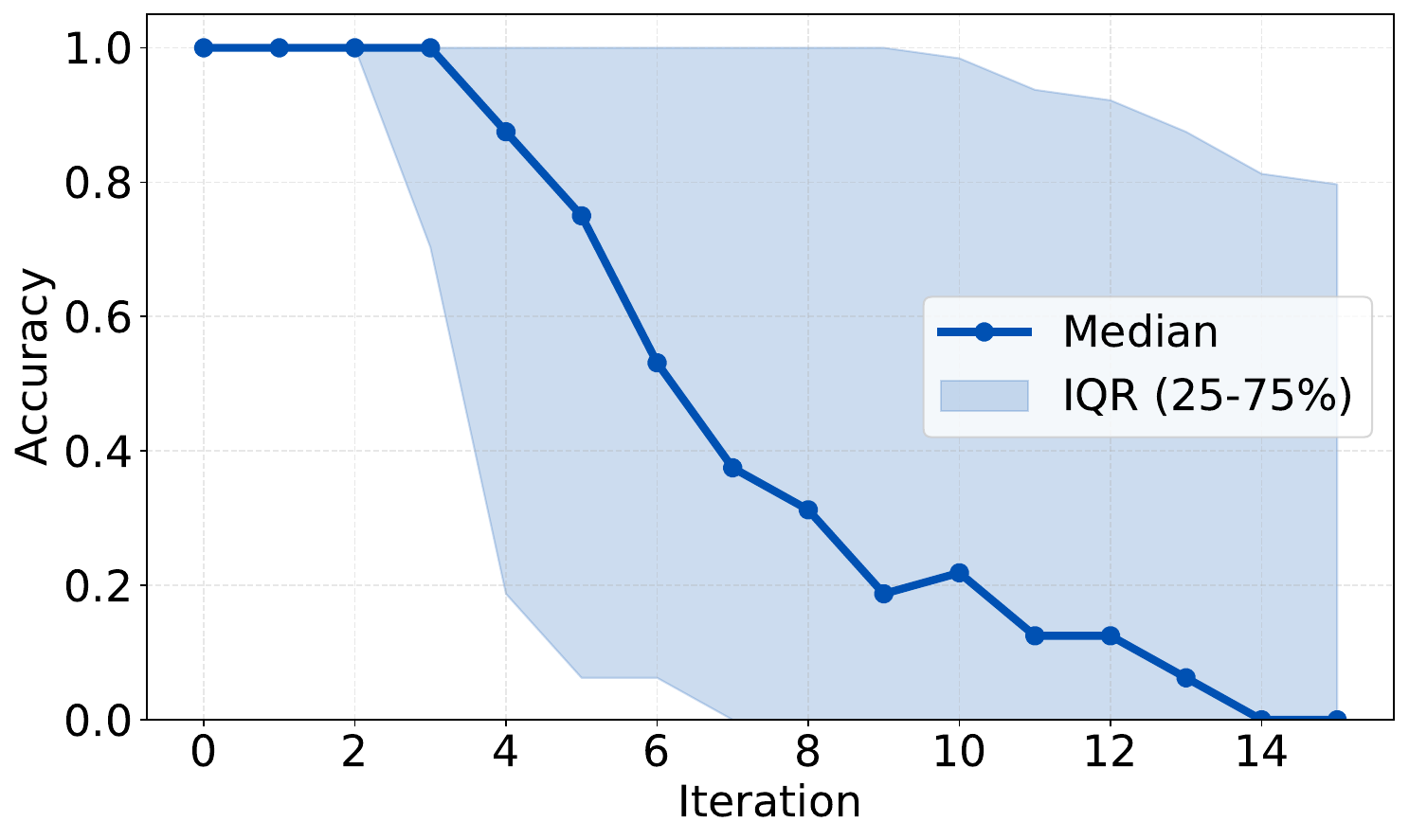}
    \caption{The model's average beam accuracy.}
    \label{fig:beam_acc_history_gsms_qwen_32b}
  \end{subfigure}
  \caption{\textbf{Beam search finds increasingly difficult problem variations.} Across iterations, variation difficulty increases (a), while \emph{Qwen-2.5-32B-Instruct’s} beam accuracy decreases (b). Lines show medians and interquartile ranges across runs over all \emph{GSM-Symbolic} templates.}
  \label{fig:beam_search_gsms_qwen_32b}
\end{figure*}

\begin{figure*}[tbp]
  \centering
  \begin{subfigure}{0.45\textwidth}
    \centering
    \includegraphics[width=\linewidth]{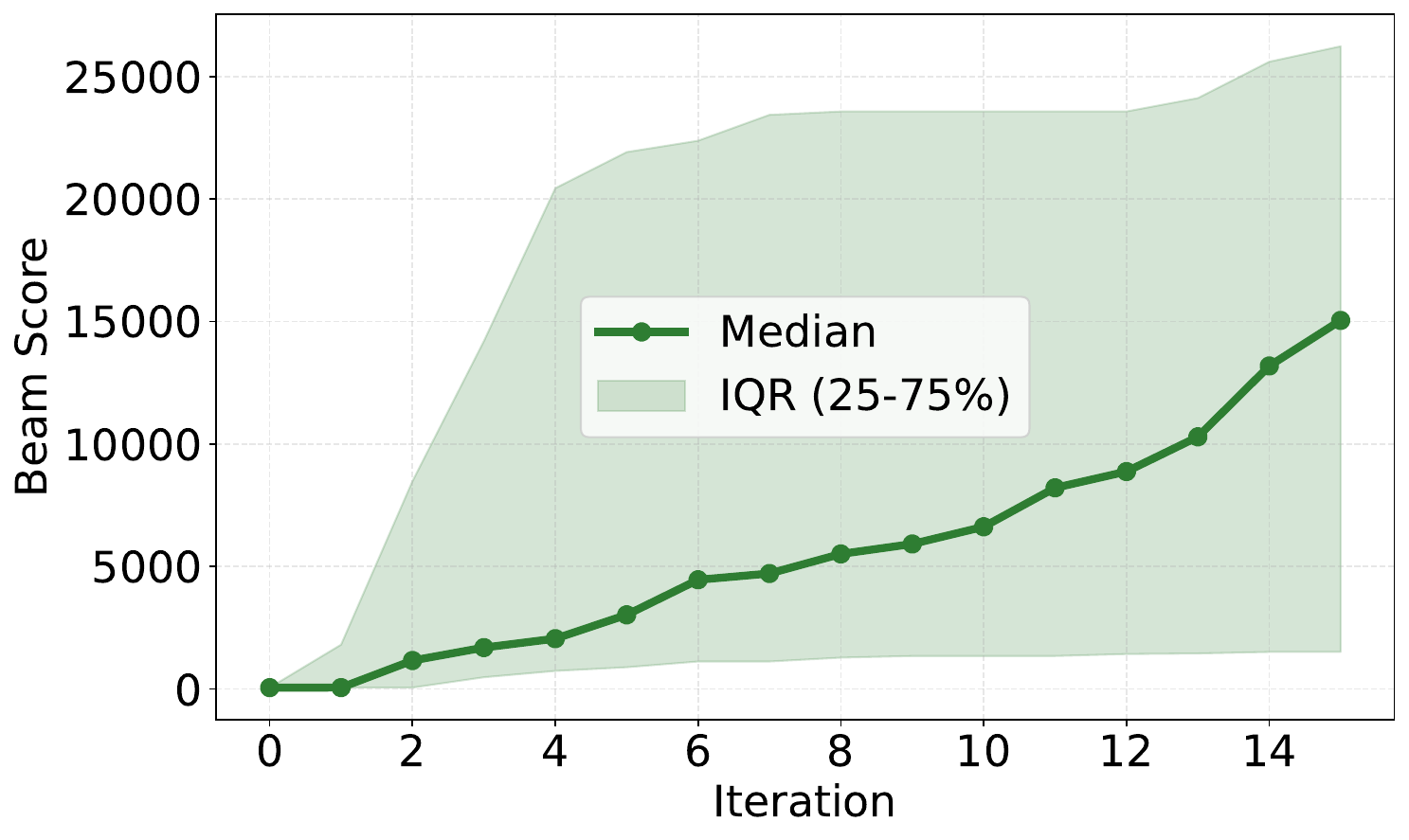}
    \caption{Top score $f^* = \text{MD}_{\mathcal{H}}^*$ in the current beam.}
    \label{fig:beam_score_history_finchain_llama_70b}
  \end{subfigure}
  \hspace{0.05\textwidth}
  \begin{subfigure}{0.45\textwidth}
    \centering
    \includegraphics[width=\linewidth]{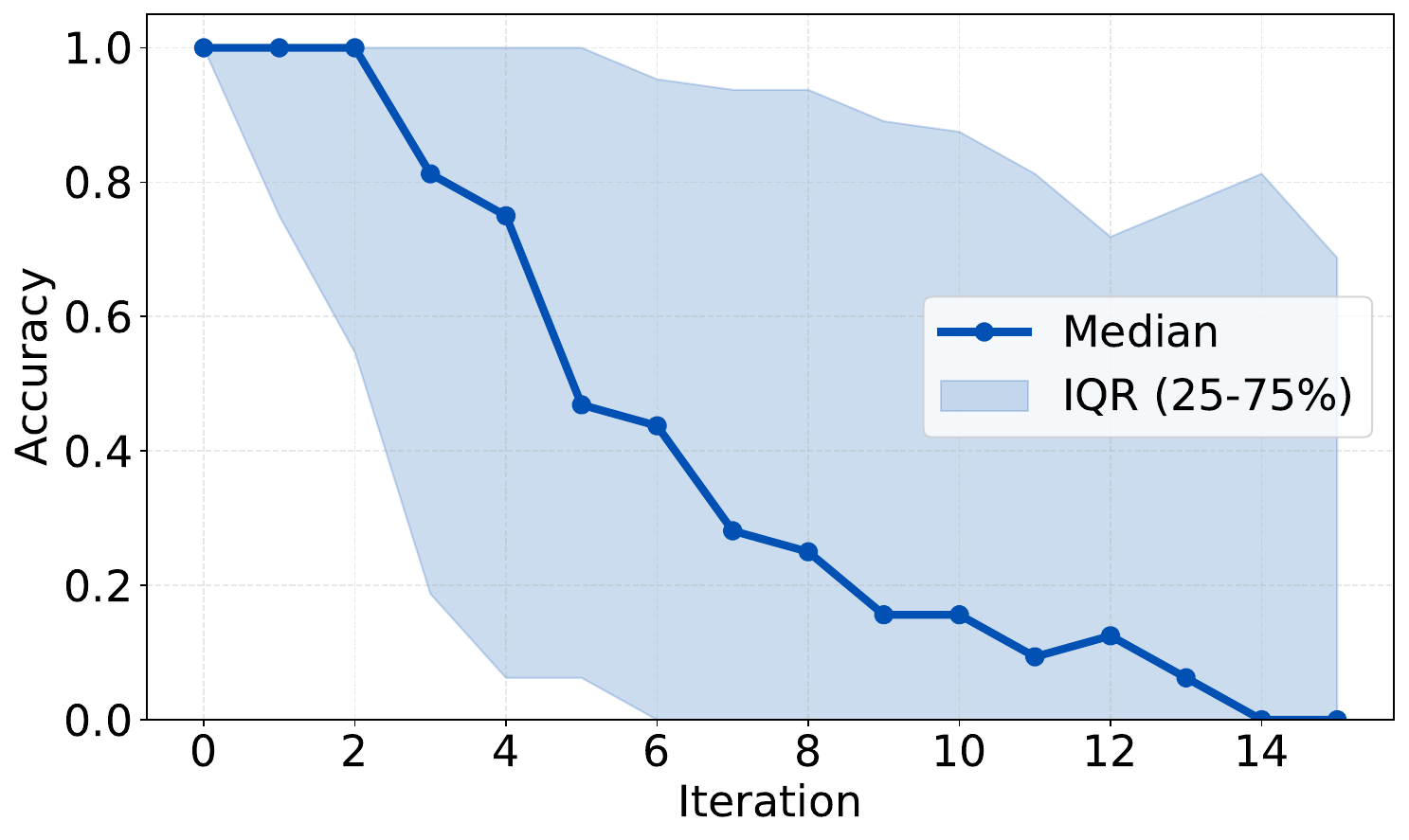}
    \caption{The model's average beam accuracy.}
    \label{fig:beam_acc_history_finchain_llama_70b}
  \end{subfigure}
  \caption{\textbf{Beam search finds increasingly difficult problem variations.} Across iterations, variation difficulty increases (a), while \emph{Llama-3.1-70B-Instruct’s} beam accuracy decreases (b). Lines show medians and interquartile ranges across runs over all \emph{FinChain} templates.}
  \label{fig:beam_search_finchain_llama_70b}
\end{figure*}

\begin{figure*}[tbp]
  \centering
  \begin{subfigure}{0.45\textwidth}
    \centering
    \includegraphics[width=\linewidth]{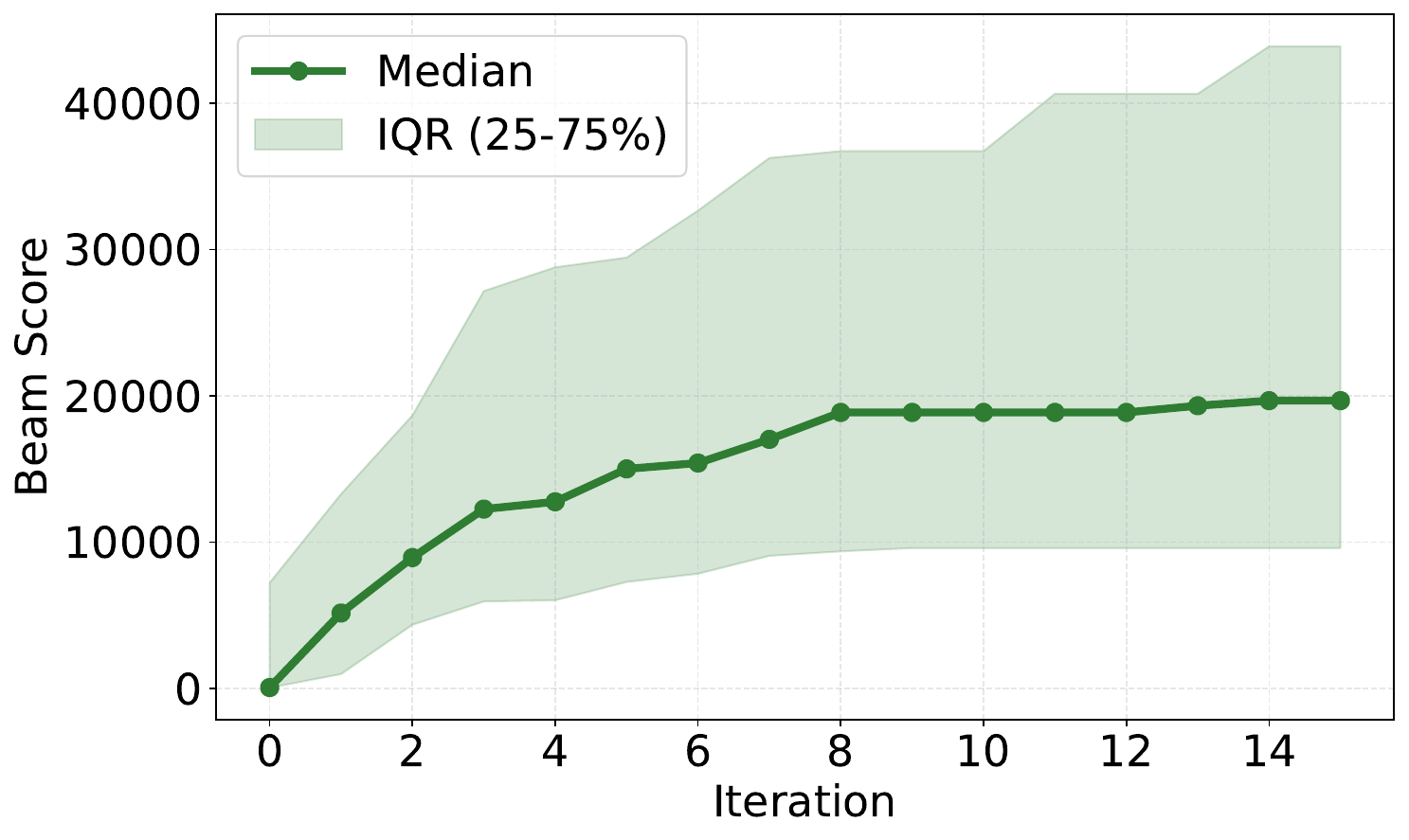}
    \caption{Top score $f^* = \text{MD}_{\mathcal{H}}^*$ in the current beam.}
    \label{fig:beam_score_history_engtrace_qwen_7b}
  \end{subfigure}
  \hspace{0.05\textwidth}
  \begin{subfigure}{0.45\textwidth}
    \centering
    \includegraphics[width=\linewidth]{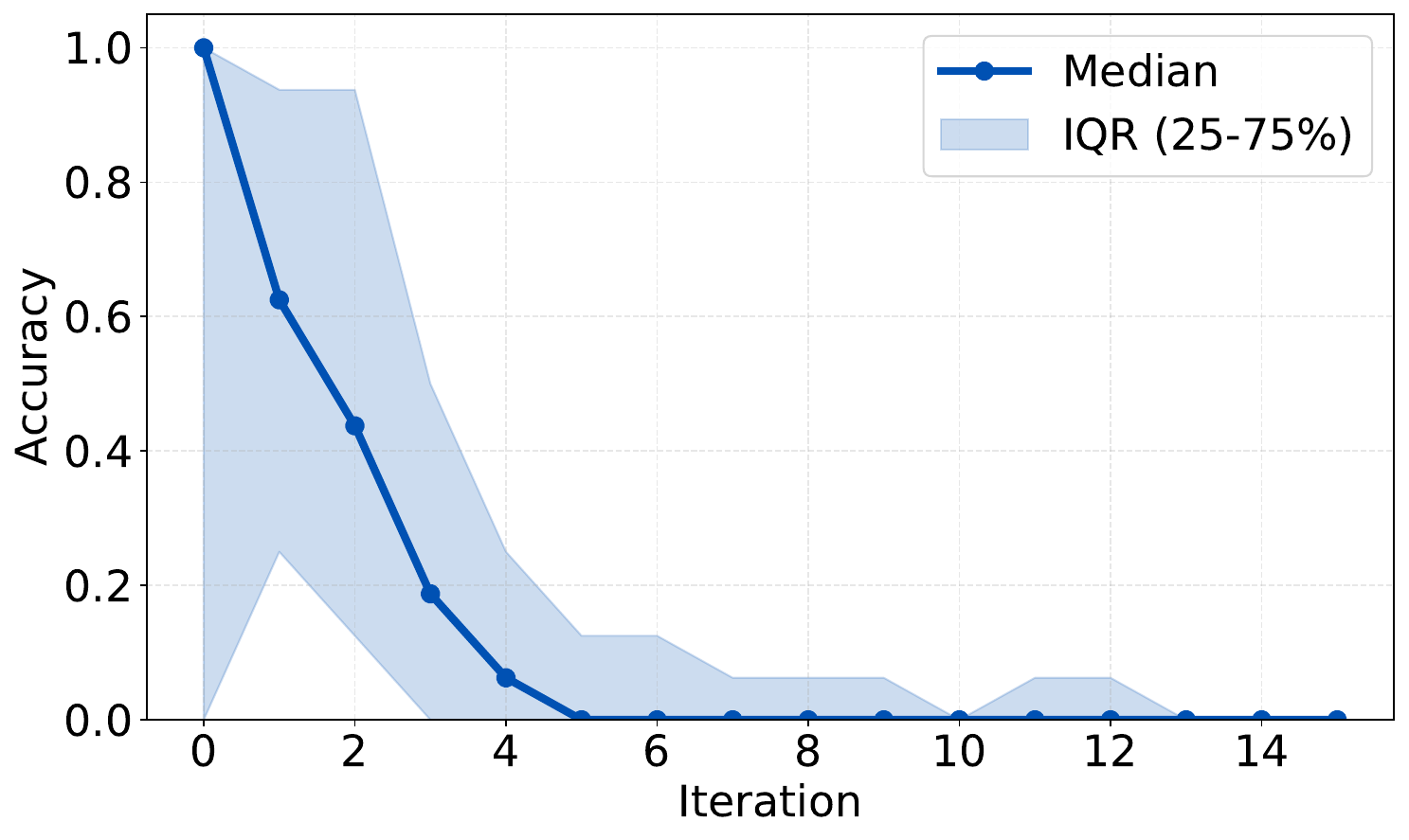}
    \caption{The model's average beam accuracy.}
    \label{fig:beam_acc_history_engtrace_qwen_7b}
  \end{subfigure}
  \caption{\textbf{Beam search finds increasingly difficult problem variations.} Across iterations, variation difficulty increases (a), while \emph{Qwen-2.5-7B-Instruct’s} beam accuracy decreases (b). Lines show medians and interquartile ranges across runs over all \emph{EngTrace} templates.}
  \label{fig:beam_search_engtrace_qwen_7b}
\end{figure*}

As described in Section~\ref{sec:lpds}, we identify increasingly difficult problem variations by running the discrete prompt optimization procedure introduced in Algorithm~\ref{alg:beam_search} for $T = 15$ iterations. The algorithm is based on beam search: at each iteration, each of the $w = 16$ active nodes (problem variations) in the current beam is expanded into a set of $b = 16$ candidate variations, and the top-$w$ variations with the highest estimated difficulty $\text{MD}_{\mathcal{H}}$ form the updated beam. As the search progresses, the difficulty of the problem variations in the beam $(p_i, \cdot) \in \mathcal{B}$ increases, while the model's accuracy on these variations decreases. This trend can be observed across different models and datasets. Figure~\ref{fig:beam_search_gsms_qwen_7b} presents results for Qwen-2.5-7B-Instruct and templates from GSM-Symbolic; Figure~\ref{fig:beam_search_gsms_qwen_32b} shows a similar trend for Qwen-2.5-32B-Instruct on GSM-Symbolic; Figure~\ref{fig:beam_search_finchain_llama_70b} illustrates results for Llama-3.1-70B-Instruct and FinChain templates; and Figure~\ref{fig:beam_search_engtrace_qwen_7b} shows results for Qwen-2.5-7B-Instruct on templates from EngTrace. Notably, upon completion of the search, the models' accuracy on the problem variations in the final beam is $0\%$ (median, averaged over all templates).

\begin{figure*}[tbp]
  \centering
  \begin{subfigure}{0.475\textwidth}
    \centering
    \includegraphics[width=\linewidth]{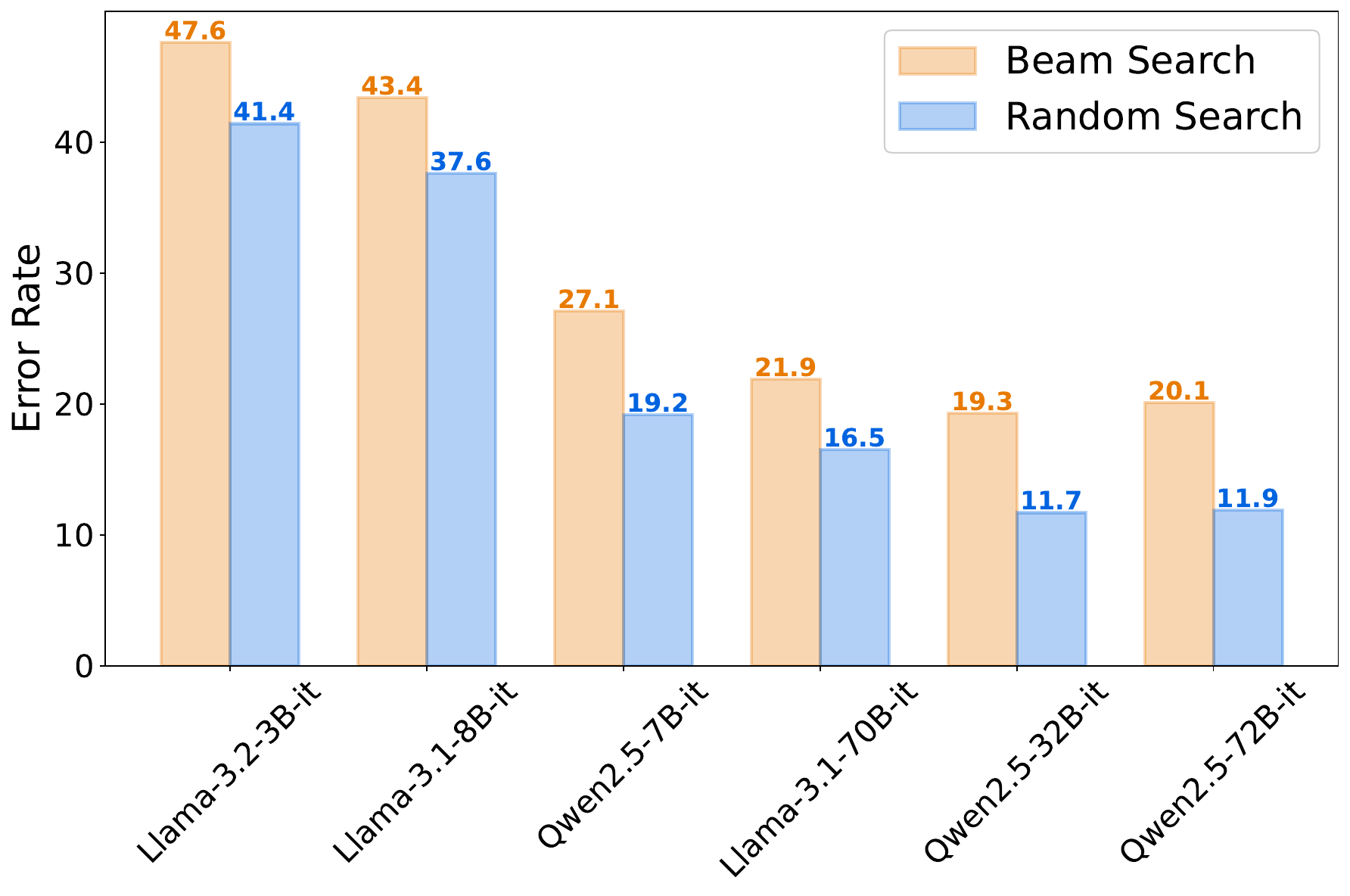}
    \caption{FinChain.}
    \label{fig:error_rates_finchain}
  \end{subfigure}
  \hspace{0.025\textwidth}
  \begin{subfigure}{0.475\textwidth}
    \centering
    \includegraphics[width=\linewidth]{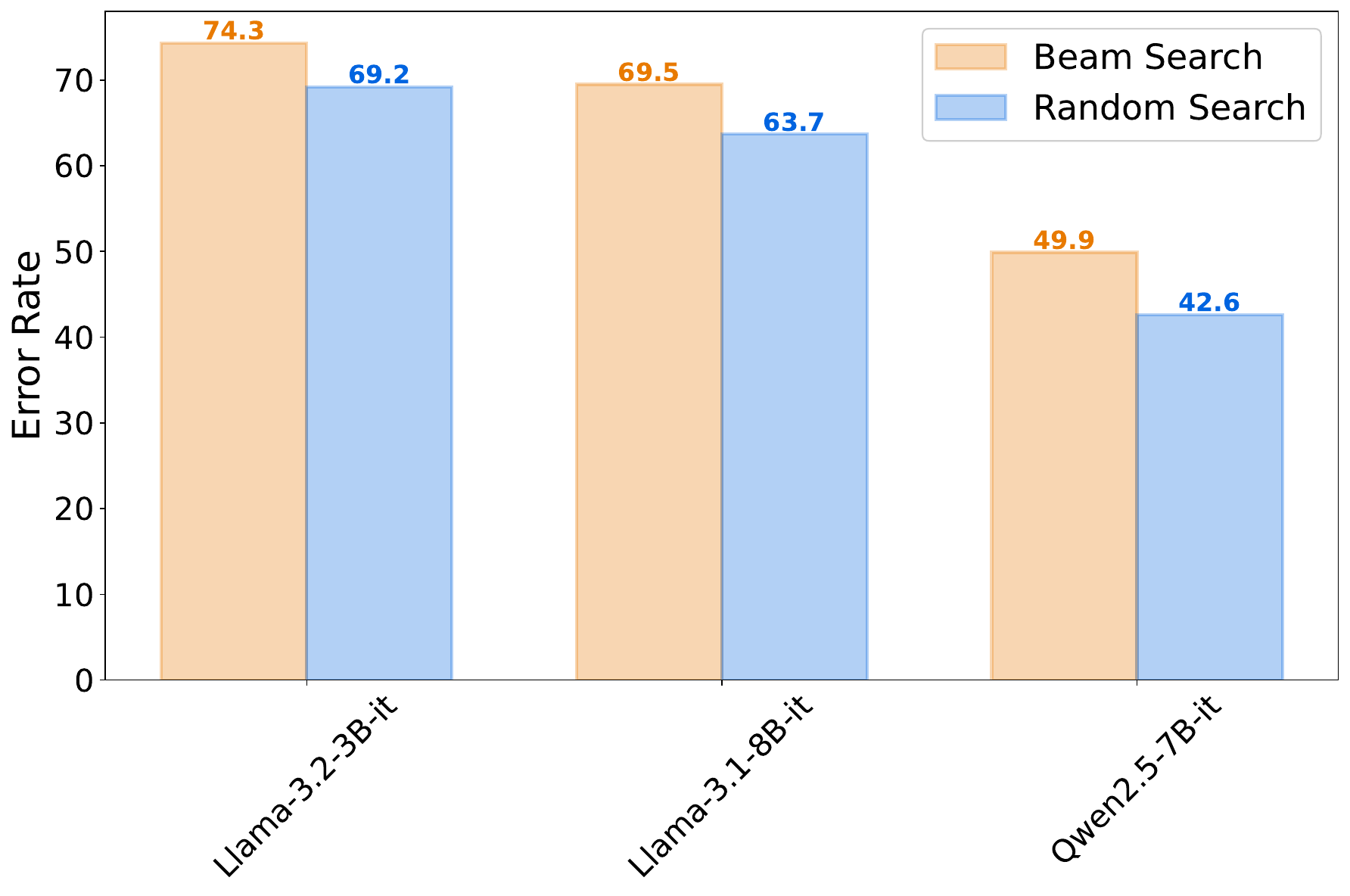}
    \caption{EngTrace.}
    \label{fig:error_rates_engtrace}
  \end{subfigure}
  \caption{\textbf{Beam search consistently yields higher error rates (\%) than random sampling.} Error rates of models for samples from FinChain and EngTrace across search strategies.}
  \label{fig:error_rates}
\end{figure*}

\subsubsection{Exposing model failures}\label{app:evaluating_robustness}
Following the experiments on GSM-Symbolic described in Section~\ref{subsec:lpds_expose_failure} (Figure~\ref{fig:beam_search_vs_random_search_gsms}), we evaluate the robustness of different models to problem variations from FinChain and EngTrace. Since FinChain and EngTrace are not derived from an existing base benchmark, we measure robustness using the models' error rates ($1 - \text{accuracy}$) rather than a performance drop. Analogous to the experiments described in Section~\ref{subsec:lpds_expose_failure}, we evaluate models on the problem variations explored during beam search and compare the resulting performance to that on an equal number of problem variations randomly sampled from a given template. Results for different models on FinChain and EngTrace are shown in Figure~\ref{fig:error_rates}. Across all models, error rates are consistently higher on problem variations identified via beam search than on randomly sampled ones. For example, on FinChain, Qwen-2.5-72B-Instruct reaches an error rate of $20.1\%$ on beam-search-selected variations, compared to $11.9\%$ on randomly sampled variations. We further observe that error rates are higher on both FinChain and EngTrace than on GSM-Symbolic, indicating that these datasets contain more challenging templates. In particular, EngTrace appears especially challenging, with error rates of up to $74.3\%$ for Llama-3.2-3B-Instruct on variations identified via beam search.

\begin{figure*}[tbp]
  \centering
  \begin{subfigure}[t]{0.31\textwidth}
    \centering
    \includegraphics[width=\linewidth]{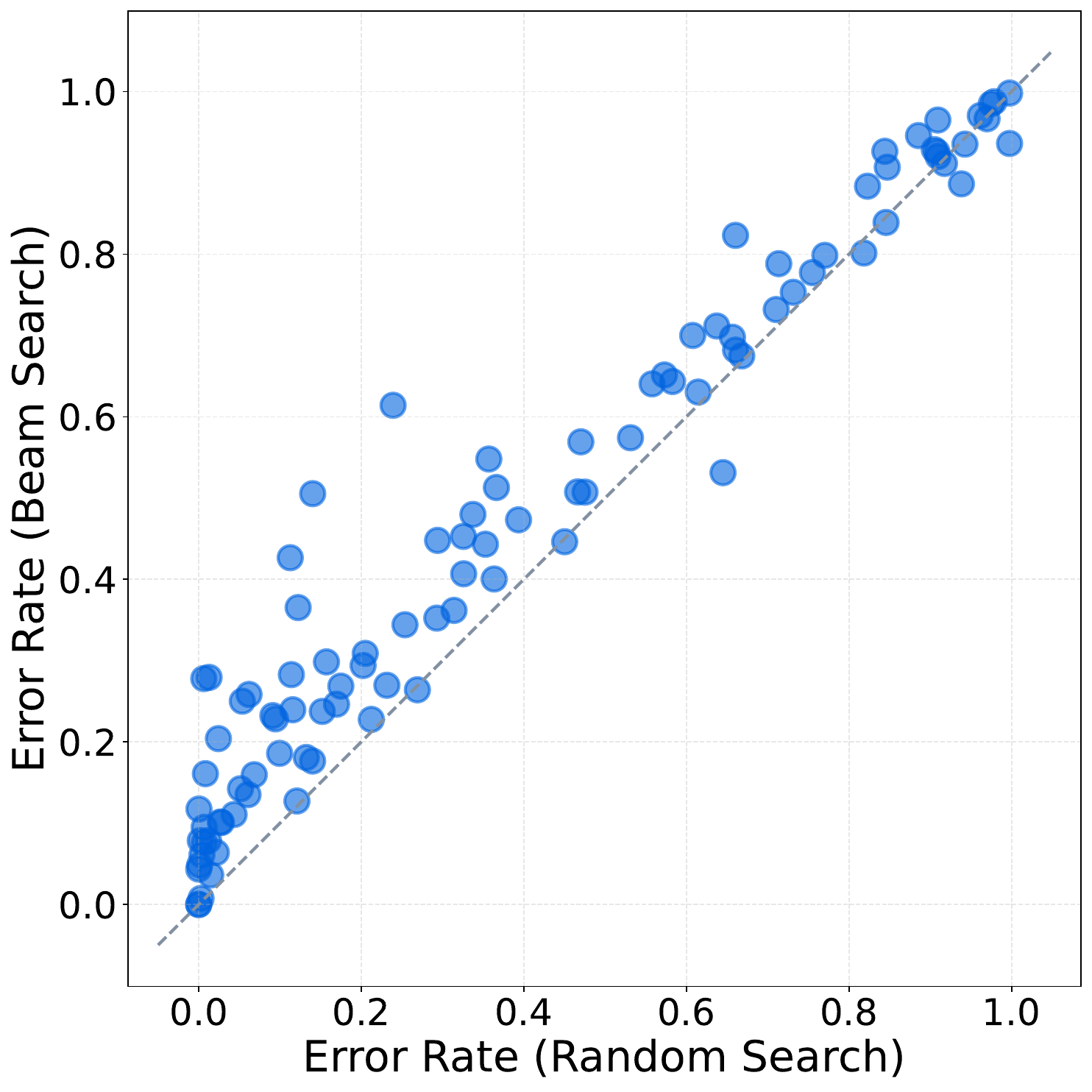}
    \caption{Llama-3.2-3B-it}
    \label{fig:error_rate_comparison_gsms_llama_3b}
  \end{subfigure}
  \hspace{0.01\textwidth}
  \begin{subfigure}[t]{0.31\textwidth}
    \centering
    \includegraphics[width=\linewidth]{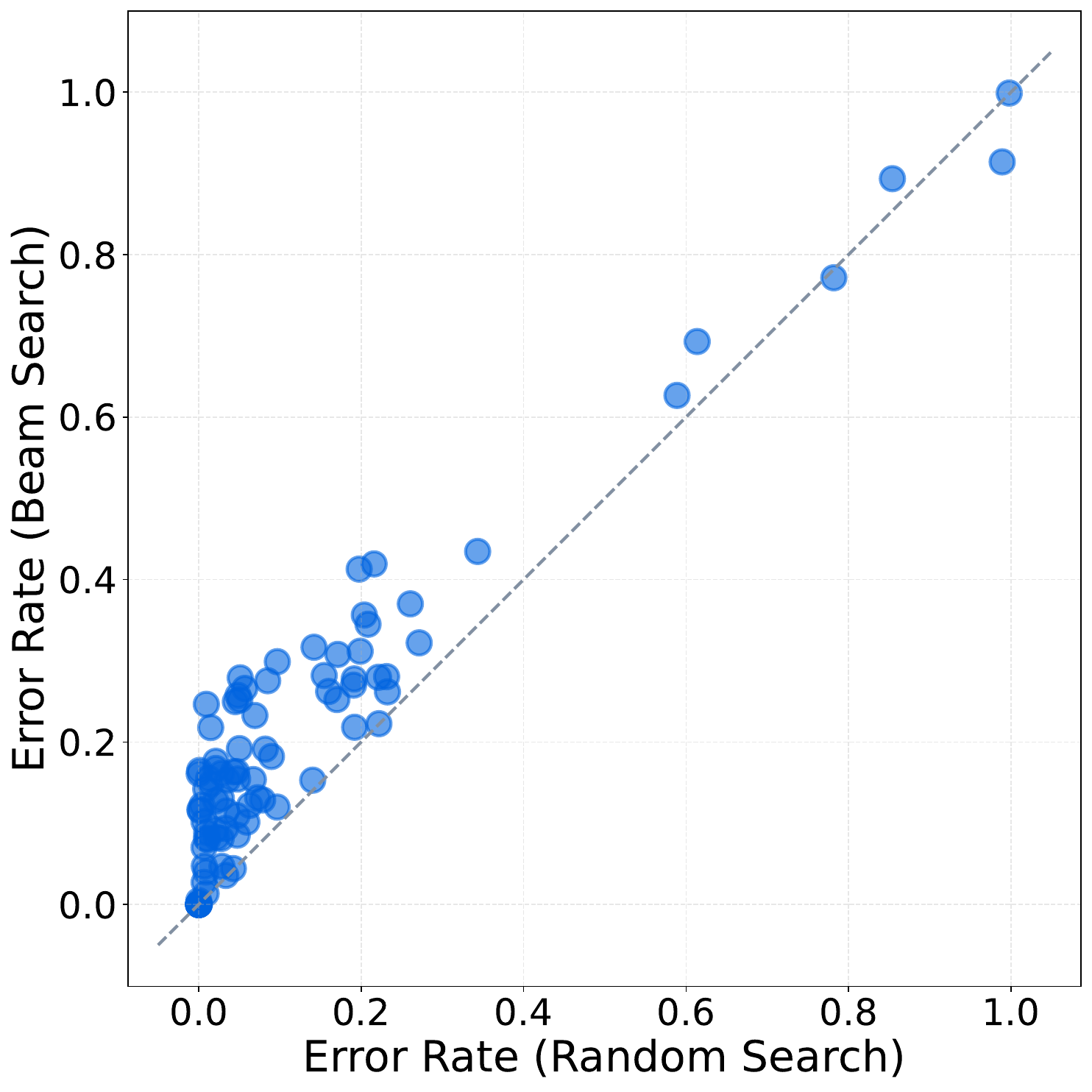}
    \caption{Qwen-2.5-7B-it}
    \label{fig:error_rate_comparison_gsms_qwen_7b}
  \end{subfigure}
  \hspace{0.01\textwidth}
  \begin{subfigure}[t]{0.31\textwidth}
    \centering
    \includegraphics[width=\linewidth]{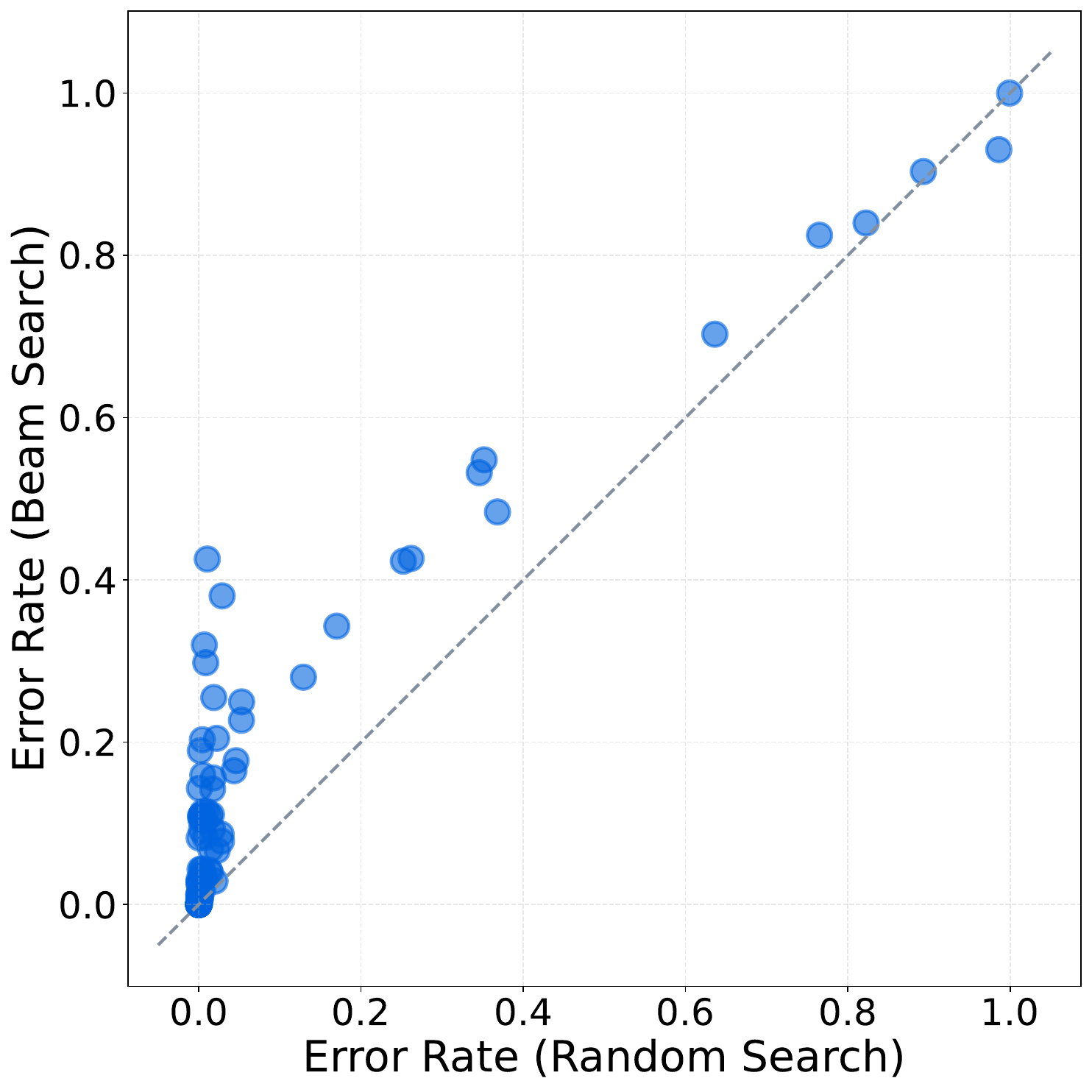}
    \caption{Qwen-2.5-72B-it}
    \label{fig:error_rate_comparison_gsms_qwen_72b}
  \end{subfigure}
  \caption{\textbf{Error rate comparison.} Models' error rates per \emph{GSM-Symbolic} template (beam search vs.\ random search).}
  \label{fig:error_rate_comparison_gsms}
\end{figure*}

\begin{figure*}[tbp]
  \centering
  \begin{subfigure}[t]{0.31\textwidth}
    \centering
    \includegraphics[width=\linewidth]{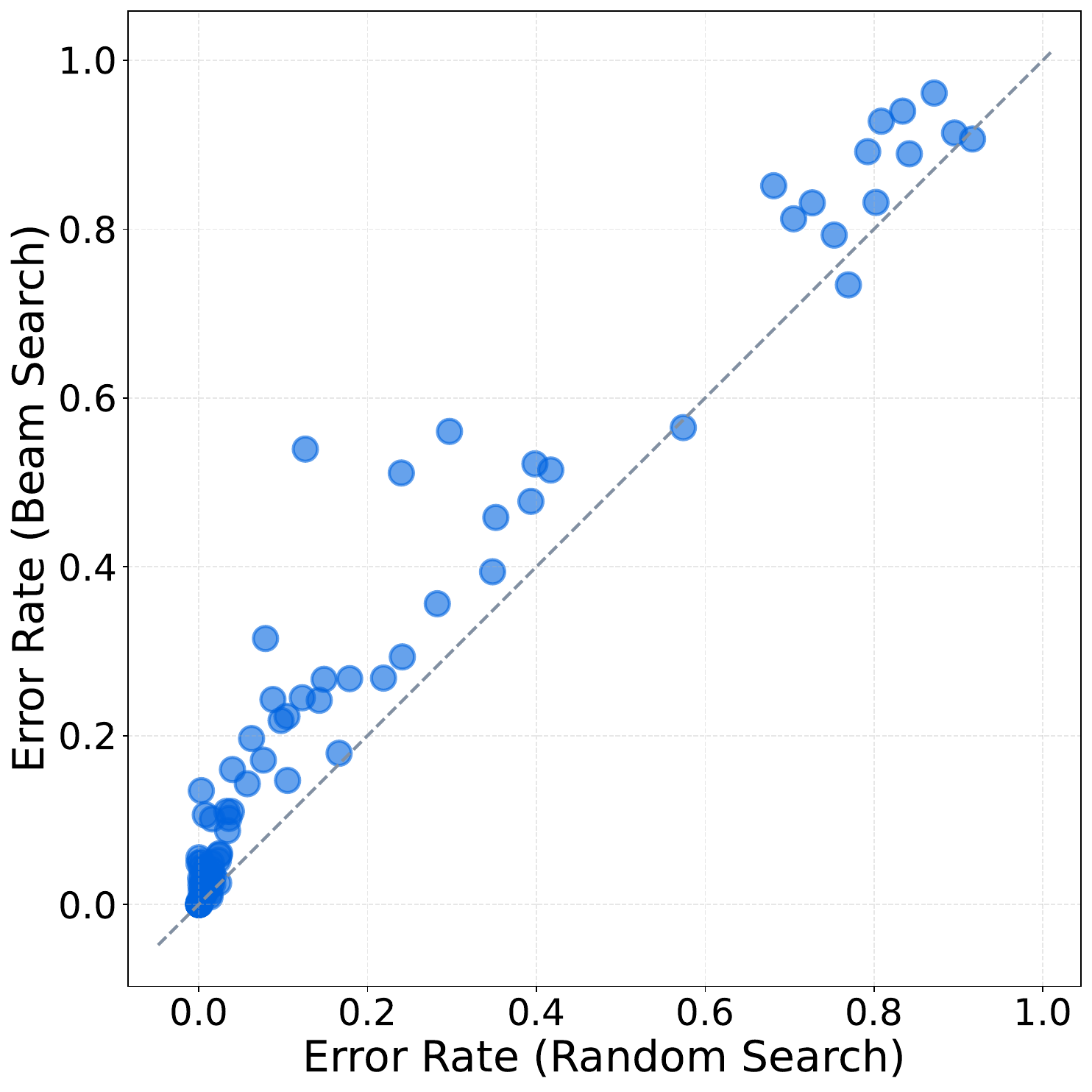}
    \caption{Llama-3.2-70B-it}
    \label{fig:error_rate_comparison_finchain_llama_70b}
  \end{subfigure}
  \hspace{0.01\textwidth}
  \begin{subfigure}[t]{0.31\textwidth}
    \centering
    \includegraphics[width=\linewidth]{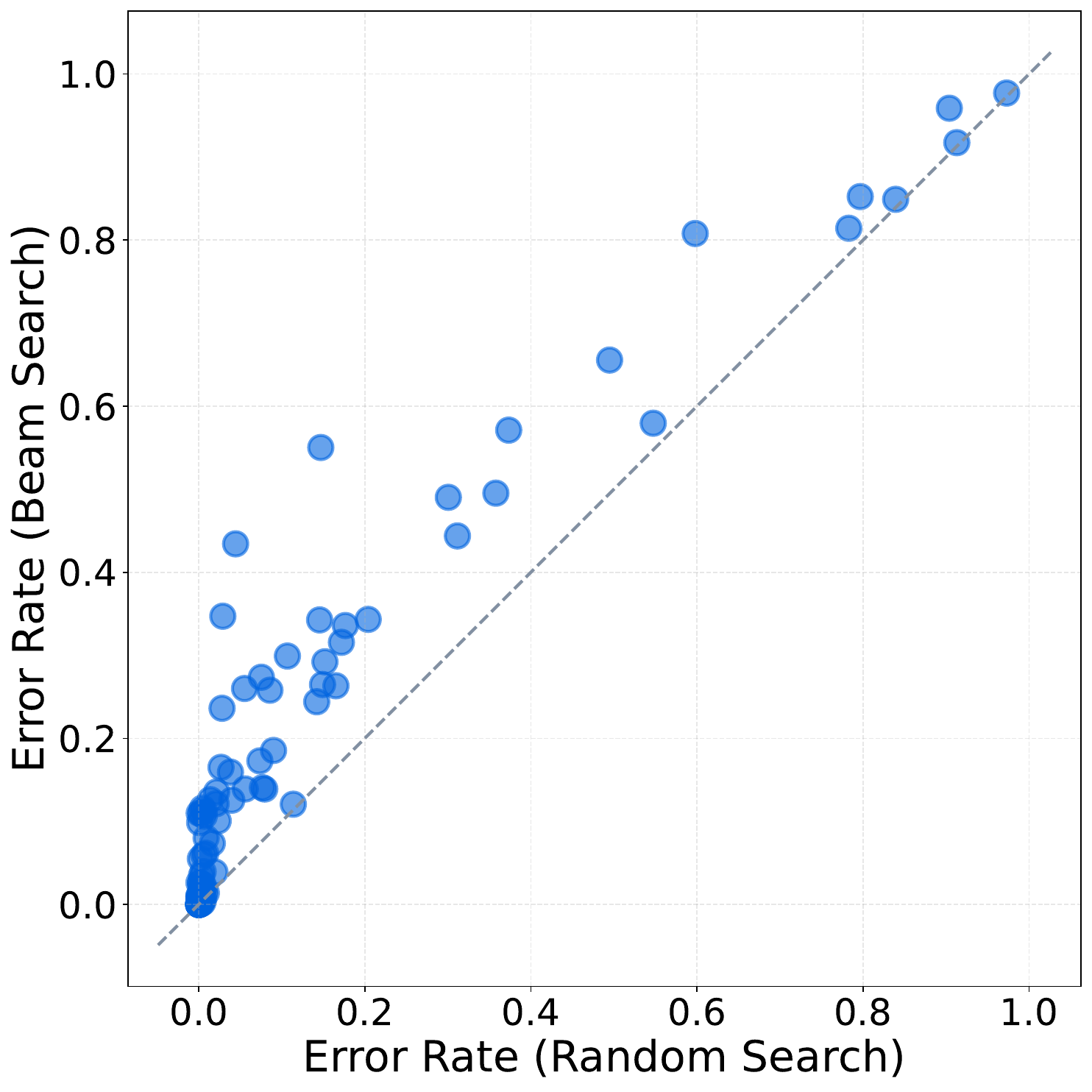}
    \caption{Qwen-2.5-32B-it}
    \label{fig:error_rate_comparison_finchain_qwen_32b}
  \end{subfigure}
  \hspace{0.01\textwidth}
  \begin{subfigure}[t]{0.31\textwidth}
    \centering
    \includegraphics[width=\linewidth]{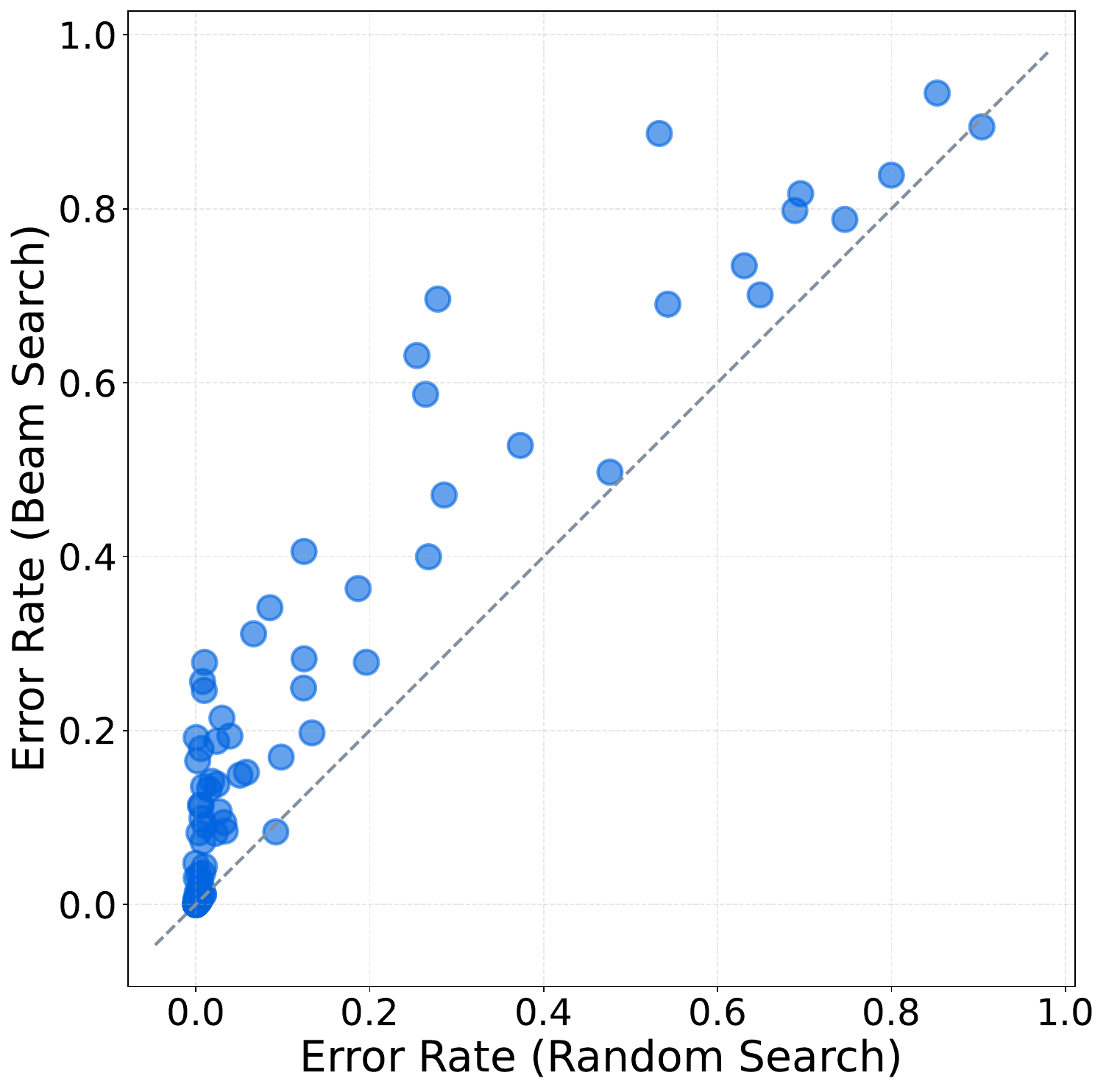}
    \caption{Qwen-2.5-72B-it}
    \label{fig:error_rate_comparison_finchain_qwen_72b}
  \end{subfigure}
  \caption{\textbf{Error rate comparison.} Models' error rates per \emph{FinChain} template (beam search vs.\ random search).}
  \label{fig:error_rate_comparison_finchain}
\end{figure*}

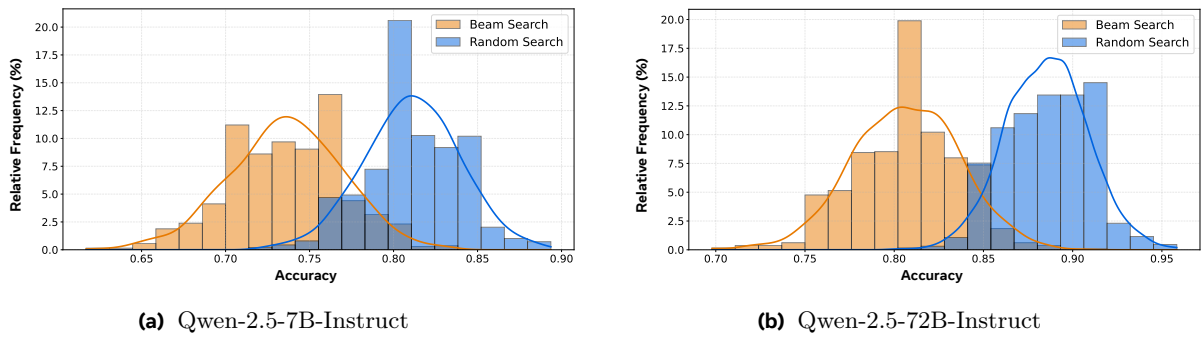
\begin{figure*}[tbp]
  \centering
  \begin{subfigure}{0.435\textwidth}
    \centering
    \begin{tikzpicture}
        \node[inner sep=0] (img) at (0,0) {%
            \includegraphics[
                width=\linewidth,
                trim={1.cm 1.cm 0.cm 0.cm},
                clip
            ]{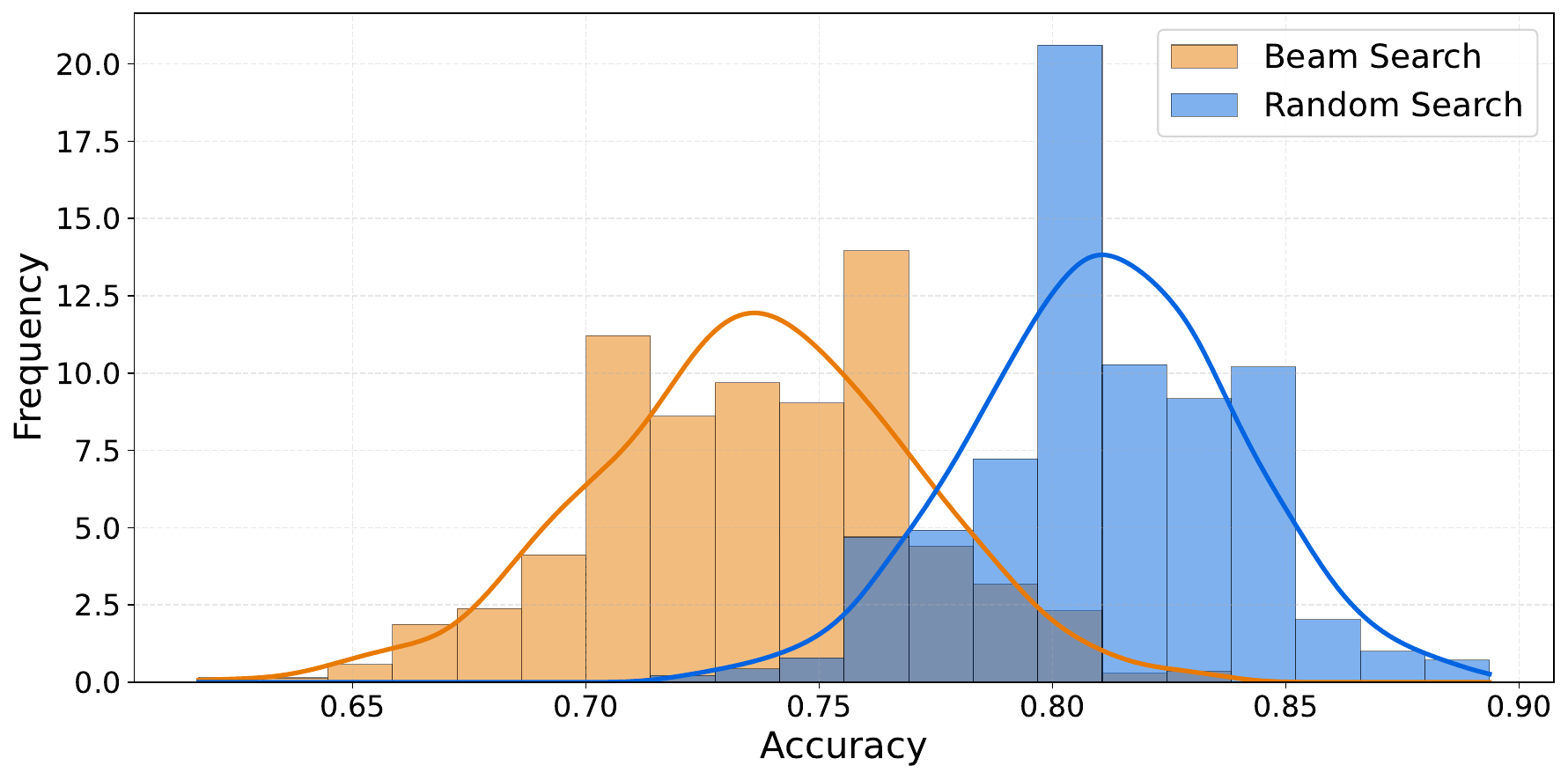}
            };
        \node[
            rotate=90,
            font=\fontsize{6pt}{7pt}\selectfont\sffamily,
            anchor=south
        ] at ([xshift=0.005cm]img.west) {Relative Frequency (\%)};
        \node[
            font=\fontsize{6pt}{7pt}\selectfont\sffamily,
            anchor=north,
        ] at ([yshift=0.075cm]img.south) {Accuracy};
    \end{tikzpicture}
    \caption{Qwen-2.5-7B-Instruct}
    \label{fig:performance_variance_finchain_qwen_7b}
  \end{subfigure}
  \hspace{0.05\textwidth}
  \begin{subfigure}{0.435\textwidth}
    \centering
    \begin{tikzpicture}
        \node[inner sep=0] (img) at (0,0) {%
            \includegraphics[
                width=\linewidth,
                trim={1.cm 1.cm 0.cm 0.cm},
                clip
            ]{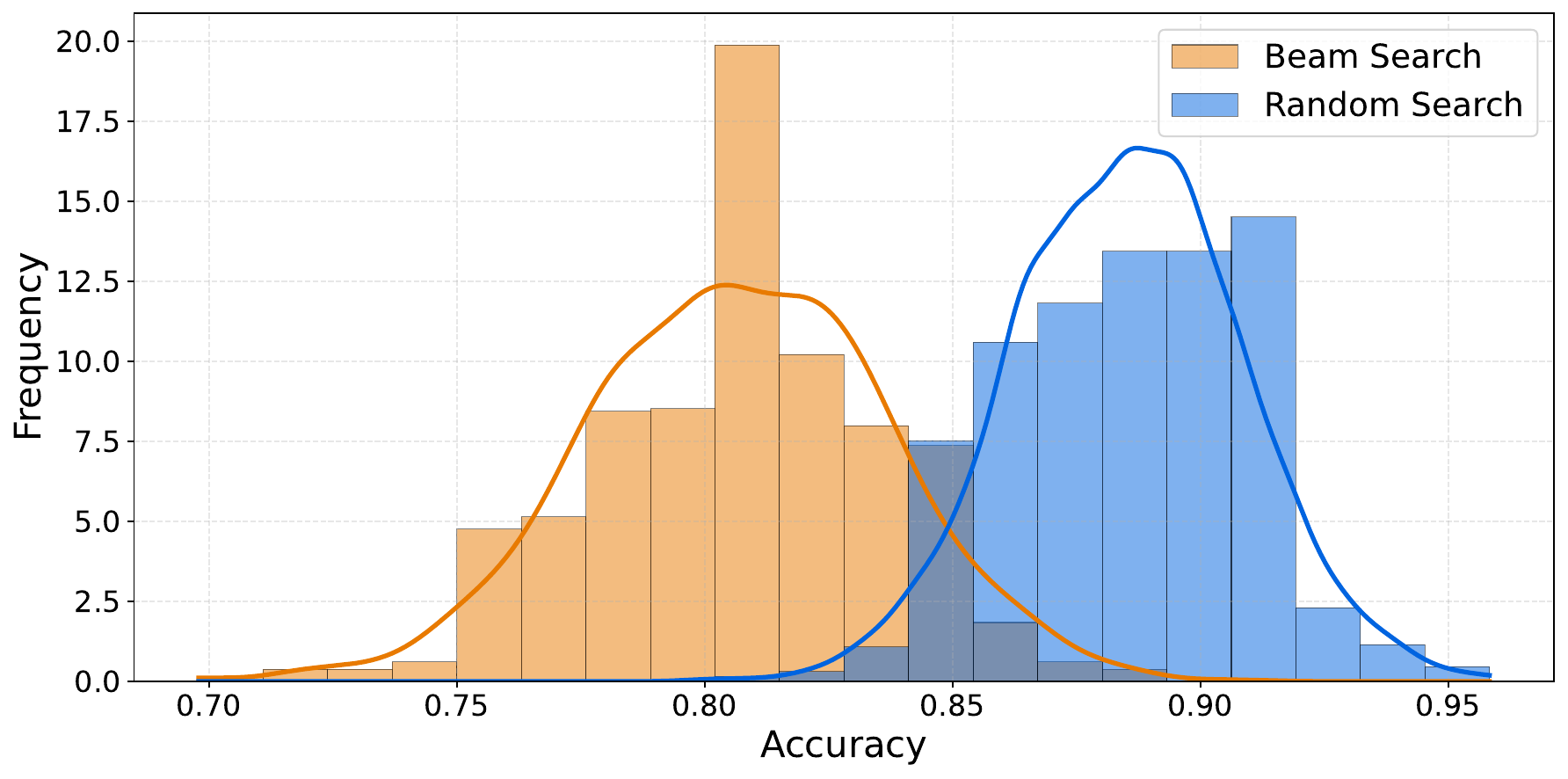}
        };
        \node[
            rotate=90,
            font=\fontsize{6pt}{7pt}\selectfont\sffamily,
            anchor=south
        ] at ([xshift=0.005cm]img.west) {Relative Frequency (\%)};
        \node[
            font=\fontsize{6pt}{7pt}\selectfont\sffamily,
            anchor=north,
        ] at ([yshift=0.075cm]img.south) {Accuracy};
    \end{tikzpicture}
    \caption{Qwen-2.5-72B-Instruct}
    \label{fig:performance_variance_finchain_qwen_72b}
    \end{subfigure}
  \caption{\textbf{Empirical accuracy distributions.} Problem variations are identified from \emph{FinChain} via beam vs.\ random search.}
  \label{fig:performance_variance_finchain}
\end{figure*}

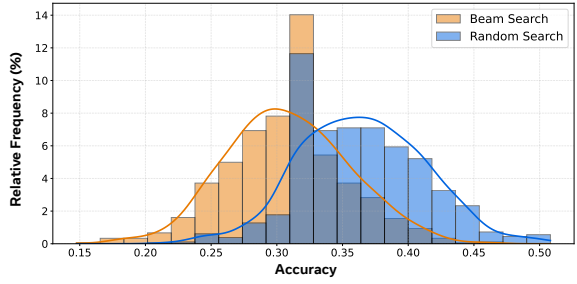
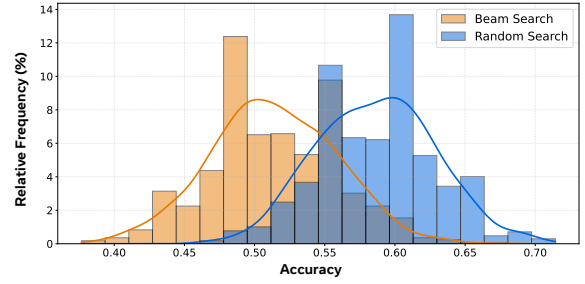
\begin{figure*}[tbp]
  \centering
  \begin{subfigure}{0.435\textwidth}
    \centering
    \begin{tikzpicture}
     \node[inner sep=0] (img) at (0,0) {%
            \includegraphics[
                width=\linewidth,
                trim={1.cm 1.cm 0.cm 0.cm},
                clip
            ]{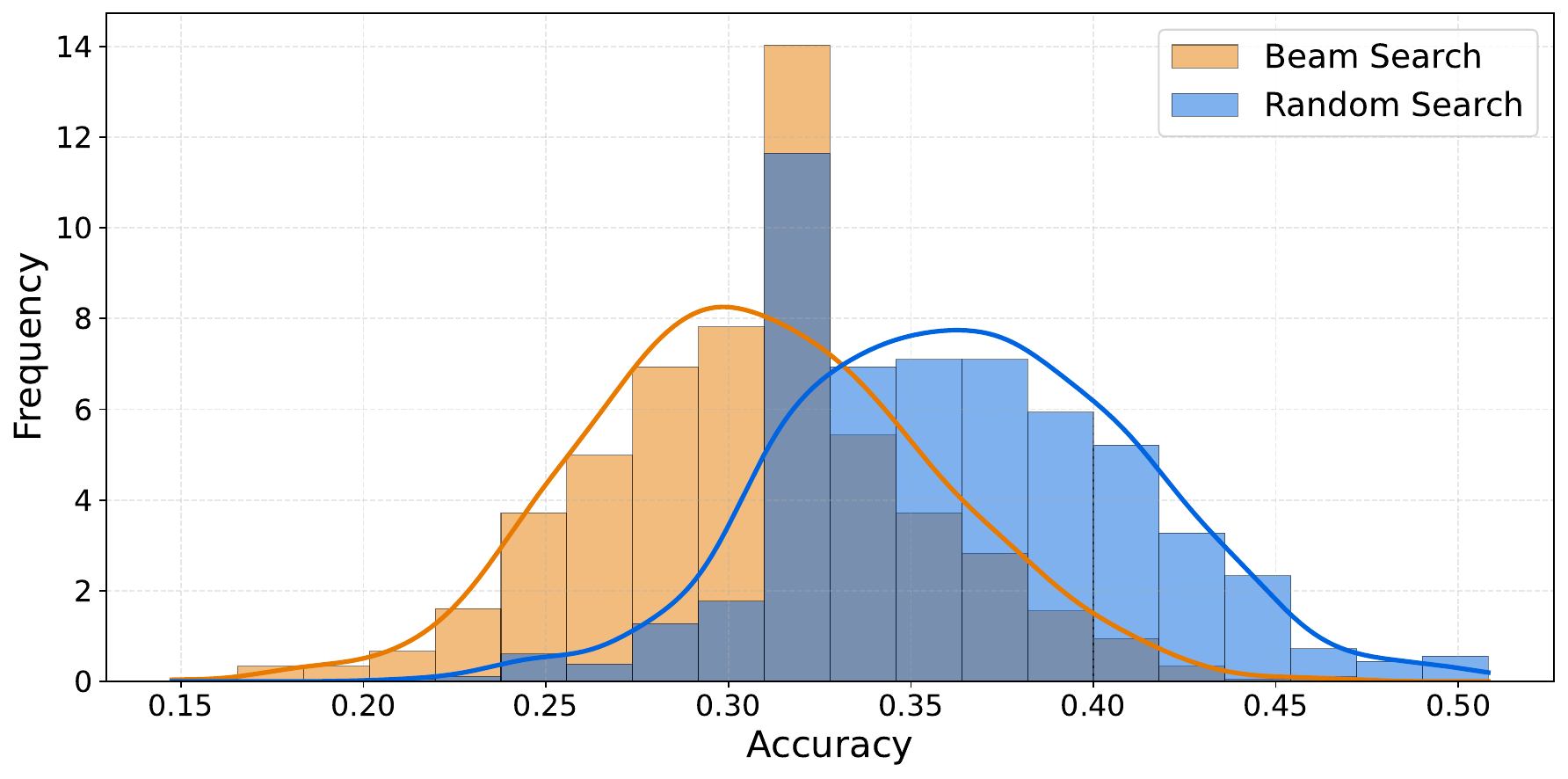}
    };
    \node[
            rotate=90,
            font=\fontsize{6pt}{7pt}\selectfont\sffamily,
            anchor=south
        ] at ([xshift=0.005cm]img.west) {Relative Frequency (\%)};
        \node[
            font=\fontsize{6pt}{7pt}\selectfont\sffamily,
            anchor=north,
        ] at ([yshift=0.075cm]img.south) {Accuracy};
    \end{tikzpicture}
    \caption{Llama-3.1-8B-Instruct}
    \label{fig:performance_variance_engtrace_llama_8b}
  \end{subfigure}
  \hspace{0.05\textwidth}
  \begin{subfigure}{0.435\textwidth}
    \centering
    \begin{tikzpicture}
     \node[inner sep=0] (img) at (0,0) {%
            \includegraphics[
                width=\linewidth,
                trim={1.cm 1.cm 0.cm 0.cm},
                clip
            ]{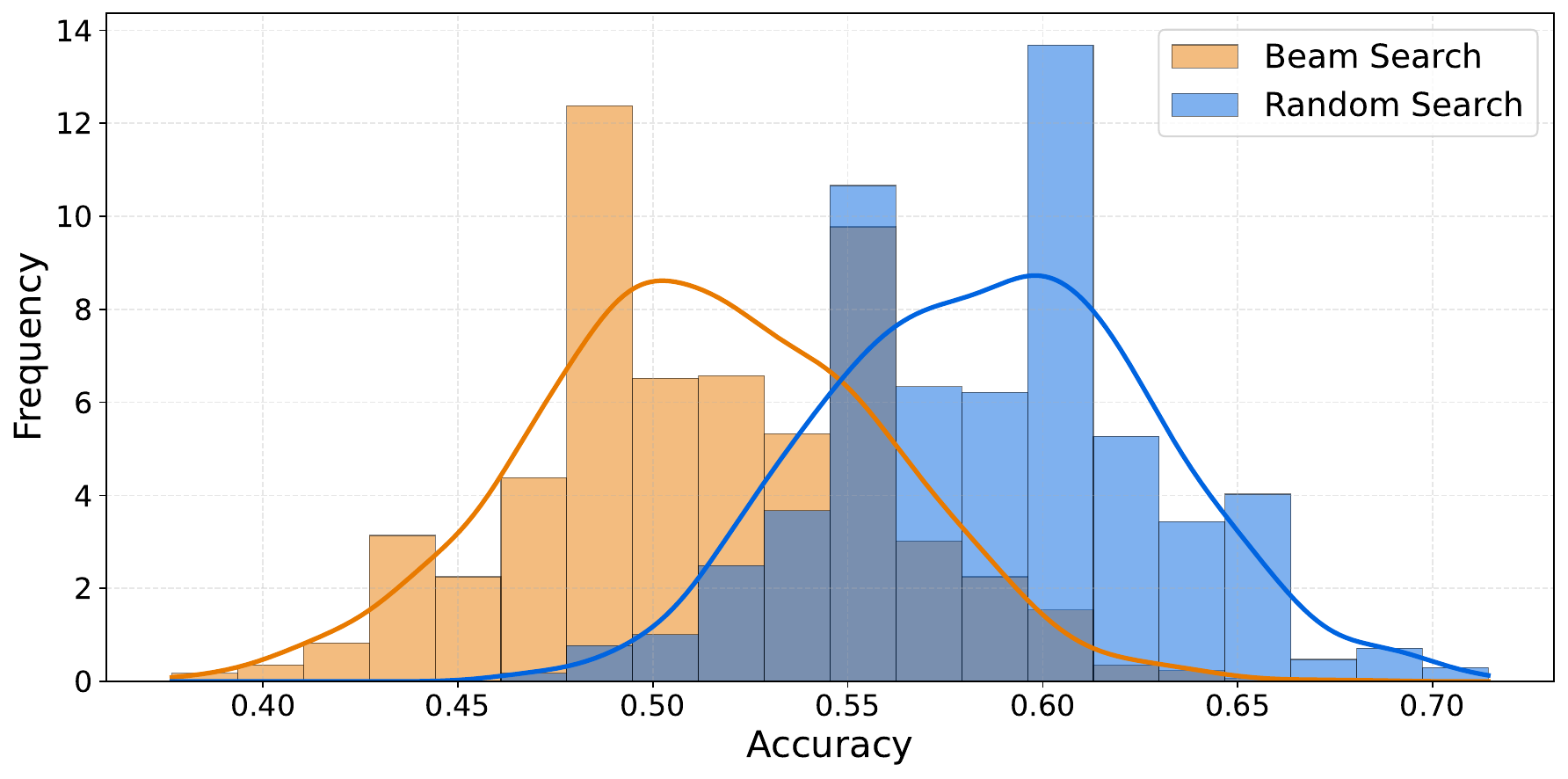}
    };
    \node[
            rotate=90,
            font=\fontsize{6pt}{7pt}\selectfont\sffamily,
            anchor=south
        ] at ([xshift=0.005cm]img.west) {Relative Frequency (\%)};
        \node[
            font=\fontsize{6pt}{7pt}\selectfont\sffamily,
            anchor=north,
        ] at ([yshift=0.075cm]img.south) {Accuracy};
    \end{tikzpicture}
    \caption{Qwen-2.5-7B-Instruct}
    \label{fig:performance_variance_engtrace_qwen_7b}
  \end{subfigure}
  \caption{\textbf{Empirical accuracy distributions.} Problem variations are identified from \emph{EngTrace} via beam vs.\ random search.}
  \label{fig:performance_variance_engtrace}
\end{figure*}

Figures~\ref{fig:error_rate_comparison_gsms} and~\ref{fig:error_rate_comparison_finchain} further show that, across models and datasets, the error rates per template are almost always higher for variations explored during beam search than for variations randomly sampled. Finally, Figures~\ref{fig:performance_variance_finchain} and~\ref{fig:performance_variance_engtrace} illustrate the variability of model performance on samples from FinChain and EngTrace, respectively. As described in Section~\ref{subsec:lpds_expose_failure}, we randomly select one problem variation from the set of samples explored during search and compute the resulting average accuracy over all templates. We repeat this procedure 1,000 times to obtain an empirical accuracy distribution for each search strategy. Across all models and datasets, we find that, for beam search, the distribution is shifted to the left compared to random search, indicating that the models are less robust than random sampling suggests.

\begin{figure*}[tbp]
  \centering
  \begin{subfigure}[t]{0.31\textwidth}
    \centering
    \includegraphics[width=\linewidth]{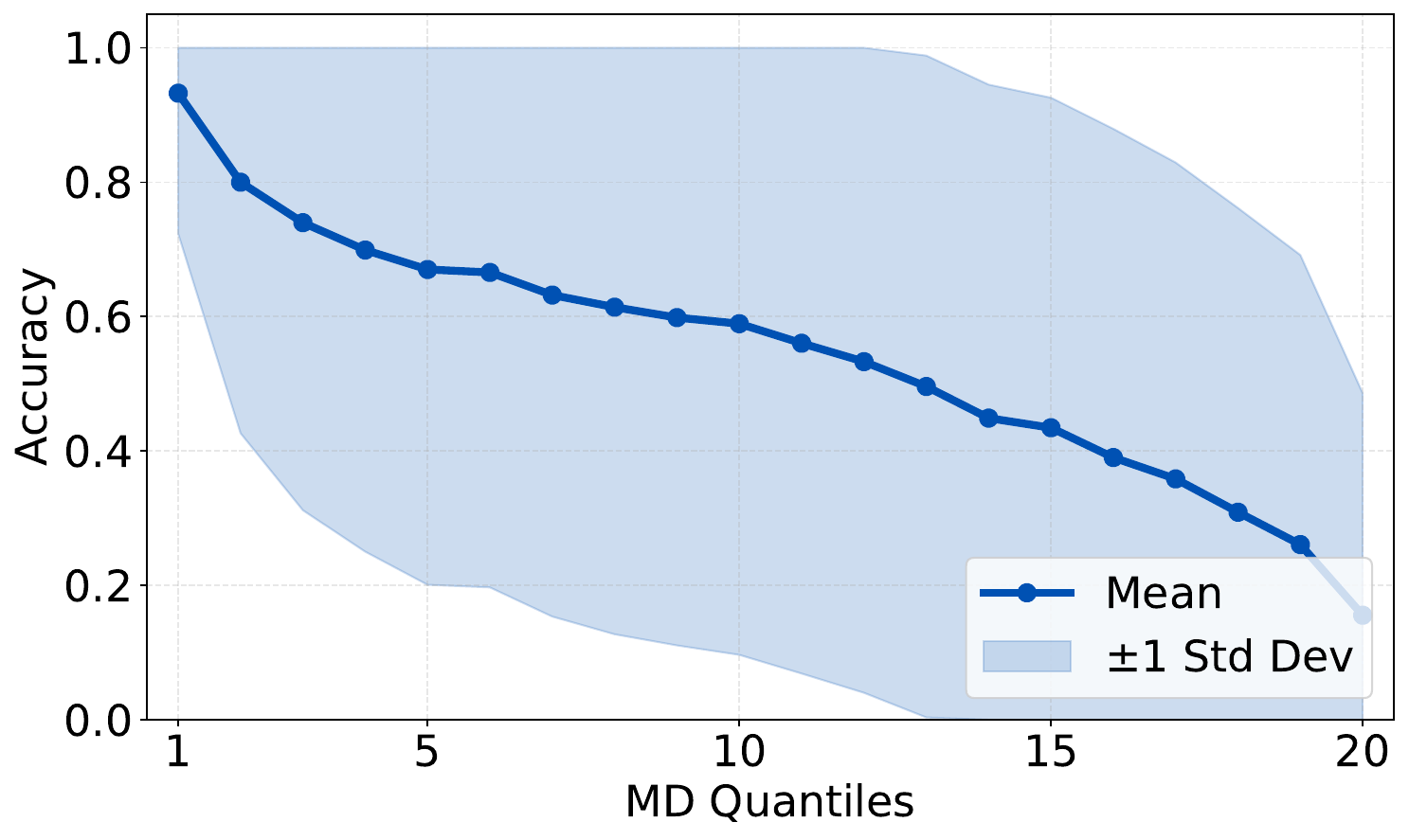}
    \caption{Llama-3.1-8B-it}
    \label{fig:reliability_plot_finchain_llama_8b}
  \end{subfigure}
  \hspace{0.01\textwidth}
  \begin{subfigure}[t]{0.31\textwidth}
    \centering
    \includegraphics[width=\linewidth]{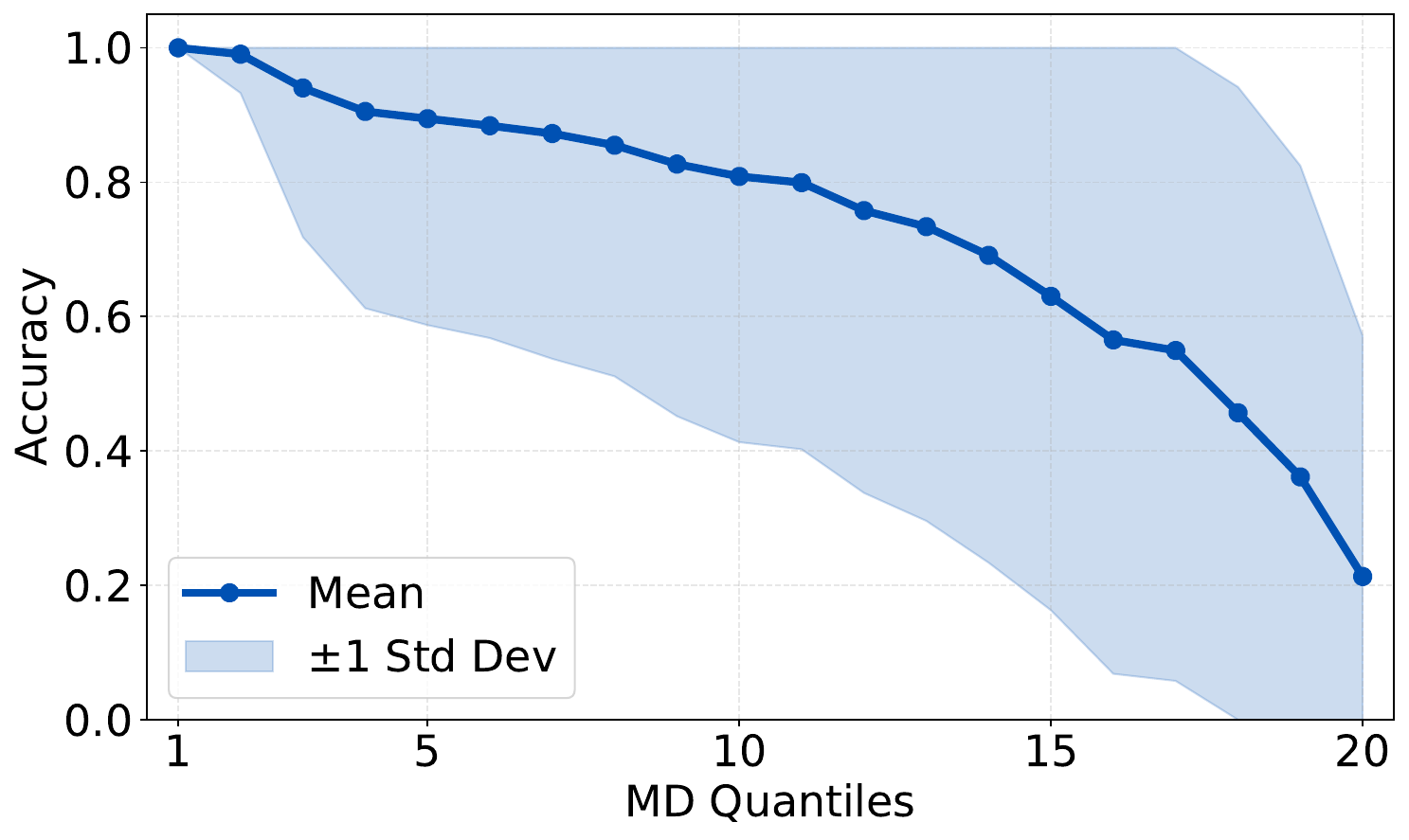}
    \caption{Qwen-2.5-7B-it}
    \label{fig:reliability_plot_finchain_qwen_7b}
  \end{subfigure}
  \hspace{0.01\textwidth}
  \begin{subfigure}[t]{0.31\textwidth}
    \centering
    \includegraphics[width=\linewidth]{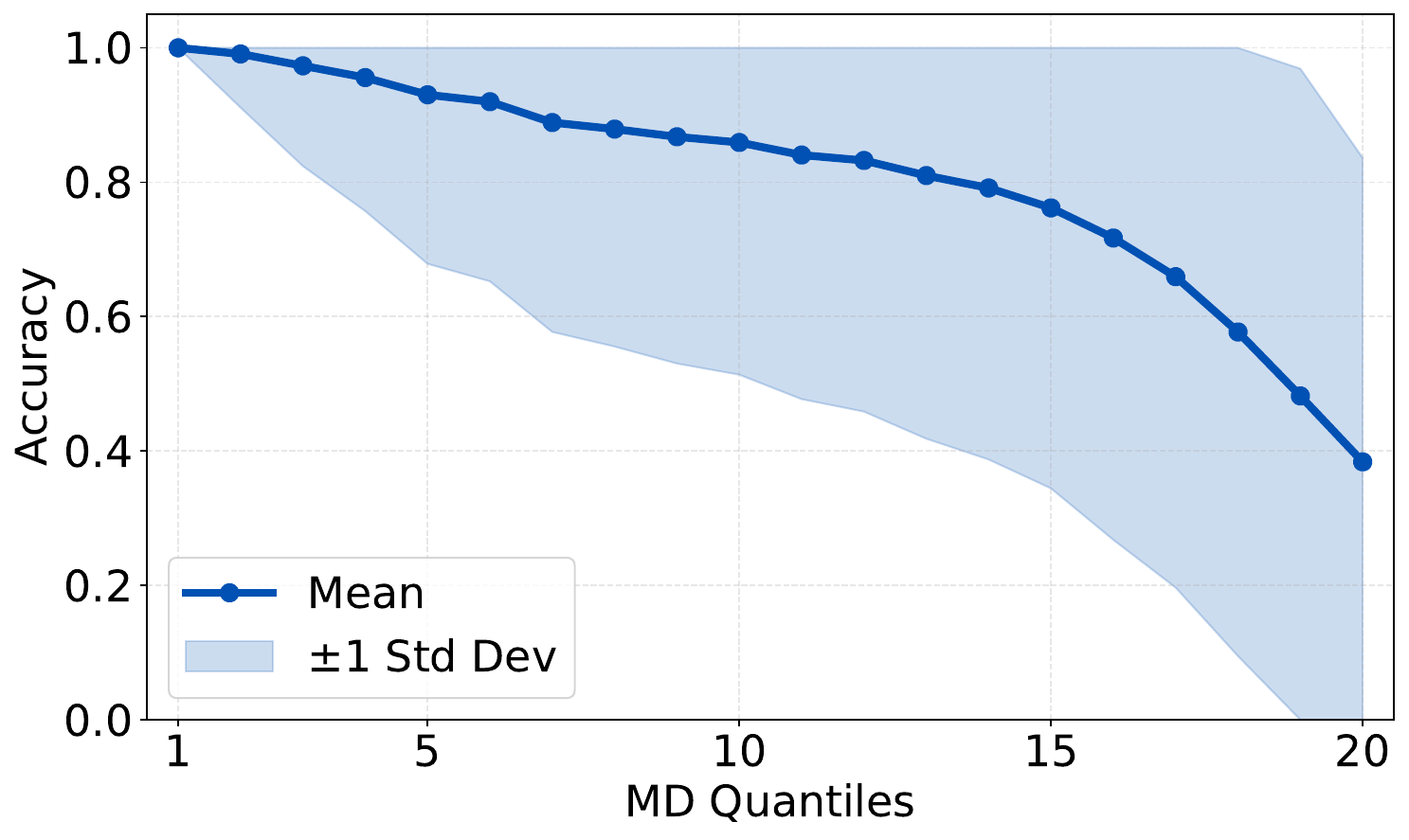}
    \caption{Qwen-2.5-72B-it}
    \label{fig:reliability_plot_finchain_qwen_72b}
  \end{subfigure}
  \caption{\textbf{Accuracy on FinChain subsets of increasing difficulty.} Variations are identified via beam search; curves show means across \emph{FinChain} templates, with standard deviation.}
  \label{fig:reliability_plot_finchain}
\end{figure*}

\begin{figure*}[tbp]
  \centering
  \begin{subfigure}[t]{0.31\textwidth}
    \centering
    \includegraphics[width=\linewidth]{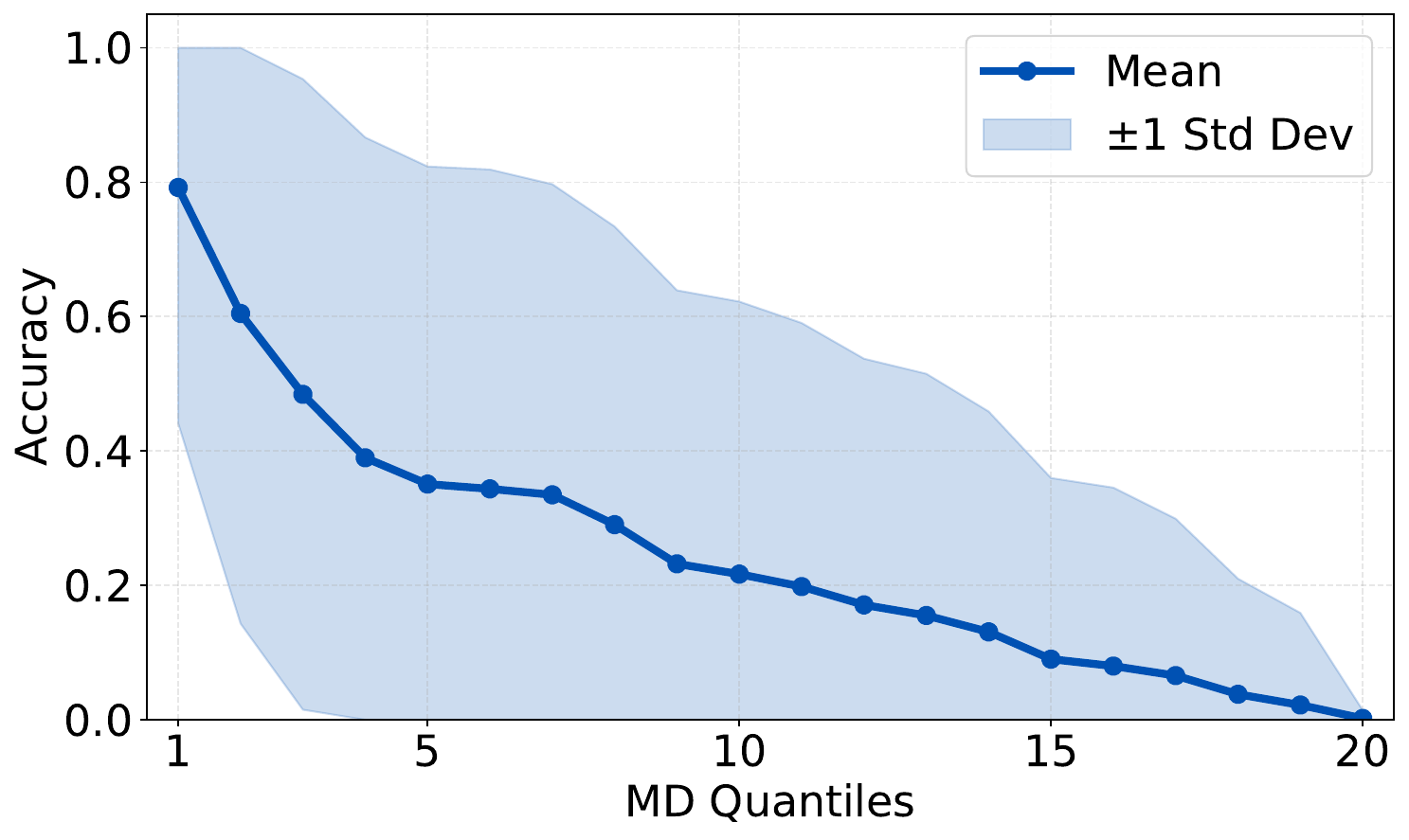}
    \caption{Llama-3.2-3B-it}
    \label{fig:reliability_plot_engtrace_llama_3b}
  \end{subfigure}
  \hspace{0.01\textwidth}
  \begin{subfigure}[t]{0.31\textwidth}
    \centering
    \includegraphics[width=\linewidth]{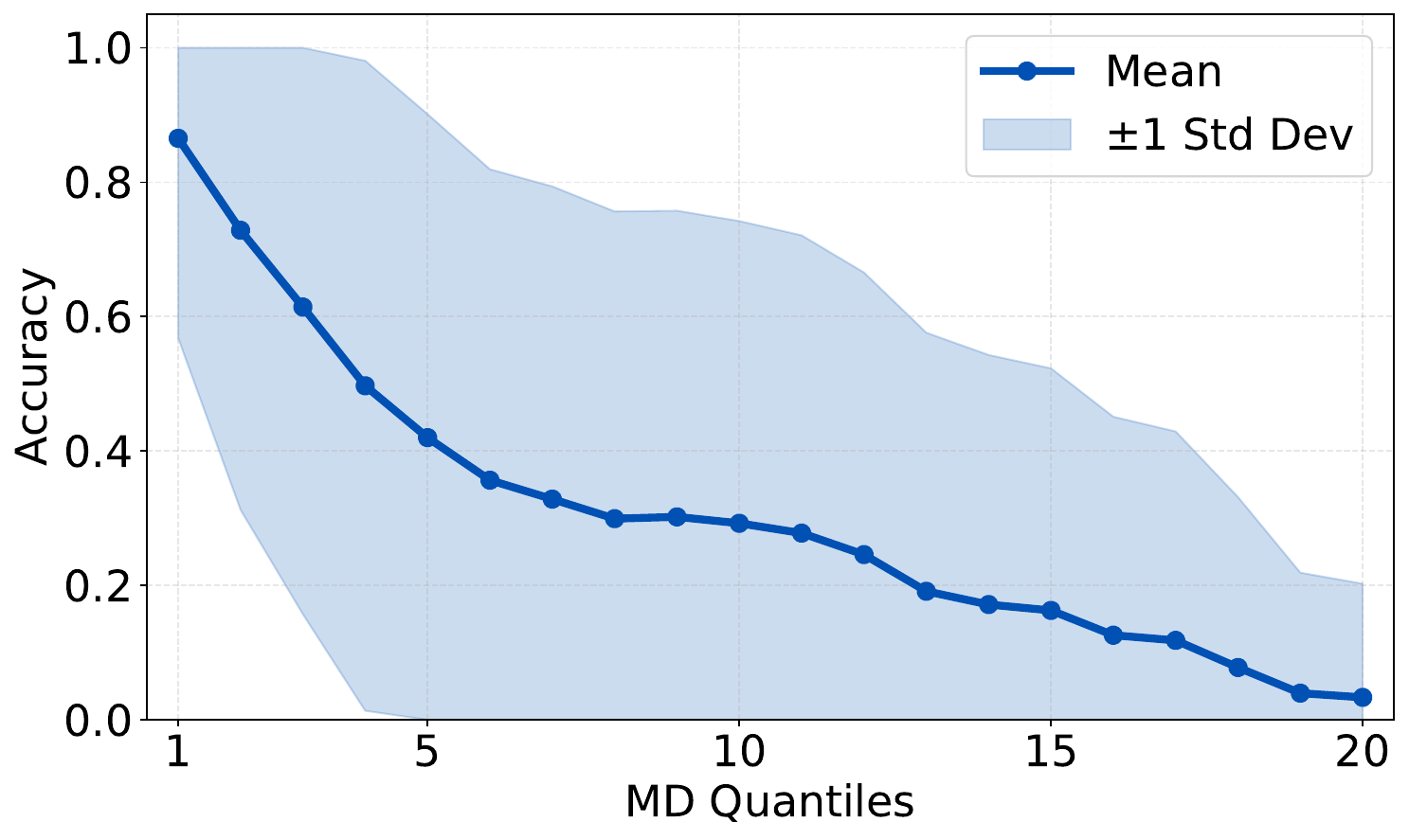}
    \caption{Llama-3.1-8B-it}
    \label{fig:reliability_plot_engtrace_llama_8b}
  \end{subfigure}
  \hspace{0.01\textwidth}
  \begin{subfigure}[t]{0.31\textwidth}
    \centering
    \includegraphics[width=\linewidth]{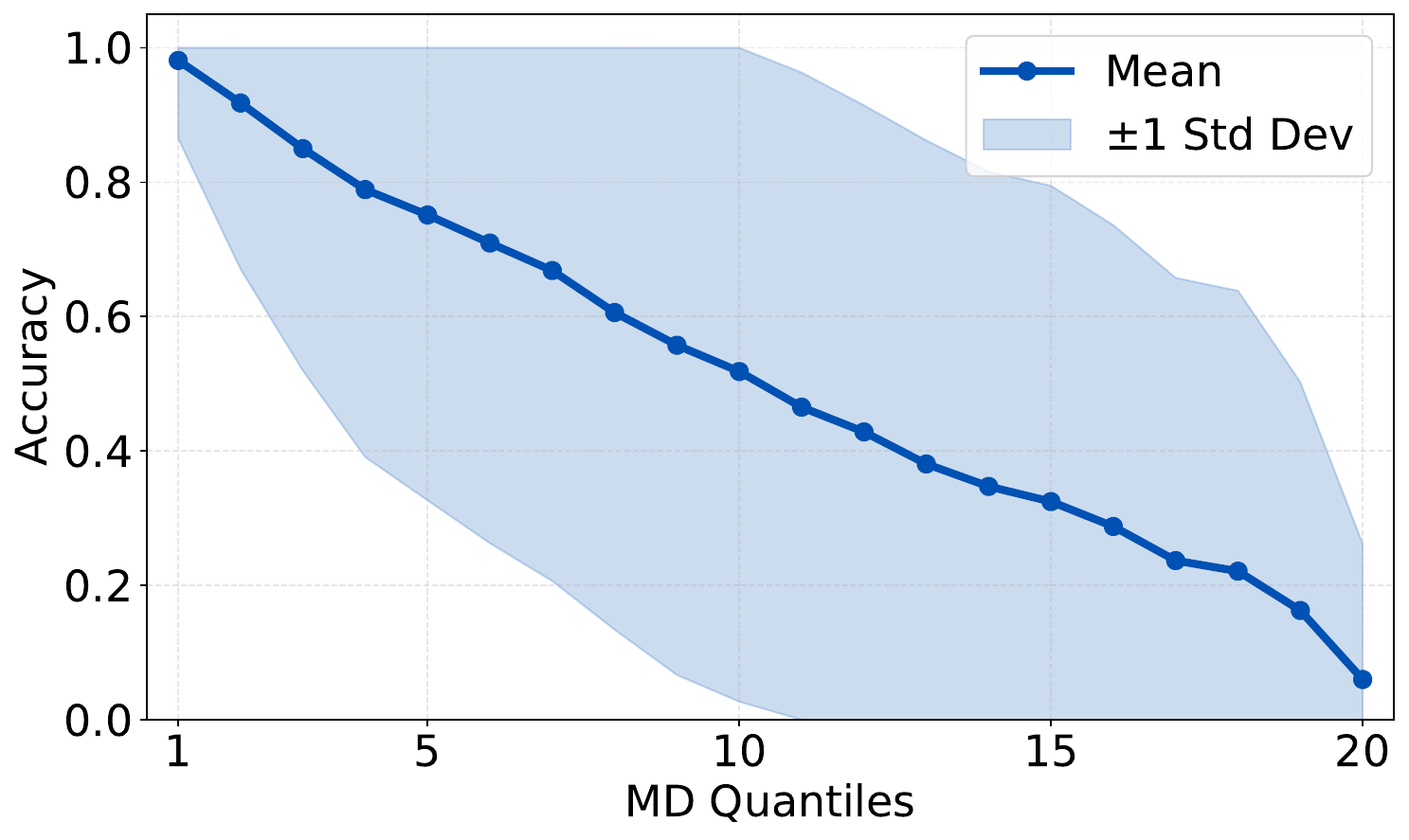}
    \caption{Qwen-2.5-7B-it}
    \label{fig:reliability_plot_engtrace_qwen_7b}
  \end{subfigure}
  \caption{\textbf{Accuracy on EngTrace subsets of increasing difficulty.} Variations are identified via beam search; curves show means across \emph{EngTrace} templates, with standard deviation.}
  \label{fig:reliability_plot_engtrace}
\end{figure*}

\paragraph{\textbf{Performance decreases as difficulty scales.}} Figures~\ref{fig:reliability_plot_finchain} and~\ref{fig:reliability_plot_engtrace} illustrate that model performance decreases as the difficulty of problem variations increases. As described in Section~\ref{subsec:lpds_expose_failure}, we group problem variations into equal-sized subsets of increasing difficulty, $\text{MD}_{\mathcal{H}}$, and evaluate model accuracy on each subset. Across models and datasets, we observe a similar trend that performance deteriorates with increasing difficulty. However, we note important differences: (i) smaller models are less robust than larger models; (ii) smaller models from the Qwen family (e.g., Qwen-2.5-7B-Instruct) appear more robust than similarly sized models from the Llama family (e.g., Llama-3.1-8B-Instruct); and (iii) performance drops fastest and most substantially for problem variations from EngTrace. 

\begin{figure*}[tbp]
  \centering
  \begin{subfigure}[t]{0.31\textwidth}
    \centering
    \includegraphics[width=\linewidth]{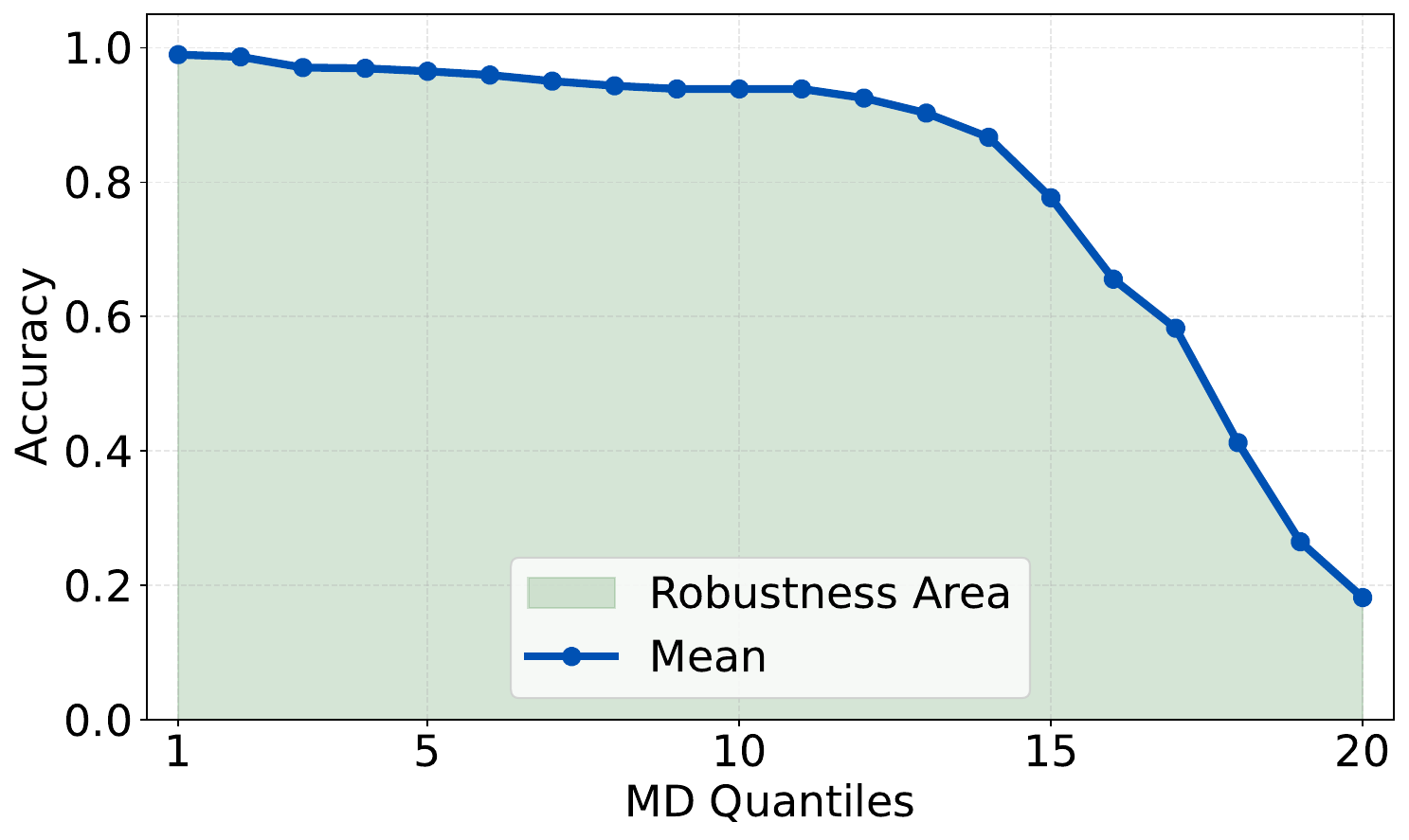}
    \caption{Qwen-2.5-7B-it}
    \label{fig:reliability_area_gsms_qwen_7b}
  \end{subfigure}
  \hspace{0.01\textwidth}
  \begin{subfigure}[t]{0.31\textwidth}
    \centering
    \includegraphics[width=\linewidth]{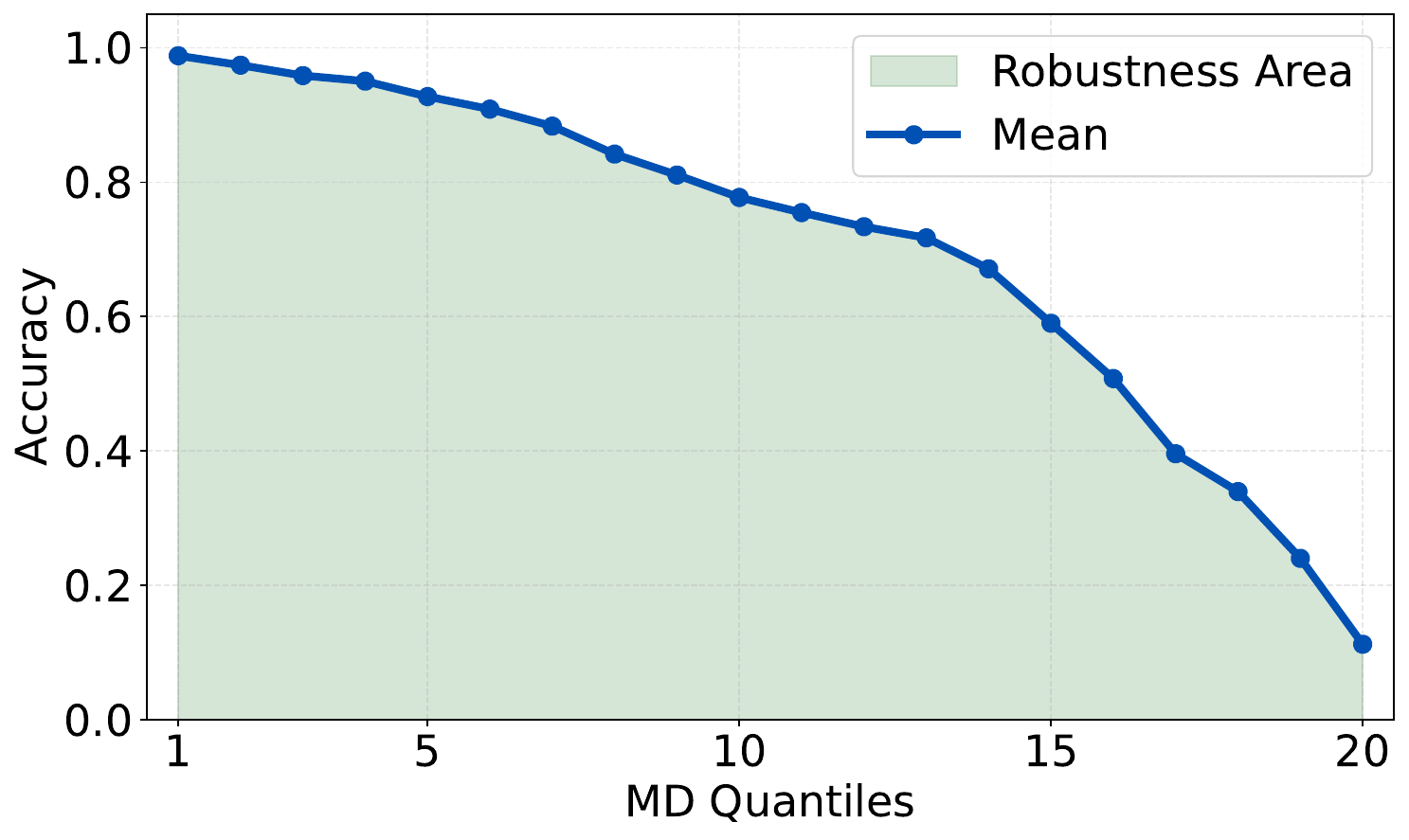}
    \caption{Llama-3.1-8B-it}
    \label{fig:reliability_area_gsms_llama_8b}
  \end{subfigure}
  \hspace{0.01\textwidth}
  \begin{subfigure}[t]{0.31\textwidth}
    \centering
    \includegraphics[width=\linewidth]{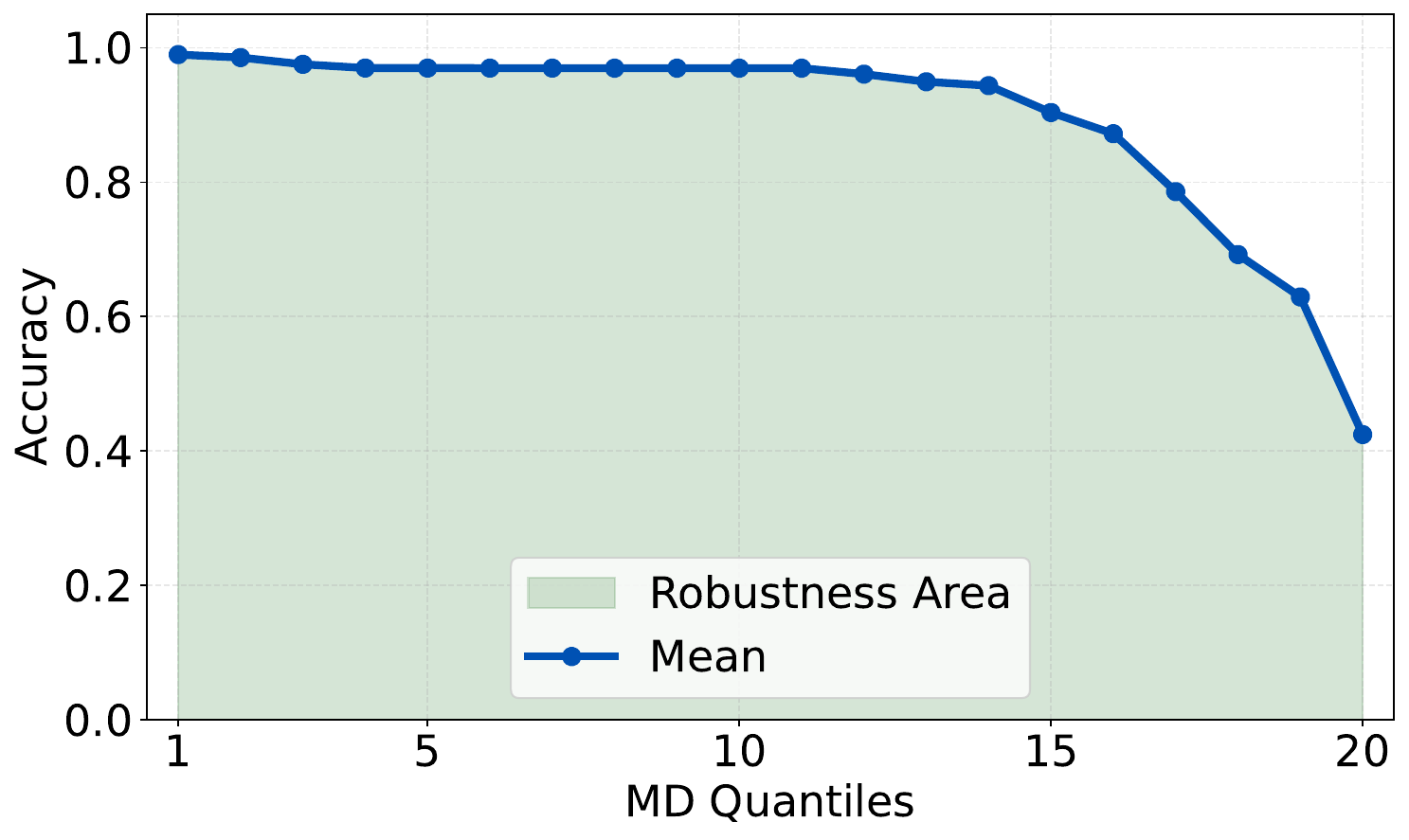}
    \caption{Llama-3.1-70B-it}
    \label{fig:reliability_area_gsms_llama_70b}
  \end{subfigure}
  \caption{\textbf{Robustness area.} Results from different models for samples from \emph{GSM-Symbolic} identified via beam search.}
  \label{fig:reliability_area_gsms}
\end{figure*}

To better quantify robustness to variation difficulty, we compute the normalized area under the accuracy curve (see Figure~\ref{fig:reliability_area_gsms} for a conceptual overview). This enables a more concise comparison across models: a value of 1 corresponds to perfect accuracy across all quantiles, whereas a value of 0 corresponds to total failure across all quantiles. Table~\ref{tab:drs_comparison} summarizes results for all models and datasets. For instance, on problem variations from GSM-Symbolic, Qwen-2.5-7B-Instruct achieves a score of 0.82, Llama-3.1-8B-Instruct achieves 0.71, and Llama-3.1-70B-Instruct achieves 0.90. These results underscore that smaller models from the Qwen family are more robust to increases in difficulty than similarly sized models from the Llama family, and that smaller models are generally less robust than larger models.

\begin{table*}[bp]
    \centering
    \scriptsize
    \begin{tabular}{l|c|ccc|c}
        \toprule
        \textbf{Train set} & \textbf{GSM-S} & \textbf{GSM-S$_{\mathrm{\textbf{low}}}$} & \textbf{GSM-S$_{\mathrm{\textbf{mid}}}$} & \textbf{GSM-S$_{\mathrm{\textbf{high}}}$} & \textbf{GSM8K (100)} \\ 
        \midrule
        -           & 69.29 & 0.00 & 0.00 & 0.00 & 86.00 \\
        \midrule
        $\mathrm{Q_{\mathrm{low}}}$ (100\%)  & 72.93 & 78.20 & 60.15 & 36.84 & 79.00 \\
        $\mathrm{Q_{\mathrm{mid}}}$ (100\%)  & 81.11 & 75.94 & 85.71 & 53.38 & 88.00 \\
        $\mathrm{Q_{\mathrm{high}}}$ (100\%)  & 83.03 & 73.68 & 85.71 & \textbf{85.71} & 86.00 \\
        \midrule
        $\mathrm{Q_{\mathrm{low}}}$ (30\%) - $\mathrm{Q_{\mathrm{mid}}}$ (70\%)  & 82.53 & 83.46 & 84.21 & 56.39 & 84.00 \\
        $\mathrm{Q_{\mathrm{low}}}$ (70\%) - $\mathrm{Q_{\mathrm{mid}}}$ (30\%)  & 84.24 & 85.71 & 82.71 & 56.39 & 89.00 \\
        $\mathrm{Q_{\mathrm{low}}}$ (30\%) - $\mathrm{Q_{\mathrm{high}}}$ (70\%)  & 86.36 & \textbf{86.47} & 87.22 & 81.95 & 82.00 \\
        $\mathrm{Q_{\mathrm{low}}}$ (70\%) - $\mathrm{Q_{\mathrm{high}}}$ (30\%)  & 82.53 & 84.21 & 78.95 & 69.92 & 85.00 \\
        $\mathrm{Q_{\mathrm{mid}}}$ (30\%) - $\mathrm{Q_{\mathrm{high}}}$ (70\%)  & 84.75 & 74.44 & 86.47 & 83.46 & 88.00 \\
        $\mathrm{Q_{\mathrm{mid}}}$ (70\%) - $\mathrm{Q_{\mathrm{high}}}$ (30\%)  & 82.83 & 75.94 & 87.97 & 75.19 & 87.00 \\
        $\mathrm{Q_{\mathrm{low}}}$ (50\%) - $\mathrm{Q_{\mathrm{mid}}}$ (30\%) - $\mathrm{Q_{\mathrm{high}}}$ (20\%)  & 84.04 & 84.21 & 80.45 & 68.42 & \textbf{92.00} \\
        $\mathrm{Q_{\mathrm{low}}}$ (20\%) - $\mathrm{Q_{\mathrm{mid}}}$ (30\%) - $\mathrm{Q_{\mathrm{high}}}$ (50\%) & \textbf{88.18} & \textbf{86.47} & \textbf{91.73} & 83.46 & 91.00 \\
        $\mathrm{Q_{\mathrm{low}}}$ (20\%) - $\mathrm{Q_{\mathrm{mid}}}$ (20\%) - $\mathrm{Q_{\mathrm{high}}}$ (60\%) & 87.27 & 82.71 & 88.72 & 83.46 &  91.00 \\
        $\mathrm{Q_{\mathrm{low}}}$ (10\%) - $\mathrm{Q_{\mathrm{mid}}}$ (10\%) - $\mathrm{Q_{\mathrm{high}}}$ (80\%) & 86.87 & 80.45 & 88.72 & \textbf{85.71} &  88.00 \\
        Random sampling  & 86.16 & 80.45 & 87.97 & 72.93 & 88.00 \\
        \bottomrule
    \end{tabular}
    \caption{\textbf{Accuracy after fine-tuning on GSM-Symbolic subsets of increasing difficulty.} Accuracy (\%) of Llama-3.1-8B-Instruct on beam search samples from GSM-Symbolic and corresponding base problems from GSM8K (100) before and after SFT. Highest-performing accuracy values are shown in bold.}
    \label{tab:full_sft_gsms_llama_8b}
\end{table*}

\subsection{Fine-tuning on difficult problem variations}\label{app:improving_robustness}
As described in Section~\ref{sec:improving_robustness} and further detailed in Appendix~\ref{app:training_details}, we fine-tune Llama-3.1-8B-Instruct on problem variations identified via beam search to improve the model's robustness. Specifically, we evaluate whether training on more challenging problems improves robustness to logic-preserving variations more than training on easier ones. In this section, we present additional results for Llama-3.1-8B-Instruct fine-tuned on mixtures of $\mathrm{Q}_{\mathrm{low}}$--$\mathrm{Q}_{\mathrm{high}}$ data, as well as for Llama-3.1-8B-Instruct trained on variations from FinChain.

\paragraph{\textbf{Fine-tuning on mixtures of difficulty subsets.}} In Table~\ref{tab:full_sft_gsms_llama_8b}, we analyze performance after fine-tuning on mixtures of $\mathrm{Q}_{\mathrm{low}}$--$\mathrm{Q}_{\mathrm{high}}$ data. Overall, performance on GSM-S improves more when the training mix contains a substantial amount of $\mathrm{Q}_{\mathrm{high}}$ data. For instance, training on a mix of 70\% $\mathrm{Q}_{\mathrm{low}}$ and 30\% $\mathrm{Q}_{\mathrm{high}}$ data yields an accuracy of 82.53\% on GSM-S, whereas training on 30\% $\mathrm{Q}_{\mathrm{low}}$ and 70\% $\mathrm{Q}_{\mathrm{high}}$ data achieves 86.36\%. Similarly, training on a mix of 30\% $\mathrm{Q}_{\mathrm{mid}}$ and 70\% $\mathrm{Q}_{\mathrm{high}}$ data yields higher accuracy (84.75\%) than training on 70\% $\mathrm{Q}_{\mathrm{mid}}$ and 30\% $\mathrm{Q}_{\mathrm{high}}$ data (82.83\%). This pattern is consistent with the results discussed in Section~\ref{sec:improving_robustness}. In addition, we find that training on a mix that includes data from all subsets ($\mathrm{Q}_{\mathrm{low}}$--$\mathrm{Q}_{\mathrm{high}}$) improves performance more than training on a single subset or on a combination of two subsets. The best performance is obtained by fine-tuning Llama-3.1-8B-Instruct on a training mixture comprising 20\% from $\mathrm{Q}_{\mathrm{low}}$, 30\% from $\mathrm{Q}_{\mathrm{mid}}$, and 50\% from $\mathrm{Q}_{\mathrm{high}}$. Overall, we find that training on an informed mixture of samples from different difficulty levels improves performance on variations from GSM-Symbolic more than fine-tuning on a random data mix, particularly for variations associated with high difficulty scores (GSM-S$_{\mathrm{high}}$).

\paragraph{\textbf{Fine-tuning on variations from FinChain.}} Analogous to the experiments on GSM-Symbolic (Section~\ref{sec:improving_robustness}), we partition the training data from FinChain into three equally sized subsets of increasing difficulty ($\mathrm{Q}_{\mathrm{low}}$, $\mathrm{Q}_{\mathrm{mid}}$, and $\mathrm{Q}_{\mathrm{high}}$). Table~\ref{tab:full_sft_per_template_val_filter_finchain} summarizes the model's performance after fine-tuning on each subset, as well as after fine-tuning on mixtures of $\mathrm{Q}_{\mathrm{low}}$--$\mathrm{Q}_{\mathrm{high}}$. As described in Section~\ref{sec:improving_robustness}, in addition to accuracy on the full test set (FinChain), we also report performance on three equally sized test subsets with increasing difficulty (FinChain$_{\mathrm{low}}$, FinChain$_{\mathrm{mid}}$, and FinChain$_{\mathrm{high}}$). Overall, we observe trends similar to those for GSM-Symbolic (Table~\ref{tab:full_sft_gsms_llama_8b}). On the full test set, fine-tuning on any training subset ($\mathrm{Q}_{\mathrm{low}}$--$\mathrm{Q}_{\mathrm{high}}$) improves performance relative to the untrained model. Moreover, fine-tuning on more difficult subsets ($\mathrm{Q}_{\mathrm{mid}}$ and $\mathrm{Q}_{\mathrm{high}}$) yields slightly larger gains than fine-tuning on the easiest subset ($\mathrm{Q}_{\mathrm{low}}$) (64.44\% and 64.56\% vs.\ 60.11\%). Examining the difficulty-specific test sets, training on $\mathrm{Q}_{\mathrm{low}}$ produces the largest gains on the lowest-difficulty subset FinChain$_{\mathrm{low}}$ (68.52\%), but performance is substantially lower on FinChain$_{\mathrm{mid}}$ (30.86\%) and FinChain$_{\mathrm{high}}$ (25.93\%). In contrast, training on $\mathrm{Q}_{\mathrm{mid}}$ improves performance on both FinChain$_{\mathrm{low}}$ (66.05\%) and FinChain$_{\mathrm{mid}}$ (57.41\%), with moderate improvements on FinChain$_{\mathrm{high}}$ (41.36\%). Training on $\mathrm{Q}_{\mathrm{high}}$ yields the most consistent performance across difficulty levels, with accuracies of 46.91\% on FinChain$_{\mathrm{low}}$, 55.56\% on FinChain$_{\mathrm{mid}}$, and 63.58\% on FinChain$_{\mathrm{high}}$. These results reinforce the findings in Section~\ref{sec:improving_robustness}, suggesting that training on more difficult samples improves robustness to logic-preserving variations more consistently across difficulty levels than training on easier variations. Table~\ref{tab:full_sft_per_template_val_filter_finchain} further reports performance after fine-tuning on mixtures of $\mathrm{Q}_{\mathrm{low}}$--$\mathrm{Q}_{\mathrm{high}}$. Similar to the results on GSM-Symbolic, we find that training on mixtures that include data from all three subsets improves performance more than training on a single subset or on mixtures of only two subsets. Consistent with the GSM-Symbolic experiments, the best performance is obtained by fine-tuning the model on a mixture comprising 20\% from $\mathrm{Q}_{\mathrm{low}}$, 30\% from $\mathrm{Q}_{\mathrm{mid}}$, and 50\% from $\mathrm{Q}_{\mathrm{high}}$.

\begin{table*}[tbp]
    \centering
    \scriptsize
    \begin{tabular}{l|cccc}
        \toprule
        \textbf{Train set} & \textbf{FinChain} & \textbf{FinChain$_{\mathrm{\textbf{low}}}$} & \textbf{FinChain$_{\mathrm{\textbf{mid}}}$} & \textbf{FinChain$_{\mathrm{\textbf{high}}}$} \\ 
        \midrule
        -           & 54.44 & 0.00 & 0.00 & 0.00 \\
        \midrule
        $\mathrm{Q_{\mathrm{low}}}$ (100\%)  & 60.11 & 68.52 & 30.86 & 25.93 \\
        $\mathrm{Q_{\mathrm{mid}}}$ (100\%)  & 64.44 & 66.05 & 57.41 & 41.36 \\
        $\mathrm{Q_{\mathrm{high}}}$ (100\%)  & 64.56 & 46.91 & 55.56 & 63.58 \\
        \midrule
        $\mathrm{Q_{\mathrm{low}}}$ (30\%) - $\mathrm{Q_{\mathrm{mid}}}$ (70\%) & 66.33 & \textbf{70.37} & 51.23 & 37.04 \\
        $\mathrm{Q_{\mathrm{low}}}$ (70\%) - $\mathrm{Q_{\mathrm{mid}}}$ (30\%)  & 65.89 & 64.81 & 48.15 & 40.74 \\
        $\mathrm{Q_{\mathrm{low}}}$ (30\%) - $\mathrm{Q_{\mathrm{high}}}$ (70\%)  & 69.89 & 60.49 & 50.00 & 56.17 \\
        $\mathrm{Q_{\mathrm{low}}}$ (70\%) - $\mathrm{Q_{\mathrm{high}}}$ (30\%)  & 69.67 & 65.43 & 50.62 & 57.41 \\
        $\mathrm{Q_{\mathrm{mid}}}$ (30\%) - $\mathrm{Q_{\mathrm{high}}}$ (70\%)  & 69.67 & 64.81 & 60.49 & \textbf{64.81} \\
        $\mathrm{Q_{\mathrm{mid}}}$ (70\%) - $\mathrm{Q_{\mathrm{high}}}$ (30\%)  & 63.00 & 58.02 & 50.00 & 48.77 \\
        $\mathrm{Q_{\mathrm{low}}}$ (50\%) - $\mathrm{Q_{\mathrm{mid}}}$ (30\%) - $\mathrm{Q_{\mathrm{high}}}$ (20\%)  & 71.89 & 68.52 & \textbf{61.11} & 57.41 \\
        $\mathrm{Q_{\mathrm{low}}}$ (20\%) - $\mathrm{Q_{\mathrm{mid}}}$ (30\%) - $\mathrm{Q_{\mathrm{high}}}$ (50\%) & \textbf{72.11} & 69.14 & 52.47 & 60.49 \\
        $\mathrm{Q_{\mathrm{low}}}$ (20\%) - $\mathrm{Q_{\mathrm{mid}}}$ (20\%) - $\mathrm{Q_{\mathrm{high}}}$ (60\%) & 71.00 & 68.52 & 57.41 & 61.11 \\
        $\mathrm{Q_{\mathrm{low}}}$ (10\%) - $\mathrm{Q_{\mathrm{mid}}}$ (10\%) - $\mathrm{Q_{\mathrm{high}}}$ (80\%) & 69.56 & 68.52 & 53.09 & 56.79 \\
        Random sampling  & 69.89 & 68.52 & 51.85 & 54.94 \\
        \bottomrule
    \end{tabular}
    \caption{\textbf{Accuracy after fine-tuning on FinChain subsets of increasing difficulty.} Accuracy (\%) of Llama-3.1-8B-Instruct on beam search samples from FinChain before and after SFT. Highest-performing accuracy values are shown in bold.}
    \label{tab:full_sft_per_template_val_filter_finchain}
\end{table*}

% Llama family overview
\begin{table}[bp]
\centering
\scriptsize
\begin{adjustbox}{width=\linewidth,center}
\begin{tabular}{
  lcl
  | S[table-format=1.3]
    S[table-format=-1.3]
    >{\hspace{-1.0em}}l
  | S[table-format=1.3]
    S[table-format=-1.3]
    >{\hspace{-1.0em}}l
  | S[table-format=1.3]
    S[table-format=-1.3]
    >{\hspace{-1.0em}}l
}
\toprule
& \multicolumn{1}{c}{\multirow{2}{*}{\textbf{Span}}}
& \multicolumn{1}{c}{\multirow{2}{*}{\textbf{Metric}}}
& \multicolumn{3}{c}{\textbf{GSM-Symbolic}}
& \multicolumn{3}{c}{\textbf{FinChain}}
& \multicolumn{3}{c}{\textbf{EngTrace}} \\
\cmidrule(lr){4-6}\cmidrule(lr){7-9}\cmidrule(lr){10-12}
& &
& \multicolumn{1}{c}{\textbf{AUC}}
& \multicolumn{2}{c}{\textbf{Odds Ratio}}
& \multicolumn{1}{c}{\textbf{AUC}}
& \multicolumn{2}{c}{\textbf{Odds Ratio}}
& \multicolumn{1}{c}{\textbf{AUC}}
& \multicolumn{2}{c}{\textbf{Odds Ratio}} \\
\toprule

\multirow{14}{*}{\rotatebox{90}{Llama-3.2-3B-Instruct}}
& \multirow{5}{*}{Input}
& Perplexity           & 0.530 & 0.809 & (0.79, 0.83) & 0.516 & 0.886 & (0.87, 0.90) & 0.524 & 0.342 & (0.33, 0.35) \\
& & Entropy            & 0.517 & 0.842 & (0.82, 0.86) & 0.516 & 0.763 & (0.74, 0.78) & 0.511 & 0.746 & (0.72, 0.77) \\
& & Self-Certainty         & 0.518 & 1.205 & (1.18, 1.23) & 0.514 & 1.285 & (1.25, 1.32) & 0.489 & 0.992 & (0.96, 1.03) \\
& & $\text{MD}_{\mathcal{E}}$      & 0.539 & 0.918 & (0.89, 0.94) & 0.558 & 0.809 & (0.79, 0.83) & 0.582 & 0.192 & (0.18, 0.20) \\
& & $\text{KNN}_\mathcal{E}$     & 0.533 & 0.795 & (0.78, 0.81) & 0.541 & 0.669 & (0.65, 0.68) & 0.564 & 0.351 & (0.34, 0.37) \\
\cmidrule{2-12}

& \multirow{9}{*}{Output}
& Perplexity           & 0.493 & 0.933 & (0.91, 0.96) & 0.503 & 1.015 & (0.99, 1.04) & 0.478 & 1.053 & (1.02, 1.08) \\
& & Entropy            & 0.533 & 0.773 & (0.76, 0.79) & 0.536 & 0.860 & (0.84, 0.88) & 0.463 & 1.303 & (1.26, 1.35) \\
& & Self-Certainty         & 0.539 & 1.270 & (1.24, 1.30) & 0.529 & 1.131 & (1.10, 1.16) & 0.479 & 0.780 & (0.75, 0.81) \\
& & $\text{LD}_{\text{min}}$    & 0.760 & 0.214 & (0.21, 0.22) & 0.698 & \textbf{0.123} & (0.12, 0.13) & 0.799 & 0.081 & (0.08, 0.08) \\
& & $\text{LD}_{\text{max}}$    & 0.611 & 0.254 & (0.25, 0.26) & 0.580 & 0.127 & (0.12, 0.13) & 0.598 & 0.106 & (0.10, 0.11) \\
& & $\text{LD}_{\text{mean}}$   & 0.708 & 0.186 & (0.18, 0.19) & 0.685 & 0.226 & (0.22, 0.23) & 0.751 & 0.141 & (0.14, 0.15) \\
& & $\text{LD}_{\text{median}}$ & 0.687 & 0.228 & (0.22, 0.23) & 0.662 & 0.350 & (0.34, 0.36) & 0.720 & 0.222 & (0.21, 0.23) \\
& & $\text{MD}_{\mathcal{H}}$    & \textbf{0.799} & \textbf{0.134} & (0.13, 0.14) & \textbf{0.707} & 0.151 & (0.14, 0.16) & \textbf{0.820} & \textbf{0.004} & (0.00, 0.00) \\
& & $\text{KNN}_{\mathcal{H}}$   & 0.673 & 0.442 & (0.43, 0.45) & 0.614 & 0.425 & (0.41, 0.44) & 0.709 & 0.128 & (0.12, 0.14) \\
\cmidrule{1-12}

\multirow{14}{*}{\rotatebox{90}{Llama-3.1-8B-Instruct}}
& \multirow{5}{*}{Input}
& Perplexity           & 0.53 & 0.841 & (0.82, 0.86) & 0.494 & 1.035 & (0.66, 1.61) & 0.536 & 0.310 & (0.30, 0.32) \\
& & Entropy            & 0.51 & 0.861 & (0.84, 0.88) & 0.501 & 0.990 & (0.96, 1.02) & 0.527 & 0.429 & (0.42, 0.44) \\
& & Self-Certainty         & 0.51 & 1.105 & (1.08, 1.13) & 0.498 & 0.926 & (0.90, 0.95) & 0.512 & 1.392 & (1.36, 1.43) \\
& & $\text{MD}_{\mathcal{E}}$      & 0.54 & 0.915 & (0.89, 0.94) & 0.539 & 0.825 & (0.80, 0.85) & 0.574 & 0.248 & (0.24, 0.26) \\
& & $\text{KNN}_{\mathcal{E}}$     & 0.53 & 0.718 & (0.70, 0.73) & 0.535 & 0.562 & (0.55, 0.58) & 0.574 & 0.441 & (0.43, 0.45) \\
\cmidrule{2-12}

& \multirow{9}{*}{Output}
& Perplexity           & 0.436 & 0.959 & (0.94, 0.98) & 0.461 & 1.026 & (1.00, 1.05) & 0.476 & 1.077 & (1.05, 1.10) \\
& & Entropy            & 0.507 & 0.964 & (0.94, 0.98) & 0.534 & 0.842 & (0.82, 0.86) & 0.522 & 0.869 & (0.85, 0.89) \\
& & Self-Certainty         & 0.561 & 1.359 & (1.33, 1.39) & 0.536 & 1.139 & (1.11, 1.16) & 0.548 & 1.195 & (1.16, 1.23) \\
& & $\text{LD}_{\text{min}}$    & 0.783 & 0.183 & (0.18, 0.19) & 0.702 & \textbf{0.157} & (0.15, 0.17) & 0.761 & 0.198 & (0.19, 0.21) \\
& & $\text{LD}_{\text{max}}$    & 0.610 & 0.277 & (0.27, 0.28) & 0.556 & 0.158 & (0.15, 0.16) & 0.500 & 0.726 & (0.71, 0.75) \\
& & $\text{LD}_{\text{mean}}$   & 0.745 & 0.146 & (0.14, 0.15) & 0.694 & 0.271 & (0.26, 0.28) & 0.735 & 0.133 & (0.13, 0.14) \\
& & $\text{LD}_{\text{median}}$ & 0.725 & 0.177 & (0.17, 0.18) & 0.654 & 0.419 & (0.41, 0.43) & 0.707 & 0.189 & (0.18, 0.19) \\
& & $\text{MD}_{\mathcal{H}}$    & \textbf{0.804} & \textbf{0.138} & (0.13, 0.14) & \textbf{0.737} & 0.487 & (0.47, 0.51) & \textbf{0.777} & \textbf{0.021} & (0.02, 0.02) \\
& & $\text{KNN}_{\mathcal{H}}$   & 0.709 & 0.395 & (0.38, 0.41) & 0.645 & 0.586 & (0.57, 0.61) & 0.674 & 0.227 & (0.22, 0.24) \\
\cmidrule{1-12}

\multirow{14}{*}{\rotatebox{90}{Llama-3.1-70B-Instruct}}
& \multirow{5}{*}{Input}
& Perplexity           & 0.518 & 0.920 & (0.89, 0.96) & 0.500 & 0.981 & (0.95, 1.01) & 0.544 & 0.254 & (0.25, 0.26) \\
& & Entropy            & 0.530 & 0.763 & (0.73, 0.80) & 0.522 & 0.692 & (0.67, 0.72) & 0.537 & 0.373 & (0.36, 0.38) \\
& & Self-Certainty         & 0.521 & 1.198 & (1.15, 1.25) & 0.521 & 2.399 & (2.32, 2.49) & 0.530 & 2.323 & (2.26, 2.38) \\
& & $\text{MD}_{\mathcal{E}}$      & 0.558 & 0.862 & (0.81, 0.91) & 0.533 & 0.882 & (0.85, 0.92) & 0.576 & 0.235 & (0.23, 0.24) \\
& & $\text{KNN}_{\mathcal{E}}$     & 0.550 & 0.470 & (0.45, 0.49) & 0.516 & 0.619 & (0.59, 0.64) & 0.565 & 0.464 & (0.45, 0.48) \\
\cmidrule{2-12}

& \multirow{9}{*}{Output}
& Perplexity           & 0.319 & 2.413 & (2.25, 2.58) & 0.480 & 1.057 & (1.03, 1.09) & 0.455 & 1.811 & (1.71, 1.91) \\
& & Entropy            & 0.682 & 0.315 & (0.30, 0.33) & 0.596 & 0.711 & (0.69, 0.73) & 0.610 & 0.584 & (0.57, 0.60) \\
& & Self-Certainty         & 0.740 & 5.312 & (5.13, 5.50) & 0.621 & 2.643 & (2.56, 2.73) & 0.622 & 2.537 & (2.47, 2.60) \\
& & $\text{LD}_{\text{min}}$    & 0.835 & 0.165 & (0.16, 0.17) & 0.657 & 0.421 & (0.41, 0.44) & 0.694 & 0.226 & (0.22, 0.23) \\
& & $\text{LD}_{\text{max}}$    & 0.698 & \textbf{0.070} & (0.07, 0.07) & 0.569 & \textbf{0.112} & (0.11, 0.12) & 0.534 & 0.364 & (0.35, 0.37) \\
& & $\text{LD}_{\text{mean}}$   & 0.818 & 0.077 & (0.07, 0.08) & 0.613 & 0.330 & (0.32, 0.34) & 0.668 & 0.212 & (0.21, 0.22) \\
& & $\text{LD}_{\text{median}}$ & 0.795 & 0.091 & (0.09, 0.10) & 0.598 & 0.408 & (0.40, 0.42) & 0.656 & 0.273 & (0.27, 0.28) \\
& & $\text{MD}_{\mathcal{H}}$    & \textbf{0.867} & 0.159 & (0.15, 0.17) & \textbf{0.671} & 0.384 & (0.37, 0.40) & \textbf{0.710} & \textbf{0.123} & (0.12, 0.13) \\
& & $\text{KNN}_{\mathcal{H}}$   & 0.802 & 0.215 & (0.21, 0.22) & 0.604 & 0.597 & (0.58, 0.61) & 0.641 & 0.336 & (0.32, 0.35) \\
\bottomrule
\end{tabular}
\end{adjustbox}
\caption{\textbf{Predictive power of metrics for answer correctness.} AUC scores and odds ratios for models from the \emph{Llama} family across problem variations of \emph{GSM-Symbolic}, \emph{FinChain}, and \emph{EngTrace}. Most significant scores for each model are shown in bold.}
\label{tab:correlation_llama_family}
\end{table}

% Qwen family overview
\begin{table}[tbp]
\centering
\scriptsize
\begin{adjustbox}{width=\linewidth,center}
\begin{tabular}{
  lcl
  | S[table-format=1.3]
    S[table-format=-1.3]
    >{\hspace{-1.0em}}l
  | S[table-format=1.3]
    S[table-format=-1.3]
    >{\hspace{-1.0em}}l
  | S[table-format=1.3]
    S[table-format=-1.3]
    >{\hspace{-1.0em}}l
}
\toprule
& \multicolumn{1}{c}{\multirow{2}{*}{\textbf{Span}}}
& \multicolumn{1}{c}{\multirow{2}{*}{\textbf{Metric}}}
& \multicolumn{3}{c}{\textbf{GSM-Symbolic}}
& \multicolumn{3}{c}{\textbf{FinChain}}
& \multicolumn{3}{c}{\textbf{EngTrace}} \\
\cmidrule(lr){4-6}\cmidrule(lr){7-9}\cmidrule(lr){10-12}
& &
& \multicolumn{1}{c}{\textbf{AUC}}
& \multicolumn{2}{c}{\textbf{Odds Ratio}}
& \multicolumn{1}{c}{\textbf{AUC}}
& \multicolumn{2}{c}{\textbf{Odds Ratio}}
& \multicolumn{1}{c}{\textbf{AUC}}
& \multicolumn{2}{c}{\textbf{Odds Ratio}} \\
\toprule

\multirow{14}{*}{\rotatebox{90}{Qwen-2.5-7B-Instruct}}
& \multirow{5}{*}{Input}
& Perplexity           & 0.508 & 0.938 & (0.91, 0.97) & 0.509 & 0.759 & (0.73, 0.79) & 0.542 & 0.290 & (0.28, 0.30) \\
& & Entropy            & 0.500 & 1.002 & (0.97, 1.03) & 0.508 & 0.863 & (0.84, 0.89) & 0.501 & 1.088 & (1.06, 1.11) \\
& & Self-Certainty         & 0.499 & 1.018 & (0.99, 1.05) & 0.509 & 1.195 & (1.16, 1.24) & 0.502 & 1.177 & (1.15, 1.21) \\
& & $\text{MD}_{\mathcal{E}}$      & 0.518 & 0.969 & (0.94, 1.00) & 0.530 & 0.901 & (0.87, 0.93) & 0.547 & 0.321 & (0.31, 0.33) \\
& & $\text{KNN}_{\mathcal{E}}$     & 0.519 & 0.783 & (0.76, 0.81) & 0.508 & 0.893 & (0.86, 0.92) & 0.546 & 0.495 & (0.48, 0.51) \\
\cmidrule{2-12}
& \multirow{9}{*}{Output}
& Perplexity           & 0.451 & 1.035 & (0.99, 1.08) & 0.479 & 1.287 & (1.19, 1.39) & 0.450 & 1.147 & (1.11, 1.19) \\
& & Entropy            & 0.464 & 1.256 & (1.22, 1.29) & 0.478 & 1.127 & (1.10, 1.16) & 0.457 & 1.198 & (1.17, 1.23) \\
& & Self-Certainty         & 0.495 & 0.952 & (0.92, 0.98) & 0.489 & 0.943 & (0.92, 0.97) & 0.467 & 0.854 & (0.83, 0.88) \\
& & $\text{LD}_{\text{min}}$    & 0.726 & 0.276 & (0.27, 0.28) & 0.660 & \textbf{0.338} & (0.33, 0.35) & 0.732 & 0.252 & (0.24, 0.26) \\
& & $\text{LD}_{\text{max}}$    & 0.569 & 0.499 & (0.48, 0.51) & 0.498 & 1.034 & (0.62, 1.73) & 0.588 & 0.146 & (0.14, 0.15) \\
& & $\text{LD}_{\text{mean}}$   & 0.667 & 0.327 & (0.32, 0.34) & 0.610 & 0.358 & (0.35, 0.37) & 0.696 & 0.218 & (0.21, 0.23) \\
& & $\text{LD}_{\text{median}}$ & 0.640 & 0.431 & (0.42, 0.44) & 0.586 & 0.450 & (0.44, 0.46) & 0.670 & 0.342 & (0.33, 0.35) \\
& & $\text{MD}_{\mathcal{H}}$   & \textbf{0.760} & \textbf{0.157} & (0.15, 0.16) & \textbf{0.666} & 0.570 & (0.55, 0.59) & \textbf{0.750} & \textbf{0.024} & (0.02, 0.03) \\
& & $\text{KNN}_{\mathcal{H}}$  & 0.635 & 0.491 & (0.47, 0.51) & 0.575 & 0.767 & (0.74, 0.79) & 0.662 & 0.256 & (0.24, 0.27) \\
\cmidrule{1-12}

\multirow{14}{*}{\rotatebox{90}{Qwen-2.5-32B-Instruct}}
& \multirow{5}{*}{Input}
& Perplexity           & 0.514 & 0.896 & (0.86, 0.94) & 0.503 & 0.975 & (0.93, 1.03) & 0.525 & 0.432 & (0.42, 0.44) \\
& & Entropy            & 0.510 & 0.933 & (0.90, 0.97) & 0.522 & 0.750 & (0.72, 0.78) & 0.533 & 0.518 & (0.51, 0.53) \\
& & Self-Certainty         & 0.512 & 1.090 & (1.05, 1.13) & 0.527 & 1.361 & (1.31, 1.41) & 0.516 & 1.600 & (1.56, 1.64) \\
& & $\text{MD}_{\mathcal{E}}$      & 0.528 & 0.969 & (0.91, 1.03) & 0.537 & 0.962 & (0.92, 1.01) & 0.548 & 0.311 & (0.30, 0.33) \\
& & $\text{KNN}_{\mathcal{E}}$     & 0.531 & 0.667 & (0.64, 0.70) & 0.540 & 0.630 & (0.60, 0.66) & 0.541 & 0.477 & (0.46, 0.49) \\
\cmidrule{2-12}
& \multirow{9}{*}{Output}
& Perplexity           & 0.423 & 1.972 & (1.84, 2.11) & 0.483 & 1.791 & (1.66, 1.94) & 0.488 & 1.087 & (1.06, 1.12) \\
& & Entropy            & 0.435 & 1.542 & (1.48, 1.60) & 0.512 & 1.107 & (1.06, 1.16) & 0.506 & 1.008 & (0.98, 1.03) \\
& & Self-Certainty         & 0.458 & 0.732 & (0.70, 0.76) & 0.510 & 1.008 & (0.97, 1.05) & 0.496 & 0.949 & (0.93, 0.97) \\
& & $\text{LD}_{\text{min}}$    & 0.694 & 0.374 & (0.36, 0.39) & 0.656 & 0.398 & (0.38, 0.41) & 0.698 & 0.285 & (0.28, 0.29) \\
& & $\text{LD}_{\text{max}}$    & 0.587 & 0.345 & (0.33, 0.36) & 0.561 & \textbf{0.242} & (0.24, 0.25) & 0.559 & 0.342 & (0.33, 0.35) \\
& & $\text{LD}_{\text{mean}}$   & 0.644 & 0.365 & (0.35, 0.38) & 0.671 & 0.281 & (0.27, 0.29) & 0.644 & 0.355 & (0.35, 0.36) \\
& & $\text{LD}_{\text{median}}$ & 0.630 & 0.423 & (0.41, 0.44) & 0.651 & 0.345 & (0.34, 0.35) & 0.606 & 0.481 & (0.47, 0.49) \\
& & $\text{MD}_{\mathcal{H}}$   & \textbf{0.770} & \textbf{0.246} & (0.24, 0.26) & \textbf{0.704} & 0.316 & (0.30, 0.33) & \textbf{0.732} & \textbf{0.104} & (0.10, 0.11) \\
& & $\text{KNN}_{\mathcal{H}}$  & 0.652 & 0.563 & (0.55, 0.58) & 0.643 & 0.715 & (0.69, 0.74) & 0.648 & 0.348 & (0.33, 0.36) \\
\bottomrule
\end{tabular}
\end{adjustbox}
\caption{\textbf{Predictive power of metrics for answer correctness.} AUC scores and odds ratios for models from the \emph{Qwen} family across problem variations of \emph{GSM-Symbolic}, \emph{FinChain}, and \emph{EngTrace}. Most significant scores for each model are shown in bold.}
\label{tab:correlation_qwen_family}
\end{table}

\begin{table}[tbp]
\centering
\small
\begin{tabular}{
  lcl
  | S[table-format=1.3]
    S[table-format=-1.3]
    >{\hspace{-0.8em}}l
  | S[table-format=1.3]
    S[table-format=-1.3]
    >{\hspace{-0.8em}}l
}
\toprule
& \multicolumn{1}{c}{\multirow{2}{*}{\textbf{Span}}}
& \multicolumn{1}{c}{\multirow{2}{*}{\textbf{Metric}}}
& \multicolumn{3}{c}{\textbf{GSM-Symbolic}}
& \multicolumn{3}{c}{\textbf{FinChain}} \\
\cmidrule(lr){4-6}\cmidrule(lr){7-9}
& &
& \multicolumn{1}{c}{\textbf{AUC}}
& \multicolumn{2}{c}{\textbf{Odds Ratio}}
& \multicolumn{1}{c}{\textbf{AUC}}
& \multicolumn{2}{c}{\textbf{Odds Ratio}} \\
\toprule

\multirow{14}{*}{\rotatebox{90}{Qwen-2.5-72B-Instruct}}
& \multirow{7}{*}{Input}
& Perplexity           & 0.525 & 0.828 & (0.79, 0.87) & 0.486 & 1.496 & (1.41, 1.59) \\
& & Entropy            & 0.489 & 1.064 & (1.02, 1.12) & 0.509 & 0.903 & (0.87, 0.94) \\
& & Self-Certainty         & 0.504 & 1.065 & (1.01, 1.12) & 0.515 & 1.385 & (1.34, 1.43) \\
& & $\text{MD}_{\mathcal{E}}$      & 0.538 & 0.968 & (0.90, 1.04) & 0.531 & 0.991 & (0.95, 1.03) \\
& & $\text{KNN}_{\mathcal{E}}$     & 0.540 & 0.589 & (0.56, 0.62) & 0.533 & 0.482 & (0.46, 0.50) \\

\cmidrule{2-9}

& \multirow{5}{*}{Output}
& Perplexity          & 0.332 & 10.257 & (9.52, 11.05) & 0.520 & 0.912 & (0.85, 0.98) \\
& & Entropy           & 0.376 & 1.587 & (1.52, 1.66) & 0.543 & 0.805 & (0.78, 0.83) \\
& & Self-Certainty         & 0.392 & 0.568 & (0.54, 0.59) & 0.553 & 1.344 & (1.30, 1.39) \\
& & $\text{LD}_{\text{min}}$ & 0.755 & 0.258 & (0.25, 0.27) & 0.694 & 0.311 & (0.30, 0.32) \\
& & $\text{LD}_{\text{max}}$ & 0.584 & 0.351 & (0.33, 0.37) & 0.563 & \textbf{0.066} & (0.06, 0.07) \\
& & $\text{LD}_{\text{mean}}$ & 0.708 & 0.238 & (0.23, 0.25) & 0.662 & 0.266 & (0.26, 0.27) \\
& & $\text{LD}_{\text{median}}$ & 0.690 & 0.305 & (0.29, 0.32) & 0.600 & 0.406 & (0.39, 0.42) \\
& & $\text{MD}_{\mathcal{H}}$   & \textbf{0.835} & \textbf{0.157} & (0.15, 0.17) & \textbf{0.703} & 0.381 & (0.37, 0.39) \\
& & $\text{KNN}_{\mathcal{H}}$  & 0.746 & 0.337 & (0.32, 0.35) & 0.620 & 0.743 & (0.73, 0.76) \\

\bottomrule
\end{tabular}
\caption{\textbf{Predictive power of metrics for answer correctness.} AUC scores and odds ratios for \emph{Qwen-2.5-72B-Instruct} across problem variations of \emph{GSM-Symbolic} and \emph{FinChain}. Most significant scores are shown in bold.}
\label{tab:correlation_qwen_72b}
\end{table}

% Phi overview
\begin{table}[tbp]
\centering
\small

\begin{tabular}{
  lcl
  | S[table-format=1.3]
    S[table-format=-1.3]
    >{\hspace{-0.8em}}l
  | S[table-format=1.3]
    S[table-format=-1.3]
    >{\hspace{-0.8em}}l
}
\toprule
& \multicolumn{1}{c}{\multirow{2}{*}{\textbf{Span}}}
& \multicolumn{1}{c}{\multirow{2}{*}{\textbf{Metric}}}
& \multicolumn{3}{c}{\textbf{GSM-Symbolic}}
& \multicolumn{3}{c}{\textbf{FinChain}} \\
\cmidrule(lr){4-6}\cmidrule(lr){7-9}
& &
& \multicolumn{1}{c}{\textbf{AUC}}
& \multicolumn{2}{c}{\textbf{Odds Ratio}}
& \multicolumn{1}{c}{\textbf{AUC}}
& \multicolumn{2}{c}{\textbf{Odds Ratio}} \\
\toprule

\multirow{14}{*}{\rotatebox{90}{Phi-4-14B}}
& \multirow{7}{*}{Input}
& Perplexity           & 0.524 & 0.846 & (0.82, 0.88) & 0.499 & 0.969 & (0.94, 1.00) \\
& & Entropy            & 0.495 & 0.992 & (0.95, 1.03) & 0.488 & 1.116 & (1.08, 1.15) \\
& & Self-Certainty         & 0.500 & 1.002 & (0.97, 1.04) & 0.504 & 1.413 & (1.37, 1.46) \\
& & $\text{MD}_{\mathcal{E}}$      & 0.547 & 0.890 & (0.86, 0.92) & 0.559 & 0.835 & (0.81, 0.86) \\
& & $\text{KNN}_{\mathcal{E}}$     & 0.559 & 0.403 & (0.39, 0.42) & 0.580 & 0.309 & (0.30, 0.32) \\

\cmidrule{2-9}

& \multirow{5}{*}{Output}
& Perplexity          & 0.283 & 3.080 & (2.69, 3.53) & 0.381 & 1.459 & (1.38, 1.54) \\
& & Entropy           & 0.503 & 0.883 & (0.85, 0.92) & 0.470 & 1.335 & (1.29, 1.39) \\
& & Self-Certainty         & 0.481 & 0.869 & (0.84, 0.90) & 0.459 & 0.749 & (0.72, 0.78) \\
& & $\text{LD}_{\text{min}}$ & 0.842 & 0.148 & (0.14, 0.15) & 0.759 & 0.268 & (0.26, 0.28) \\
& & $\text{LD}_{\text{max}}$ & 0.721 & 0.049 & (0.05, 0.05) & 0.666 & 0.032 & (0.03, 0.03) \\
& & $\text{LD}_{\text{mean}}$ & 0.796 & 0.108 & (0.10, 0.11) & 0.723 & 0.174 & (0.17, 0.18) \\
& & $\text{LD}_{\text{median}}$ & 0.773 & 0.149 & (0.14, 0.15) & 0.685 & 0.288 & (0.28, 0.30) \\
& & $\text{MD}_{\mathcal{H}}$   & \textbf{0.911} & \textbf{0.001} & (0.00, 0.00) & \textbf{0.808} & \textbf{0.027} & (0.02, 0.03) \\
& & $\text{KNN}_{\mathcal{H}}$  & 0.784 & 0.230 & (0.22, 0.24) & 0.734 & 0.380 & (0.37, 0.39) \\

\bottomrule
\end{tabular}
\caption{\textbf{Predictive power of metrics for answer correctness.} AUC scores and odds ratios for \emph{Phi-4} across problem variations of \emph{GSM-Symbolic} and \emph{FinChain}. Most significant scores are shown in bold.}
\label{tab:correlation_phi4_14b}
\end{table}

% Cross-reference overview
\begin{table}[tbp]
\centering
\scriptsize
\begin{subtable}[t]{0.48\textwidth}
\centering
\begin{tabular}{
    cl
    | S[table-format=1.3]
    S[table-format=-1.3]
    >{\hspace{-0.8em}}l
    }
    \toprule
    \multicolumn{1}{c}{\multirow{2}{*}{\textbf{Model}}}
    & \multicolumn{1}{c}{\multirow{2}{*}{\textbf{Ref. Model}}}
    & \multicolumn{3}{c}{\textbf{GSM-Symbolic}} \\
    \cmidrule(lr){3-5}
    \multicolumn{1}{c}{} 
    & \multicolumn{1}{c}{}
    & \textbf{AUC} & \multicolumn{2}{c}{\textbf{Odds Ratio}} \\
    \toprule
    \multirow{8}{*}{\rotatebox{90}{Llama-3.2-3B}}
        & Llama-3.2-3B & 0.759 & 0.214 & (0.21, 0.22) \\
        \cmidrule{2-5}
        & Llama-3.1-8B  & 0.665 & 0.319 & (0.31, 0.33) \\
        & Llama-3.1-70B  & 0.658 & 0.327 & (0.32, 0.33) \\
        & Qwen-2.5-7B  & 0.674 & 0.301 & (0.29, 0.31) \\
        & Qwen-2.5-32B  & 0.624 & 0.393 & (0.38, 0.40) \\
        & Qwen-2.5-72B  & 0.643 & 0.350 & (0.34, 0.36) \\
        \cmidrule{2-5}
        & Ground Truth  & 0.670 & 0.233 & (0.23, 0.24) \\
    \midrule
    \multirow{8}{*}{\rotatebox{90}{Llama-3.1-8B}}
        & Llama-3.1-8B & 0.783 & 0.183 & (0.18, 0.19) \\
        \cmidrule{2-5}
        & Llama-3.2-3B  & 0.675 & 0.291 & (0.28, 0.30) \\
        & Llama-3.1-70B  & 0.679 & 0.293 & (0.29, 0.30) \\
        & Qwen-2.5-7B  & 0.674 & 0.301 & (0.29, 0.31) \\
        & Qwen-2.5-32B  & 0.658 & 0.345 & (0.34, 0.35) \\
        & Qwen-2.5-72B  & 0.673 & 0.310 & (0.30, 0.32) \\
        \cmidrule{2-5}
        & Ground Truth  & 0.700 & 0.192 & (0.19, 0.20) \\
    \midrule
    \multirow{8}{*}{\rotatebox{90}{Llama-3.1-70B}}
        & Llama-3.1-70B & 0.835 & 0.165 & (0.16, 0.17) \\
        \cmidrule{2-5}
        & Llama-3.2-3B  & 0.731 & 0.096 & (0.09, 0.10) \\
        & Llama-3.1-8B  & 0.738 & 0.150 & (0.15, 0.15) \\
        & Qwen-2.5-7B  & 0.746 & 0.122 & (0.12, 0.13) \\
        & Qwen-2.5-32B  & 0.735 & 0.110 & (0.11, 0.11) \\
        & Qwen-2.5-72B  & 0.726 & 0.127 & (0.12, 0.13) \\
        \cmidrule{2-5}
        & Ground Truth  & 0.747 & 0.086 & (0.08, 0.09) \\
    \bottomrule
\end{tabular}
\subcaption{Llama as base models}
\label{tab:cross_model_edit_score_a}
\end{subtable}
\hfill
\begin{subtable}[t]{0.48\textwidth}
\centering
\begin{tabular}{
    cl
    | S[table-format=1.3]
    S[table-format=-1.3]
    >{\hspace{-0.8em}}l
    }
    \toprule
    \multicolumn{1}{c}{\multirow{2}{*}{\textbf{Model}}}
    & \multicolumn{1}{c}{\multirow{2}{*}{\textbf{Ref. Model}}}
    & \multicolumn{3}{c}{\textbf{GSM-Symbolic}} \\
    \cmidrule(lr){3-5}
    \multicolumn{1}{c}{} 
    & \multicolumn{1}{c}{}
    & \textbf{AUC} & \multicolumn{2}{c}{\textbf{Odds Ratio}} \\
    \toprule
    \multirow{8}{*}{\rotatebox{90}{Qwen-2.5-7B}}
    & Qwen-2.5-7B & 0.726 & 0.276 & (0.27, 0.28) \\
    \cmidrule{2-5}
    & Llama-3.2-3B  & 0.638 & 0.380 & (0.37, 0.39) \\
    & Llama-3.1-8B  & 0.629 & 0.395 & (0.38, 0.41) \\
    & Llama-3.1-70B  & 0.642 & 0.343 & (0.33, 0.35) \\
    & Qwen-2.5-32B  & 0.642 & 0.421 & (0.41, 0.43) \\
    & Qwen-2.5-72B  & 0.632 & 0.416 & (0.41, 0.43) \\
    \cmidrule{2-5}
        & Ground Truth  & 0.652 & 0.238 & (0.23, 0.25) \\
    \midrule
    \multirow{8}{*}{\rotatebox{90}{Qwen-2.5-32B}}
    & Qwen-2.5-32B  & 0.694 & 0.374 & (0.36, 0.39) \\
    \cmidrule{2-5}
    & Llama-3.2-3B  & 0.565 & 0.527 & (0.51, 0.55) \\
    & Llama-3.1-8B & 0.563 & 0.572 & (0.55, 0.59) \\
    & Llama-3.1-70B  & 0.583 & 0.518 & (0.50, 0.54) \\
    & Qwen-2.5-7B  & 0.592 & 0.471 & (0.45, 0.49) \\
    & Qwen-2.5-72B  & 0.599 & 0.487 & (0.47, 0.51) \\
    \cmidrule{2-5}
        & Ground Truth  & 0.623 & 0.260 & (0.25, 0.27) \\
    \midrule
    \multirow{8}{*}{\rotatebox{90}{Qwen-2.5-72B}}
    & Qwen-2.5-72B  & 0.755 & 0.258 & (0.25, 0.27) \\
    \cmidrule{2-5}
    & Llama-3.2-3B  & 0.653 & 0.371 & (0.36, 0.38) \\
    & Llama-3.1-8B  & 0.672 & 0.289 & (0.28, 0.30) \\
    & Llama-3.1-70B  & 0.674 & 0.406 & (0.39, 0.42) \\
    & Qwen-2.5-7B  & 0.664 & 0.260 & (0.25, 0.27) \\
    & Qwen-2.5-32B  & 0.651 & 0.366 & (0.35, 0.38) \\
    \cmidrule{2-5}
        & Ground Truth  & 0.689 & 0.118 & (0.11, 0.12)  \\
\bottomrule
\end{tabular}
\subcaption{Qwen as base models}
\label{tab:cross_model_edit_score_b}
\end{subtable}
\caption{\textbf{Predictive power of minimum edit distance for answer correctness.} The set of reference responses $\mathcal{Y}$ is taken from different reference models. Problem variations are taken from \emph{GSM-Symbolic}~\citep{mirzadeh2025gsmsymbolic}.}
\label{tab:cross_model_edit_score}
\end{table}

\newpage

% Metric distributions

% Beam history

% Error rates per template

% Reliability diagrams

% DRS
\begin{table}[bp]
\centering
\small
\begin{subtable}[t]{0.31\textwidth}
\centering
\small
\begin{tabular}{l|c}
\toprule
\textbf{Model} & \textbf{DRS} \\
\toprule
    Llama-3.2-3B-it & 0.54 \\
    Llama-3.1-8B-it  & 0.71 \\
    Llama-3.1-70B-it  &  \textbf{0.90} \\
    Qwen-2.5-7B-it  & 0.82 \\
    Qwen-2.5-32B-it  & 0.86 \\
    Qwen-2.5-72B-it  & 0.86 \\
    Phi-4 (14B) & 0.89 \\
    \bottomrule
\end{tabular}
\subcaption{GSM-Symbolic}
\label{tab:drs_gsms}
\end{subtable}
\hfill
\begin{subtable}[t]{0.31\textwidth}
\centering
\small
\begin{tabular}{l|c}
\toprule
\textbf{Model} & \textbf{DRS} \\
\toprule
    Llama-3.2-3B-it & 0.50 \\
    Llama-3.1-8B-it  & 0.54 \\
    Llama-3.1-70B-it  &  0.78 \\
    Qwen-2.5-7B-it  & 0.74 \\
    Qwen-2.5-32B-it  & \textbf{0.82} \\
    Qwen-2.5-72B-it  & 0.81 \\
    \bottomrule
\end{tabular}
\subcaption{FinChain}
\label{tab:drs_finchain}
\end{subtable}
\hfill
\begin{subtable}[t]{0.31\textwidth}
\centering
\small
\begin{tabular}{l|c}
\toprule
\textbf{Model} & \textbf{DRS} \\
\toprule
    Llama-3.2-3B-it & 0.24 \\
    Llama-3.1-8B-it  & 0.30 \\
    Qwen-2.5-7B-it  & \textbf{0.51} \\
    \bottomrule
\end{tabular}
\subcaption{EngTrace}
\label{tab:drs_engtrace}
\end{subtable}
\caption{\textbf{The difficulty-robustness score (DRS).} DRS is defined as the normalized area under the accuracy–quantile curve. Values closer to 1.0 indicate better robustness.}
\label{tab:drs_comparison}
\end{table}

% EngTrace template example
\begin{figure*}[tbp]
\centering
\noindent
\begin{tcolorbox}[
    title=\small{EngChain Template},
    fontupper=\scriptsize,
    fontlower=\scriptsize,
    colback=metabg,
    colframe=metafg,
    coltitle=white,
    colbacktitle=metafg,
    coltext=metafg,
    boxsep=3pt,
    top=4pt,
    bottom=4pt,
]
% Problem Statement
\textbf{\scriptsize\textcolor{metafg}{Problem}}
\vspace{2pt}
\hrule height 0.3pt
\vspace{4pt}

An undamped spring-mass system has a mass of \highlight{magenta30}{\{\textsc{mass}\}} kg and a spring stiffness of \highlight{yellow20}{\{\textsc{stiffness}\}} N/m. The mass is given an initial displacement of \highlight{blue20}{\{\textsc{initial\_disp\_mm}\}} mm and an initial velocity of \highlight{green20}{\{\textsc{initial\_vel\_ms}\}} m/s. Determine the displacement x(t) at t = 2.

\vspace{8pt}

% Variables
\textbf{\scriptsize\textcolor{metafg}{Variables}}
\vspace{2pt}
\hrule height 0.3pt
\vspace{4pt}

\begin{tabular}{@{}l@{\,}l@{}}
\highlight{magenta30}{\textsc{mass}} & = \{1.00, 1.01, $\dots$, 500.00\}\\
\highlight{yellow20}{\textsc{stiffness}} & = \{1000, 1001, $\dots$, 19999, 200000\}\\
\highlight{blue20}{\textsc{initial\_disp\_mm}} & = \{-100, -99, -98, $\dots$, 99, 100\}\\
\highlight{green20}{\textsc{initial\_vel\_ms}} & = \{-5.00, -4.99, -4.98, $\dots$, 4.99, 5.00\}\\
\end{tabular}

\vspace{8pt}

% Conditions
\textbf{\scriptsize\textcolor{metafg}{Conditions}}
\vspace{2pt}
\hrule height 0.3pt
\vspace{4pt}

\begin{tabular}{@{}l@{\,}l@{}}
\highlight{blue20}{\{\textsc{initial\_disp\_mm}\}} & $\neq$ 0\\
$|\text{\highlight{green20}{\{\textsc{initial\_vel\_ms}\}}}|$ & $\geq$ 0.01 \\

\end{tabular}

\vspace{8pt}

% Ground Truth
\textbf{\scriptsize\textcolor{metafg}{Ground-Truth Reasoning}}
\vspace{2pt}
\hrule height 0.3pt
\vspace{4pt}
Given:\\
Mass (m) = \highlight{magenta30}{\{\textsc{mass}\}} kg\\
Stiffness (k) = \highlight{yellow20}{\{\textsc{stiffness}\}} N/m\\
Initial Displacement (x(0)) = \highlight{blue20}{\{\textsc{initial\_disp\_mm}\}} mm\\
Initial Velocity (v(0)) = \highlight{green20}{\{\textsc{initial\_vel\_ms}\}} m/s

\vspace{5pt}

\textbf{Step 1}: Calculate the undamped natural frequency (omega\_n).\\
omega\_n = sqrt(k / m) = sqrt(\highlight{yellow20}{\{\textsc{stiffness}\}} / \highlight{magenta30}{\{\textsc{mass}\}}) rad/s

\vspace{5pt}

\textbf{Step 2:} State the general form of the solution for an undamped system.\\
The general solution is: x(t) = A1 * cos(omega\_n * t) + A2 * sin(omega\_n * t)

\vspace{5pt}

\textbf{Step 3:} Apply the initial conditions to find the constants A1 and A2.\\
First, apply the initial displacement at t=0:\\
x(0) = A1 * cos(0) + A2 * sin(0) = A1

\vspace{5pt}

Therefore, A1 = x(0) = \highlight{blue20}{\{\textsc{initial\_disp\_mm}\}} / 1000 m

\vspace{5pt}

Next, find the derivative of x(t) to get the velocity, v(t):\\
v(t) = dx/dt = -A1 * omega\_n * sin(omega\_n * t) + A2 * omega\_n * cos(omega\_n * t)

\vspace{5pt}
Now apply the initial velocity at t=0:\\
v(0) = -A1 * omega\_n * sin(0) + A2 * omega\_n * cos(0) = A2 * omega\_n

\vspace{5pt}
Therefore, A2 = v(0) / omega\_n = \highlight{green20}{\{\textsc{initial\_vel\_ms}\}} / sqrt(\highlight{yellow20}{\{\textsc{stiffness}\}} / \highlight{magenta30}{\{\textsc{mass}\}}) m

\vspace{5pt}

\textbf{Step 4:} Substitute the constants and omega\_n into the general solution.\\
x(t) = \highlight{blue20}{\{\textsc{initial\_disp\_mm}\}} / 1000 * cos(sqrt(\highlight{yellow20}{\{\textsc{stiffness}\}} / \highlight{magenta30}{\{\textsc{mass}\}}) * t) + (\highlight{green20}{\{\textsc{initial\_vel\_ms}\}} / sqrt(\highlight{yellow20}{\{\textsc{stiffness}\}} / \highlight{magenta30}{\{\textsc{mass}\}})) * sin(sqrt(\highlight{green20}{\{\textsc{initial\_vel\_ms}\}} / \highlight{magenta30}{\{\textsc{mass}\}} * t)

\end{tcolorbox}
\caption{\textbf{Example symbolic template from EngTrace.} As illustrated, the template captures a problem's underlying logical structure by defining variables, constraints, and solution steps, while treating specific numbers, names, and items as parameterizable variables.}
\label{fig:symb_template_example_engtrace}
\end{figure*}

% Qualitative examples
\begin{figure*}[ht]
  \centering
  % Right side: top row + 3x3 matrix
  \begin{minipage}[c]{0.99\textwidth}
    % Top row spanning all columns
    \begin{minipage}[t]{1.0\textwidth}
      \centering
      \input{tikz/qualitative_examples/gsms/template_55/template_problem}
    \end{minipage}
    
    \vspace{1em}
    
    % First row (low MD)
    \begin{minipage}[t!]{0.32\textwidth}
      \centering
      \input{tikz/qualitative_examples/gsms/template_55/low_md_example_1_qwen_7b}
    \end{minipage}
    \hspace{0.005\textwidth}
    \begin{minipage}[t!]{0.32\textwidth}
      \centering
      \input{tikz/qualitative_examples/gsms/template_55/medium_md_example_1_qwen_7b}
    \end{minipage}
    \hspace{0.005\textwidth}
    \begin{minipage}[t!]{0.32\textwidth}
      \centering
      \input{tikz/qualitative_examples/gsms/template_55/high_md_example_1_qwen_7b}
    \end{minipage} 
    
    \vspace{1em}
    % Second row (medium MD)
    \begin{minipage}[t!]{0.32\textwidth}
      \centering
      \input{tikz/qualitative_examples/gsms/template_55/low_md_example_2_qwen_7b}
    \end{minipage}
    \hspace{0.005\textwidth}
    \begin{minipage}[t!]{0.32\textwidth}
      \centering
      \input{tikz/qualitative_examples/gsms/template_55/medium_md_example_2_qwen_7b}
    \end{minipage}
    \hspace{0.005\textwidth}
    \begin{minipage}[t!]{0.32\textwidth}
      \centering
      \input{tikz/qualitative_examples/gsms/template_55/high_md_example_2_qwen_7b}
    \end{minipage}
  \end{minipage}
  
  \vspace{1em}
  
  \begin{tikzpicture}
    \draw[->, thick, color=metafg] (0,0) -- (14.5,0) node[right] {$\mathbf{\text{MD}_\mathcal{H}}$};
  \end{tikzpicture}
  
  \caption{\textbf{Responses from Qwen-2.5-7B-Instruct to problem variations from template 55 of GSM-Symbolic.} Examples show how responses deviate more from correct reasoning as $\text{MD}_\mathcal{H}$ increases.}
  \label{fig:qualitative_examples_template_55_qwen_7b_instruct}
\end{figure*}

\begin{figure*}[ht]
  \centering
  
  % Right side: top row + 3x3 matrix
  \begin{minipage}[c]{0.99\textwidth}
    % Top row spanning all columns
    \begin{minipage}[t]{1.0\textwidth}
      \centering
      \input{tikz/qualitative_examples/gsms/template_74/template_problem}
    \end{minipage}
    
    \vspace{1em}
    
    % First row (low MD)
    \begin{minipage}[t!]{0.32\textwidth}
      \centering
      \input{tikz/qualitative_examples/gsms/template_74/low_md_example_1_qwen_7b}
    \end{minipage}
    \hspace{0.005\textwidth}
    \begin{minipage}[t!]{0.32\textwidth}
      \centering
      \input{tikz/qualitative_examples/gsms/template_74/medium_md_example_1_qwen_7b}
    \end{minipage}
    \hspace{0.005\textwidth}
    \begin{minipage}[t!]{0.32\textwidth}
      \centering
      \input{tikz/qualitative_examples/gsms/template_74/high_md_example_1_qwen_7b}
    \end{minipage} 
  \end{minipage}
  
  \vspace{1em}
  
  \begin{tikzpicture}
    \draw[->, thick, color=metafg] (0,0) -- (14.5,0) node[right] {$\mathbf{\text{MD}_\mathcal{H}}$};
  \end{tikzpicture}
  
  \caption{\textbf{Responses from Qwen-2.5-7B-Instruct to problem variations from template 74 of GSM-Symbolic.} Examples show how responses deviate more from correct reasoning as $\text{MD}_\mathcal{H}$ increases.}
  \label{fig:qualitative_examples_template_74_qwen_7b_instruct}
\end{figure*}

\begin{figure*}[ht]
  \centering
  
  % Right side: top row + 3x3 matrix
  \begin{minipage}[c]{0.99\textwidth}
    % Top row spanning all columns
    \begin{minipage}[t]{1.0\textwidth}
      \centering
      \input{tikz/qualitative_examples/gsms/template_74/template_problem}
    \end{minipage}
    
    \vspace{1em}
    
    % Second row (medium MD)
    \begin{minipage}[t!]{0.32\textwidth}
      \centering
      \input{tikz/qualitative_examples/gsms/template_74/low_md_example_2_qwen_7b}
    \end{minipage}
    \hspace{0.005\textwidth}
    \begin{minipage}[t!]{0.32\textwidth}
      \centering
      \input{tikz/qualitative_examples/gsms/template_74/medium_md_example_2_qwen_7b}
    \end{minipage}
    \hspace{0.005\textwidth}
    \begin{minipage}[t!]{0.32\textwidth}
      \centering
      \input{tikz/qualitative_examples/gsms/template_74/high_md_example_2_qwen_7b}
    \end{minipage}
  \end{minipage}
  
  \vspace{1em}
  
  \begin{tikzpicture}
    \draw[->, thick, color=metafg] (0,0) -- (14.5,0) node[right] {$\mathbf{\text{MD}_\mathcal{H}}$};
  \end{tikzpicture}
  
  \caption{\textbf{Further responses from Qwen-2.5-7B-Instruct to problem variations from template 74 of GSM-Symbolic.} Examples show how responses deviate more from correct reasoning as $\text{MD}_\mathcal{H}$ increases.}
  \label{fig:qualitative_examples_template_74_b_qwen_7b_instruct}
\end{figure*}

\begin{figure*}[ht]
  \centering
  
  % Right side: top row + 3x3 matrix
  \begin{minipage}[c]{0.99\textwidth}
    % Top row spanning all columns
    \begin{minipage}[t]{1.0\textwidth}
      \centering
      \input{tikz/qualitative_examples/gsms/template_79/template_problem}
    \end{minipage}
    
    \vspace{1em}
    
    % First row (low MD)
    \begin{minipage}[t!]{0.32\textwidth}
      \centering
      \input{tikz/qualitative_examples/gsms/template_79/low_md_example_1_qwen_7b}
    \end{minipage}
    \hspace{0.005\textwidth}
    \begin{minipage}[t!]{0.32\textwidth}
      \centering
      \input{tikz/qualitative_examples/gsms/template_79/medium_md_example_1_qwen_7b}
    \end{minipage}
    \hspace{0.005\textwidth}
    \begin{minipage}[t!]{0.32\textwidth}
      \centering
      \input{tikz/qualitative_examples/gsms/template_79/high_md_example_1_qwen_7b}
    \end{minipage} 
    
    \vspace{1em}
    % Second row (medium MD)
    \begin{minipage}[t!]{0.32\textwidth}
      \centering
      \input{tikz/qualitative_examples/gsms/template_79/low_md_example_2_qwen_7b}
    \end{minipage}
    \hspace{0.005\textwidth}
    \begin{minipage}[t!]{0.32\textwidth}
      \centering
      \input{tikz/qualitative_examples/gsms/template_79/medium_md_example_2_qwen_7b}
    \end{minipage}
    \hspace{0.005\textwidth}
    \begin{minipage}[t!]{0.32\textwidth}
      \centering
      \input{tikz/qualitative_examples/gsms/template_79/high_md_example_2_qwen_7b}
    \end{minipage}
  \end{minipage}
  
  \vspace{1em}
  
  \begin{tikzpicture}
    \draw[->, thick, color=metafg] (0,0) -- (14.5,0) node[right] {$\mathbf{\text{MD}_\mathcal{H}}$};
  \end{tikzpicture}
  
  \caption{\textbf{Responses from Qwen-2.5-7B-Instruct to problem variations from template 79 of GSM-Symbolic.} Examples show how responses deviate more from correct reasoning as $\text{MD}_\mathcal{H}$ increases.}
  \label{fig:qualitative_examples_template_79_qwen_7b_instruct}
\end{figure*}

% Prompts
\begin{figure*}[tbp]
\centering
\noindent
\input{tikz/prompts/gsms_prompt}
\caption{\textbf{The 5-shot CoT prompt for samples from GSM-Symbolic.} The final \{question\} is substituted with the corresponding problem variation.}
\label{fig:gsms_prompt}
\end{figure*}

\begin{figure*}[tbp]
\centering
\noindent
\input{tikz/prompts/finchain_prompt}
\caption{\textbf{The 5-shot CoT prompt for samples from FinChain.} The final \{question\} is substituted with the corresponding problem variation.}
\label{fig:finchain_prompt}
\end{figure*}

\begin{figure*}[tbp]
\centering
\noindent
\input{tikz/prompts/engtrace_prompt}
\caption{\textbf{The 5-shot CoT prompt for samples from EngTrace.} The final \{question\} is substituted with the corresponding problem variation.}
\label{fig:engtrace_prompt}
\end{figure*}

\end{document}